\documentclass[11pt]{article}

\usepackage[preprint]{acl}
\usepackage{times}
\usepackage{latexsym}
\usepackage[T1]{fontenc}
\usepackage[utf8]{inputenc}
\usepackage{microtype}
\usepackage{inconsolata}
\usepackage{graphicx}
\usepackage{amsmath}
\usepackage{soul}
\usepackage{booktabs}
\usepackage{enumitem}
\usepackage{float}

\usepackage{xcolor}
\usepackage{colortbl}
\usepackage{etoolbox}
\definecolor{tabhead}{HTML}{DFE9F6}
\definecolor{tabours}{HTML}{EEF5F1}
\definecolor{tabzebra}{HTML}{F5F7FA}
\newcommand{\headrow}{\rowcolor{tabhead}}
\newcommand{\zebrarow}{\rowcolor{tabzebra}}
\newcommand{\oursrow}{\rowcolor{tabours}}
\newcommand{\headrulebg}{\noalign{{\color{tabhead}\hrule height\cmidrulewidth}\vskip-\cmidrulewidth}}
\AtBeginEnvironment{table}{}
\AtBeginEnvironment{table*}{}
\usepackage{tcolorbox}
\tcbuselibrary{skins,breakable}
\usepackage{pifont}
\usepackage{etoc}
\usepackage{newunicodechar}
\newunicodechar{—}{\textemdash}

\title{ReviewPilot: A Multi-Agent System for Screening and Data Extraction in Systematic Reviews}
\author{
 \textbf{Zhaoqian Xue\textsuperscript{1}\thanks{Corresponding authors: Zhaoqian Xue and Lingyao Li.}},
 \textbf{Shaoze Zhou\textsuperscript{2}},
 \textbf{Xuexin Li\textsuperscript{3}},
 \textbf{Junting Mu\textsuperscript{4}},
 \textbf{Haozheng Liu\textsuperscript{5}},
\\
 \textbf{Zhongyin Wang\textsuperscript{6}},
 \textbf{Yichun He\textsuperscript{7}},
 \textbf{Renkai Ma\textsuperscript{8}},
 \textbf{Hao Zhou\textsuperscript{9}},
\\
 \textbf{Sarah Bratt\textsuperscript{1}},
 \textbf{Siyuan Ma\textsuperscript{10}},
 \textbf{Chen Chen\textsuperscript{2}},
 \textbf{Lingyao Li\textsuperscript{1}\footnotemark[2]}
\\[0.5ex]
 \normalsize
 \textsuperscript{1}University of Arizona,
 \textsuperscript{2}Florida International University,
\\
 \normalsize
 \textsuperscript{3}Vanderbilt University,
 \textsuperscript{4}University of California, Irvine,
 \textsuperscript{5}New York University,
\\
 \normalsize
 \textsuperscript{6}University of Illinois Urbana-Champaign,
 \textsuperscript{7}Broad Institute of MIT and Harvard,
\\
 \normalsize
 \textsuperscript{8}University of Cincinnati,
 \textsuperscript{9}University of South Florida,
 \textsuperscript{10}Vanderbilt University Medical Center
}

\begin{document}
\maketitle

\begin{abstract}
Systematic literature reviews follow the PRISMA process of screening scholarly works and extracting evidence from included studies. Deep research tools based on large language models (LLMs) can search and summarize the literature but do not record which studies are excluded or where extracted values come from. We develop a conversational multi-agent system called ReviewPilot that automates the PRISMA workflow from paper screening to information extraction. Researchers can tailor criteria and schemas through dialogue, and a project memory keeps their configurations and revisions. Each screening decision comes with its reason and each extracted value with its source. On epidemiological expert reviews, ReviewPilot outperforms a single-prompt LLM and AgentSLR across three LLMs, with screening macro F1 up to 0.776 and extraction F1 up to 0.788. Case studies in biomedicine and human-computer interaction show close agreement with human annotators, suggesting that ReviewPilot can transfer beyond epidemiology. Code is available at \href{https://github.com/ZhaoqianXue/ReviewPilot}{this link}.
\end{abstract}

\section{Introduction}
\label{sec:introduction}
\vspace{-2mm}

\begin{figure*}[t]
  \centering
  \vspace{-4mm}
  \includegraphics[width=\textwidth]
  {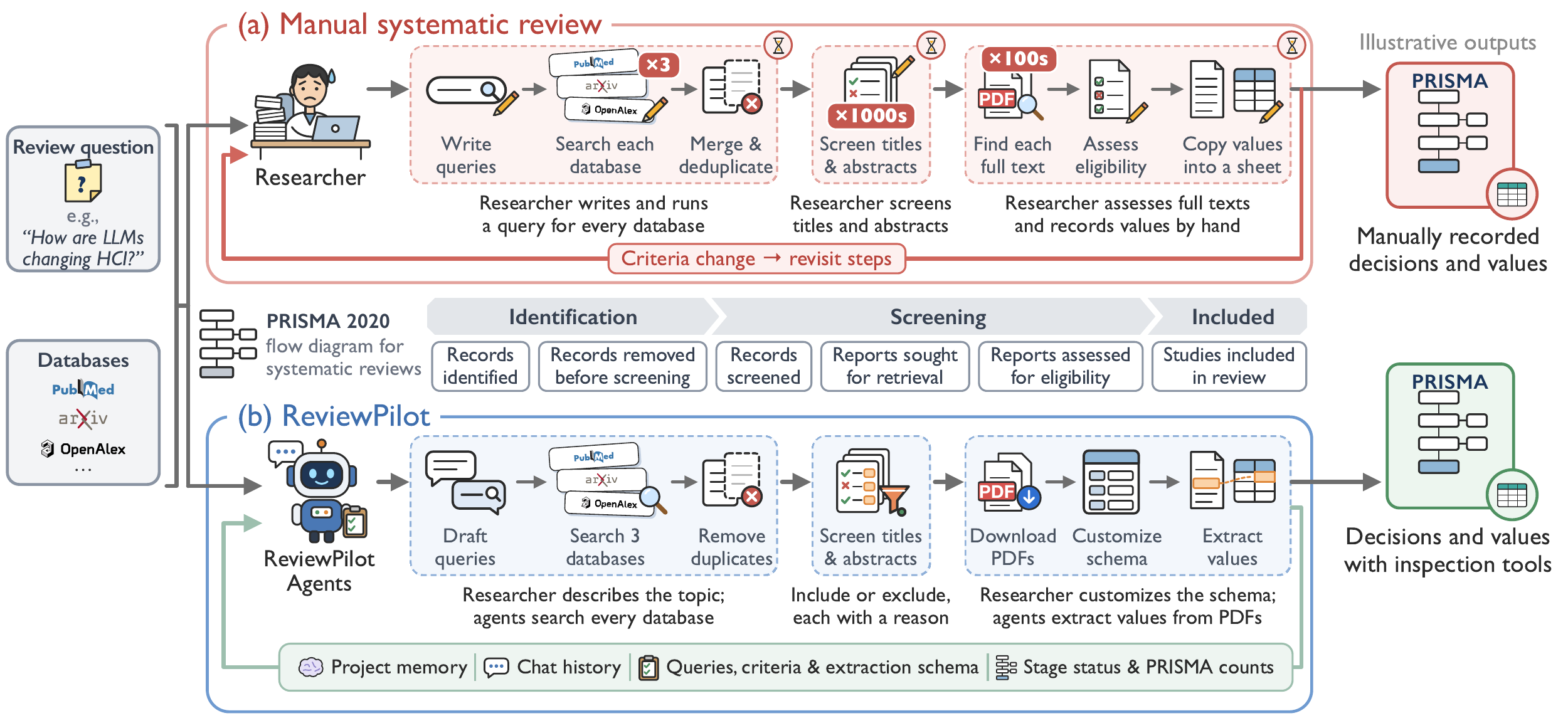}
  \caption{Manual and ReviewPilot-assisted systematic review workflows, aligned with PRISMA 2020 stages \citep{page2021prisma}. ReviewPilot coordinates search, screening, and extraction while supporting researcher inspection and revision. Project memory retains configurations, outputs, and corrections.}
  \label{fig:overview}
  \vspace{-4mm}
\end{figure*}

Systematic literature reviews depend on a sequence of judgments, including eligibility check, full-text retrieval, and information extraction \citep{higgins2023cochrane,page2021prisma}. Such a process if labor-intensive. Searches often result in high recall, so reviewers need to read thousands of titles and abstracts to find a few eligible studies \citep{van2021open}. A review can take over a year on average from registration to publication \citep{borah2017analysis}. Automation can reduce this workload, but researchers still need to inspect decisions and document the process based on the PRISMA (Preferred Reporting Items for Systematic Reviews and Meta-Analyses) 2020 process \citep{page2021prisma}.

Large language models (LLMs) are now widely used to search and summarize literature. Deep research tools such as ChatGPT \citep{openai2025deepresearch}, Gemini \citep{gemini2024deepresearch}, and Perplexity Research \citep{perplexity2025deepresearch} can generate a synthesized report. These tools help researchers explore a topic but do not follow the PRISMA process. They do not show which papers are screened or excluded and why, or where each extracted value comes from. Other review tools only limit these decisions to a single stage, such as ASReview \citep{van2021open} and LGAR \citep{jaumann2025lgar} for screening, and RobotReviewer \citep{marshall2017automating} for extraction, or an agent-based system for a few stages like AgentSLR \citep{padarha2026evaluating}.

However, designing an AI-assisted literature review workflow involves two key considerations. \textbf{First, screening criteria and field definitions need operational instructions.} Domain terms and borderline cases need to be iteratively interpreted consistently for screening and coding. \textbf{Second, outputs must reflect the current configuration.} Researchers refine criteria and fields as a review progresses \citep{higgins2023cochrane}; confirmed settings and corrections must therefore be retained, and affected downstream outputs identified. We implement these requirements within a workflow that records the counts and inclusion-exclusion criteria needed for PRISMA-guided workflow.

We present \textbf{ReviewPilot}, a conversational multi-agent system that addresses these challenges from search through information extraction (Figure~\ref{fig:overview}). Researchers define eligibility criteria and an extraction schema through natural dialogue with the system, and a Prompt Agent turns them into review-specific guidance. Agents screen records and extract information, delivering screening decisions with their reasons and extracted values linked to source passages in the papers. The result is a step-by-step record of how papers are filtered, with the records kept and excluded at each stage, and a table of the information extracted from each retrieved paper. Our contributions are as follows.

\begin{itemize}[leftmargin=*,nosep] 
\item \textbf{PRISMA-guided workflow.} ReviewPilot coordinates search, screening, retrieval, and extraction while recording counts, exclusion reasons, and available source for inspection (Section~\ref{sec:methods}). 
\item \textbf{Review-specific configuration and revision.} Researchers can set criteria and schemas through dialogue. Then, agents draft screening guidance and coding rules. Researchers can edit these instructions and confirm. Project memory design further retains these configurations and corrections (Sections~\ref{sec:prompt-design} and~\ref{sec:memory-interface}). 
\item \textbf{System-level evaluation.} On PERG, ReviewPilot achieves higher screening macro F1 and extraction F1 than a single-prompt LLM and AgentSLR under the reported settings (Tables~\ref{tab:screening} and~\ref{tab:extraction}). Two case studies document completed workflows and screening agreement with human annotators (Section~\ref{sec:case-studies}). 
\end{itemize}

\section{Related Work}
\label{sec:literature}
\vspace{-2mm}

\paragraph{LLM multi-agent systems.} Multi-agent frameworks divide tasks among LLM agents with distinct roles \citep{guo2024large}. For instance, AutoGen supports configurable conversations that combine LLMs, tools, and human input \citep{wu2024autogen}. MetaGPT encodes standard operating procedures as prompt sequences to coordinate software development and check intermediate results \citep{hong2024metagpt}. Our system, ReviewPilot, leverages the potential of multi-agent workflow to review stages.

\par\vspace{2mm}
\noindent\textbf{Deep research and literature synthesis.} Deep research tools in ChatGPT \citep{openai2025deepresearch}, Gemini \citep{gemini2024deepresearch}, and Perplexity \citep{perplexity2025deepresearch} iteratively search and analyze sources to produce reports. ChatGPT or Gemini can link citations and allows users to revise its research plan. For scientific literature, OpenScholar combines passage retrieval, cited generation, and iterative self-feedback \citep{asai2026synthesizing}, while AutoSurvey retrieves papers, constructs an outline, and delegates subsection drafting to specialized LLMs \citep{wang2024autosurvey}. These systems can synthesize reports but do not document the paper filtering process based on PRISMA 2020 \citep{page2021prisma}.

\par\vspace{2mm}
\noindent\textbf{Systematic review tools.} Some tools support individual review stages. ASReview prioritizes records for human screening through active learning from relevance judgments \citep{van2021open}; LGAR ranks abstracts zero-shot using graded, criterion-based LLM scores and dense reranking \citep{jaumann2025lgar}. RobotReviewer extracts trial characteristics and supports risk-of-bias assessment with sentences linked to the source document \citep{marshall2017automating}.

Integrated workflows extend assistance across stages. AgentSLR pairs an epidemiological literature review workflow with stage-specific evaluation and describes an interface for expert verification of extracted values \citep{padarha2026evaluating}. Its published workflow uses predefined eligibility criteria and domain-specific extraction schemas. We compare against it in Section~\ref{sec:experiments}. \citet{ren2026systematic} translate user criteria into screening rules and use multi-agent cross-review for clinical-trial screening. Their extraction workflow combines source-derived table standardization, iterative correction, and targeted human review. The commercial Elicit workflow proposes editable criteria and extraction fields and lets users return to earlier steps \citep{byun2025elicit}; it also provides evidence quotes, exclusion reasons, and PRISMA 2020 flow diagrams \citep{vardhaan2026elicit}. These workflows already combine stage automation, customization, evidence access, and human oversight.

To address these gaps, we develop ReviewPilot, a PRISMA-guided multi-agent system that coordinates literature search, screening, full-text retrieval, and information extraction. Researchers can tailor eligibility criteria and extraction schemas through dialogue, while agents generate screening guidance and per-field coding rules. Project memory retains configurations and corrections to support revision across stages (Sections~\ref{sec:prompt-design} and~\ref{sec:memory-interface}). We evaluate the complete system on PERG and examine its workflows and screening agreement with human annotators in two case studies.

\section{ReviewPilot Development}
\label{sec:methods}

\vspace{-2mm}

\begin{figure*}[t]
  \centering
  \vspace{-2mm}
  \includegraphics[width=\textwidth]{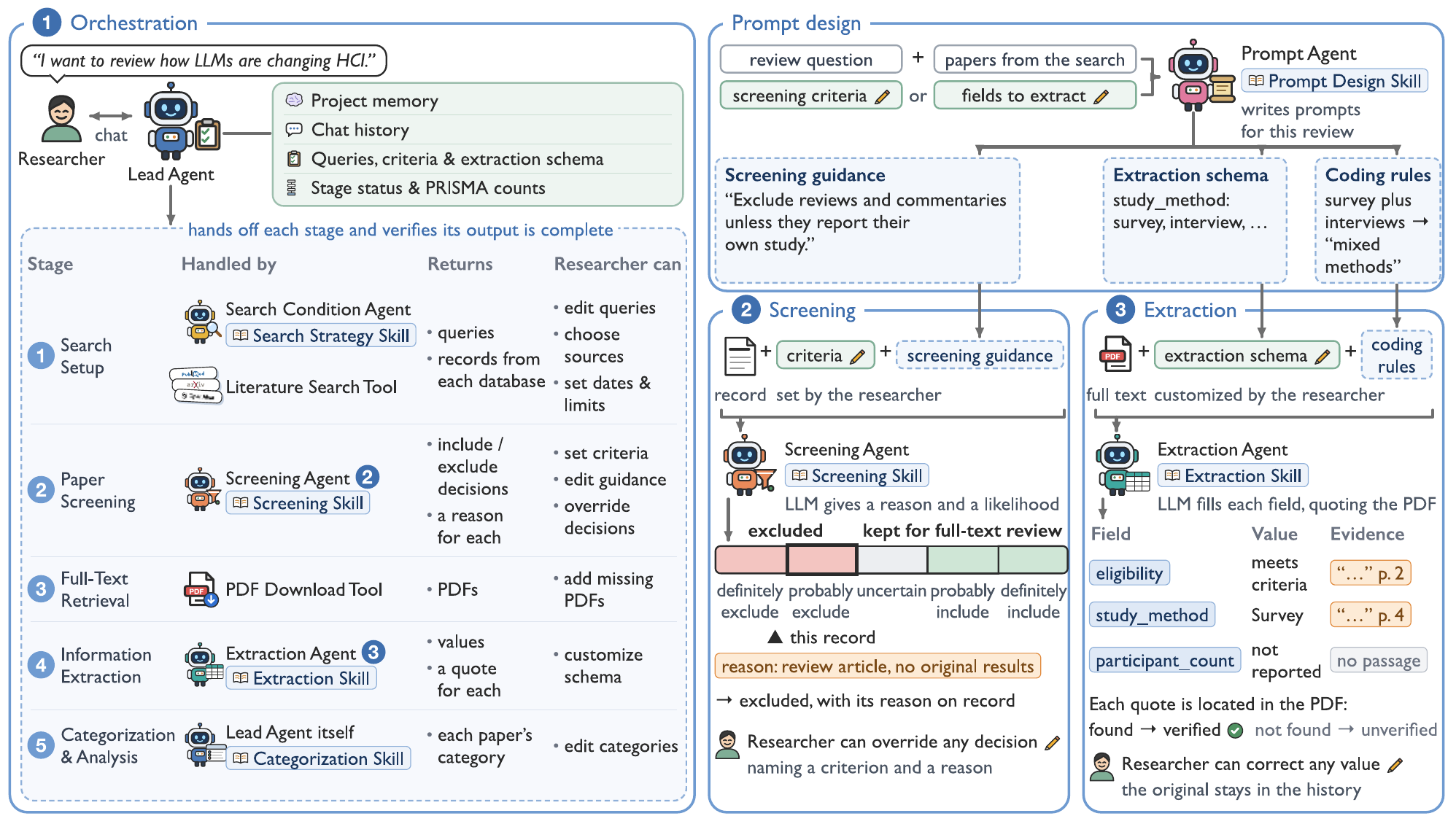}
  \caption{ReviewPilot architecture and workflow. (1)~The Lead Agent coordinates five review stages using project memory. Researchers configure stages and correct decisions or values. In the upper right, the Prompt Agent prepares screening guidance, an extraction schema, and coding rules without reviewer labels. (2)~Screening returns a rationale and one of five likelihood levels. (3)~Extraction returns values with verbatim excerpts. The label \emph{verified} means the excerpt is found in the PDF text. Orange marks screening rationales and extraction evidence.}
  \label{fig:architecture}
  \vspace{-4mm}
\end{figure*}

ReviewPilot is a conversational multi-agent system that covers five stages of a systematic literature review: search, screening, full-text retrieval, information extraction, and semantic categorization (Figure~\ref{fig:architecture}). Researchers can configure the review through dialogue or direct editing, inspect each stage's outputs, and can override screening decisions or correct extracted values.

\subsection{Architecture and Orchestration}
\label{sec:agents}

A Lead Agent coordinates four specialized agents and two tools through action contracts (Appendix~\ref{app:impl-orchestration}). The Search Condition Agent designs the search strategy, and the Prompt Agent writes review-specific instructions (Section~\ref{sec:prompt-design}). The Screening Agent deduplicates and assesses records, and the Extraction Agent extracts structured information. The Literature Search Tool executes queries, and the PDF Download Tool retrieves full texts. The Lead Agent categorizes values itself.

The Lead Agent starts a stage only when its prerequisites are met and the outputs it depends on are current. It reports completion only after checking the expected artifacts and their required fields. Agent skills are structured Markdown instructions specifying procedures for search strategy, screening, prompt design, extraction, and categorization. A fixed mapping appends the matching skill to the LLM system prompts for these tasks (Appendix~\ref{app:prompts}).

\subsection{Review-Specific Prompt Design}
\label{sec:prompt-design}

The Prompt Agent writes the review-specific parts of the screening and extraction instructions, following the prompt-design skill. After the eligibility criteria are confirmed, it writes screening guidance $\mathcal{G}$ from the review objectives, criteria, search concepts, and up to 200 candidate records (titles and abstract openings). This guidance defines the review focus, domain terms, include and exclude categories, and tie-breaking rules. After screening, it drafts an extraction schema from the objectives, the requested fields, and the abstract openings of up to three included papers. Once the schema is confirmed, it writes the coding rules $\mathcal{C}$ from up to 25 included papers, using their abstract openings and full-text sentences selected for each field.

The guidance and the coding rules are each drafted and then revised in a second LLM pass against the skill's checklist \citep{madaan2023self}. Reviewer decisions and reference extraction values are not among the generation inputs. Researchers can edit the guidance, schema, and rules and must confirm each before the stage that uses it runs (Appendices~\ref{app:prompt-agent-prompts}, \ref{app:perg-guidance}, and~\ref{app:perg-coding}).

\subsection{Search \& Screening}
\label{sec:screening}

\paragraph{Search.} \label{sec:query-generation}The Search Condition Agent identifies the concepts required for eligibility and proposes alternative terms for each concept. It builds a Boolean query by joining alternatives with OR and required concept groups with AND (Appendix~\ref{app:impl-search}). Analytical dimensions that do not determine eligibility are omitted from the required groups. Researchers refine the concepts and query through dialogue or direct editing and review the sources, date range, and per-source limits. The Literature Search Tool then searches titles and abstracts in PubMed, arXiv, and OpenAlex. The saved query, settings, counts, and errors document the search as run.

\par\vspace{2mm} \noindent\textbf{Screening.} After date filtering and deduplication (Appendix~\ref{app:impl-screening}), the model assesses each record $p_i$, its title and available abstract, against the confirmed inclusion criteria $\mathcal{I}$, exclusion criteria $\mathcal{E}$, and screening guidance $\mathcal{G}$: \vspace{-2mm} \begin{equation}
(r_i, \ell_i) = f_{\text{LLM}}(p_i, \mathcal{I}, \mathcal{E}, \mathcal{G}),
\end{equation} where $r_i$ is a one-sentence rationale naming the deciding rule and $\ell_i$ is a qualitative likelihood label. Its five levels are \emph{definitely exclude}, \emph{probably exclude}, \emph{uncertain}, \emph{probably include}, and \emph{definitely include}. The screening skill instructs the model to assess the study's own work against the review focus and criteria, not topics mentioned only as background (Appendix~\ref{app:screen-prompt}). Among these five levels, a record is excluded only when $\ell_i$ is \emph{probably exclude} or \emph{definitely exclude}. Records with the other labels are retained for full-text review. Records are also retained when the model call fails. The prompt requests the deciding rule and a verbatim title or abstract excerpt. An exclusion with an unmatched rule or unlocated excerpt is flagged for human review, but the record remains excluded.

\subsection{Paper Retrieval \& Information Extraction}
\label{sec:extraction}

\paragraph{Paper retrieval.} \label{sec:retrieval}The PDF Download Tool tries a cascade of publisher, open-access, and preprint sources for each retained record. It checks PDF validity and compares opening-page text with the title. These heuristics do not guarantee retrieval success (Appendix~\ref{app:impl-retrieval}), so it generates a report to record the success or failure retrieval for each paper, and researchers can add missing PDFs manually.

\par\vspace{2mm} \noindent\textbf{Information extraction.} The confirmed schema $\mathcal{S}=\{(f_j,\tau_j,\delta_j)\}_{j=1}^{k}$ defines each field by its name, type, and description (Appendix~\ref{app:impl-extraction}). The coding rules $\mathcal{C}$ specify how borderline values are coded. ReviewPilot places a full-text eligibility field first in the drafted schema, with options from the confirmed criteria. The model judges whether the paper remains eligible and names the criterion behind an exclusion. In a PDF assessment, an empty or unrecognized eligibility value does not exclude the paper. Excluded papers are omitted from categorization, and their exclusion reasons are recorded for PRISMA reporting.

For each available PDF, the Extraction Agent sends up to the first 400,000 characters of its text, with page markers, together with the schema and coding rules to the model (Appendix~\ref{app:extract-prompt}). For each field, the prompt requests a value $v_j$, verbatim excerpts, and page numbers. It requires an empty or null value when the supplied text neither states nor necessarily implies a value. Responses are checked for field presence, types, and value constraints. A failed check is recorded as an extraction failure for that paper. Passing these checks does not establish that the values are correct. ReviewPilot searches the normalized PDF text for the returned excerpts. A field with at least one located excerpt is marked \emph{verified}, and the first located excerpt is stored with the page on which it was located. Otherwise, the field is \emph{unverified}. Excerpt location establishes textual presence, not semantic support for the value or coverage of every value in a multi-value field.

For papers whose PDF was not retrieved, researchers can enable assessment from the abstract. When the selected model supports web search, it can seek web evidence for fields the abstract leaves open. Accepted values require located excerpts, and web pages must pass paper-identity checks. These outputs retain source labels and any source URLs without PDF page numbers and are reported separately from full-text assessments. Fields without evidence remain unavailable, and papers whose eligibility remains unavailable are not categorized.

\subsection{Semantic Categorization}
\label{sec:categorization}

The Lead Agent categorizes extracted values only for papers retained after eligibility check. For a selected field, it proposes a set of categories with short descriptions (Appendices~\ref{app:impl-categorization} and~\ref{app:categorization-prompts}). The plan supports single-label or multiple-label assignment, and an \emph{Other} category for values that do not fit. Researchers review and edit the plan before the Lead Agent assigns labels to each paper. Papers with no value for the field are marked unavailable.

\subsection{Memory Design}
\label{sec:memory-interface}

Researchers work in a conversation panel and a stage-specific canvas. A chat request that changes criteria or the schema is saved as a draft, and confirmation is a canvas action (Appendices~\ref{app:impl-interface} and~\ref{app:interface}). Project memory stores the conversation history, current configurations, confirmed configuration snapshots, and each stage's status and PRISMA counts.

In conversation, the Lead Agent receives its task instructions, the current request, the saved project state, and the conversation history. Its instructions give saved state precedence over superseded requests and distinguish drafts from confirmed configurations. Researchers reuse a previous project's configuration by importing it as a draft that again requires confirmation. The system does not retrieve memories across projects automatically or learn preferences from corrections.

To override a screening decision, researchers select a confirmed criterion or guidance rule and give a reason. They also give reasons when correcting extracted values. Original decisions and values remain in review history. A change to a confirmed configuration or reviewed record marks affected downstream outputs as outdated until rerun.

\section{Experiments \& Results}
\label{sec:experiments}

\vspace{-2mm}

We evaluate screening and structured information extraction separately against the expert annotations in PERG, a dataset released with AgentSLR \citep{padarha2026evaluating}. Extraction uses the same expert-included papers for all methods, independently of each method's screening predictions.

\subsection{Experiment Setup}
\label{sec:setup}

\paragraph{Data and reference annotations.} PERG is released under the CC BY 4.0 license. Its annotations originate from systematic reviews conducted by the Pathogen Epidemiology Review Group at Imperial College London for WHO priority pathogens \citep{cuomo2024marburg, nash2024ebola, doohan2024lassa, morgenstern2024sars, mccain2025zika, naidoo2026nipah}. Screening uses 5,614 title--abstract records for Lassa, Nipah, SARS, and Zika, with 854 expert inclusions (15.2\%). All methods apply the five inclusion and six exclusion criteria specified in AgentSLR's PERG template (Appendix~\ref{app:perg-data}). All records are evaluated without date filtering or deduplication.

Extraction uses 178 annotated papers (79 for Ebola, 6 for Lassa, 36 for SARS, and 57 for Zika). We evaluate five fields: model type, stochastic or deterministic status, code availability, transmission routes, and interventions (Table~\ref{tab:perg-fields}). PERG annotates each transmission model in a paper. Reference values are aggregated to the paper level by taking the union across models. Code availability is true if the paper reports available code for any model.

\par\vspace{2mm} \noindent\textbf{Models and evaluation metrics.} We use gpt-6-luna through the OpenAI API,\footnote{\url{https://developers.openai.com/api/docs}} and DeepSeek V4.1 Flash and GLM-5.3-Flash through Together AI.\footnote{\url{https://docs.together.ai/intro}} The \emph{single-prompt LLM} uses one instruction containing the screening criteria, or the study objective and field definitions for extraction. It retains uncertain screening cases and requests field-level excerpts for extraction, without ReviewPilot's skills, coding rules, or schema validation (Appendix~\ref{app:baselines}).

We rerun \emph{AgentSLR} using its public screening implementation and transmission-model extraction pipeline. Its prompts, parser, and internal full-text screening are preserved, as are the tool and provenance loops of its extraction stage. It reads the same PDF text as the other methods. We adapt AgentSLR's API client and replace its original high reasoning setting (Table~\ref{tab:agentslr-adaptation}). Extraction and gpt-6-luna screening use medium reasoning. Screening with the Together AI models uses provider defaults. Papers rejected by its internal screening remain in the evaluation with empty predictions.

ReviewPilot follows the screening and extraction procedures in Sections~\ref{sec:screening} and~\ref{sec:extraction}. It uses a fixed five-field extraction schema, with review-specific screening guidance and coding rules generated by the Prompt Agent. All methods receive the same papers and use the same reference annotations.

Prompt Agent uses the same model as its downstream stage. The agent receives the review objectives, criteria or field definitions, and samples of evaluation records or papers. Expert decisions and reference extraction values are excluded from the agent's inputs. ReviewPilot's screening procedure and prompt-design skill are refined using evaluation scores from these PERG pools. The pools are therefore not held out from method development.

We report one complete run per method and model. Decoding settings and output limits differ across methods and are recorded with each run (Appendix~\ref{app:run-config}). Transport errors and incomplete responses trigger retries. ReviewPilot also retries unparseable screening outputs. GLM-5.3-Flash exhausted its 65,536-token output limit on one paper. That paper is excluded from all three methods' results on this model, leaving 177 papers. All table entries are single-run point estimates and do not characterize variation across repeated model runs.

\par\vspace{2mm} \noindent\textbf{Screening metrics.} We pool the records and compute precision, recall, and F1 separately for the include and exclude classes. Each metric is then macro-averaged over the two classes. The \emph{review share}, $(\mathrm{TP}+\mathrm{FP})/N$, is the proportion of records retained for full-text review, where TP and FP denote retained expert-included and expert-excluded records, and $N$ is the pool size. Macro recall averages recall for both classes and does not measure include-class recall alone.

\par\vspace{2mm} \noindent\textbf{Extraction metrics.} Predicted and reference values are normalized with a fixed synonym mapping shared by all methods. At the paper--field level, matching values count as true positives, additional predicted values as false positives, and reference values absent from predictions as false negatives. Under AgentSLR's empty-field convention, a field empty in both prediction and reference counts as one true positive. This convention applies to model type, stochastic or deterministic status, and code availability. For paper-level scoring, AgentSLR's per-model predictions are aggregated identically to reference values. Its native model-matching evaluation is not used. Counts are pooled across papers and the five fields to compute precision, recall, and F1. AgentSLR's additional fields are not scored.

The \emph{verified-correct rate} is computed over paper--field pairs with a reference value. A pair counts as verified-correct when its complete normalized prediction matches the reference set and at least one returned excerpt is located in the paper. All methods use the same excerpt matcher. Excerpt location does not establish semantic support for the prediction \citep{gao2023enabling} or coverage of every value in a multi-value field.

\par\vspace{2mm} \noindent\textbf{Cost estimation.} We estimate API costs from returned token usage and the prices in Table~\ref{tab:models}. Costs are reported in USD per 1,000 screened records or per extraction paper. Screening costs include ReviewPilot's guidance generation, whereas extraction costs exclude coding-rule generation.

\subsection{Screening Results}
\label{sec:screening-results}

\begin{table}[t]
\centering
\caption{Screening results pooled over four PERG pathogens (5,614 records with 854 expert inclusions). Precision, recall, and F1 are macro-averaged over the include and exclude classes. Review share is the proportion of records retained for full-text review. Cost is USD per 1,000 records. Bold marks the best value within each model (highest for $\uparrow$, lowest for $\downarrow$).}
\vspace{-1mm}
\label{tab:screening}
\footnotesize
\setlength{\tabcolsep}{1.25pt}
\begin{tabular}{lccccc}
\toprule
\headrow & Macro & Macro & Macro & Review & Cost$\downarrow$ \\
\headrow Method & precision$\uparrow$ & recall$\uparrow$ & F1$\uparrow$ & share$\downarrow$ & (USD) \\
\midrule
\multicolumn{6}{l}{\textit{gpt-6-luna}} \\
\zebrarow \shortstack[l]{\rule{0pt}{1em}Single-prompt\\LLM} & 0.685 & 0.828 & 0.696 & 0.354 & \textbf{0.13} \\
AgentSLR & 0.694 & 0.833 & 0.713 & 0.331 & 0.30 \\
\oursrow ReviewPilot & \textbf{0.734} & \textbf{0.861} & \textbf{0.766} & \textbf{0.274} & 0.23 \\
\midrule
\multicolumn{6}{l}{\textit{DeepSeek V4.1 Flash}} \\
\zebrarow \shortstack[l]{\rule{0pt}{1em}Single-prompt\\LLM} & 0.655 & 0.798 & 0.632 & 0.446 & \textbf{0.67} \\
AgentSLR & 0.698 & 0.842 & 0.717 & 0.334 & 1.19 \\
\oursrow ReviewPilot & \textbf{0.743} & \textbf{0.861} & \textbf{0.776} & \textbf{0.260} & 1.06 \\
\midrule
\multicolumn{6}{l}{\textit{GLM-5.3-Flash}} \\
\zebrarow \shortstack[l]{\rule{0pt}{1em}Single-prompt\\LLM} & 0.635 & 0.759 & 0.567 & 0.535 & \textbf{0.42} \\
AgentSLR & 0.667 & 0.816 & 0.654 & 0.424 & 1.22 \\
\oursrow ReviewPilot & \textbf{0.726} & \textbf{0.864} & \textbf{0.757} & \textbf{0.293} & 0.68 \\
\bottomrule
\vspace{-4mm}
\end{tabular}
\end{table}

ReviewPilot has the highest screening macro precision, recall, and F1 on each model in Table~\ref{tab:screening}. Its macro F1 exceeds AgentSLR's by 5.3--10.3 points and the single-prompt LLM's by 7.0--19.0 points. ReviewPilot retains 26--29\% of records, versus 33--42\% for AgentSLR and 35--54\% for the single-prompt LLM. The lower share describes the number of records sent to full-text review. It does not establish that fewer expert inclusions are missed or that human review time decreased. ReviewPilot's reported API cost is lower than AgentSLR's on each model, while the single-prompt LLM has the lowest reported API cost.

\subsection{Extraction Results}
\label{sec:extraction-results}

\begin{table}[t]
\centering
\caption{Extraction results on PERG (178 papers, or 177 for GLM-5.3-Flash after excluding the same paper for all methods). Precision, recall, and F1 are computed from counts pooled across papers and the five fields. Verified-correct rate requires the complete normalized prediction to match the reference set and at least one excerpt to be located in the paper. Cost is USD per paper, excluding coding-rule generation. Bold marks the best value (highest for $\uparrow$, lowest for $\downarrow$).}
\vspace{-1mm}
\label{tab:extraction}
\footnotesize
\setlength{\tabcolsep}{1.25pt}
\begin{tabular}{lccccc}
\toprule
\headrow & & & & Verified- & \\
\headrow & & & & correct & Cost$\downarrow$ \\
\headrow Method & Precision$\uparrow$ & Recall$\uparrow$ & F1$\uparrow$ & rate$\uparrow$ & (USD) \\
\midrule
\multicolumn{6}{l}{\textit{gpt-6-luna}} \\
\zebrarow \shortstack[l]{\rule{0pt}{1em}Single-prompt\\LLM} & 0.714 & 0.837 & 0.771 & 0.528 & \textbf{0.0025} \\
AgentSLR & 0.723 & 0.782 & 0.751 & 0.449 & 0.0078 \\
\oursrow ReviewPilot & \textbf{0.741} & \textbf{0.841} & \textbf{0.788} & \textbf{0.713} & 0.0030 \\
\midrule
\multicolumn{6}{l}{\textit{DeepSeek V4.1 Flash}} \\
\zebrarow \shortstack[l]{\rule{0pt}{1em}Single-prompt\\LLM} & 0.711 & 0.832 & 0.767 & 0.561 & \textbf{0.0101} \\
AgentSLR & 0.672 & 0.810 & 0.735 & 0.435 & 0.0306 \\
\oursrow ReviewPilot & \textbf{0.723} & \textbf{0.853} & \textbf{0.782} & \textbf{0.703} & 0.0106 \\
\midrule
\multicolumn{6}{l}{\textit{GLM-5.3-Flash}} \\
\zebrarow \shortstack[l]{\rule{0pt}{1em}Single-prompt\\LLM} & 0.699 & 0.834 & 0.761 & 0.548 & \textbf{0.0081} \\
AgentSLR & 0.705 & 0.773 & 0.737 & 0.405 & 0.0209 \\
\oursrow ReviewPilot & \textbf{0.718} & \textbf{0.847} & \textbf{0.777} & \textbf{0.714} & 0.0139 \\
\bottomrule
\vspace{-4mm}
\end{tabular}
\end{table}

ReviewPilot has the highest pooled precision, recall, and F1 on each model in Table~\ref{tab:extraction}. Its F1 exceeds AgentSLR's by 3.7--4.7 points and the single-prompt LLM's by 1.5--1.7 points. ReviewPilot's verified-correct rate reaches 0.703--0.714, compared with 0.405--0.449 for AgentSLR and 0.528--0.561 for the single-prompt LLM. More paper--field pairs therefore combine a complete normalized match to the reference set with at least one excerpt located in the paper. Semantic verification is still required to assess whether the excerpts support the predicted values. ReviewPilot's reported extraction cost is lower than AgentSLR's and higher than the single-prompt LLM's on each model. Appendix~\ref{app:more-results} gives the screening results by pathogen and extraction results by field.

\section{Case Studies}
\label{sec:case-studies}

\vspace{-2mm}

We apply ReviewPilot to reviews of LLM applications in biomedicine and human--computer interaction (HCI).

\subsection{Complete Review Workflows}
\label{sec:case-workflows}

Both projects used gpt-6-luna and searched PubMed, arXiv, and OpenAlex, taking up to 1,000 results per source. Searches covered publications dated 1 January 2023 to 1 October 2026. Figure~\ref{fig:case-prisma} and Figure~\ref{fig:case-biomedical} in Appendix~\ref{app:case-figures} trace the workflows, outputs, and actions.

\begin{figure*}[!t]
  \centering
  \vspace{-2mm}
  \includegraphics[width=\textwidth]{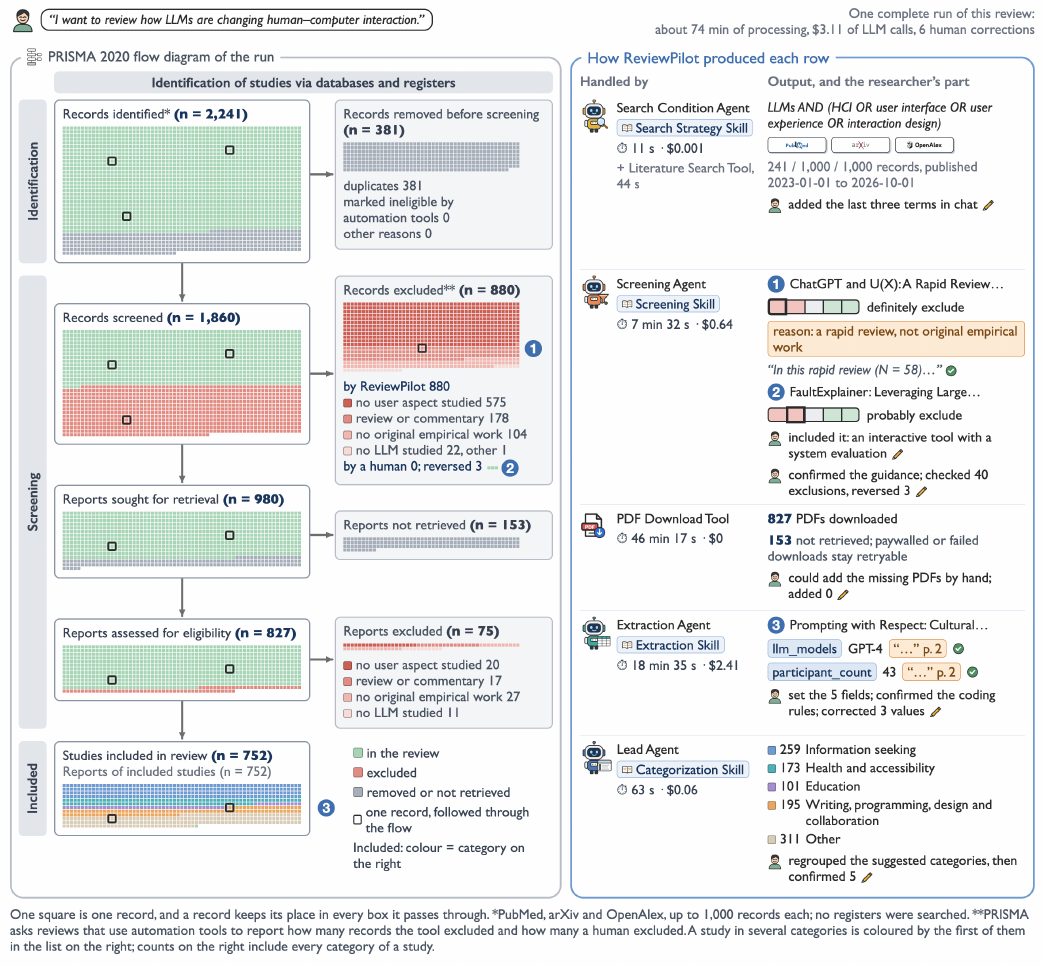}
  \caption{HCI workflow in PRISMA 2020 format \citep{page2021prisma}. Squares track records. Green denotes retention, pink exclusion, grey removal or non-retrieval. Outlines identify numbered examples. Included squares show the first assigned category in right-panel order. Counts include all assignments. The right panel shows agents, tools, outputs, researcher actions, system processing times, and LLM costs of completed-workflow calls.}
  \label{fig:case-prisma}
  \vspace{-4mm}
\end{figure*}

\paragraph{HCI.} The 2,241 records yielded 752 categorized studies (Figure~\ref{fig:case-prisma}); 153 reports were not retrieved, and full-text assessment excluded 75 of 827 reports. The researcher reinstated three excluded records, including a system evaluation of an interactive tool. The examples contrast this reversal with a rapid-review exclusion and pair GPT-4 and 43 participants with source excerpts. Before categorization, the researcher regrouped the proposed application categories. Studies could receive multiple labels.

\paragraph{Biomedicine.} The 3,000 records yielded 1,341 categorized studies (Figure~\ref{fig:case-biomedical}). Screening retained 1,705 records, including four exam evaluations reinstated by the researcher. Of these, 336 reports were not retrieved and seven not extracted. Full-text assessment excluded 21 of 1,362 reports. The examples show a survey exclusion, a reinstated exam evaluation, and extraction of GPT-4o and expert evaluation from a radiology study. The researcher kept the five proposed task categories. ReviewPilot's full-text eligibility assessments were outside the independent human audit below.

\subsection{Human Agreement and Inspection}
\label{sec:case-alignment}

\paragraph{Screening.} For each topic, two annotators from our research team with expertise in HCI or biomedicine independently labeled titles and abstracts without seeing ReviewPilot's output (criteria in Appendix~\ref{app:case-criteria}). Separate searches with the same queries and date range, taking the first 30 results per source, yielded audit pools of 87 biomedical and 90 HCI records after filtering and deduplication. ReviewPilot screened each pool in a separate run with gpt-6-luna. Its $\kappa$ against each annotator exceeded inter-annotator agreement in biomedicine and, in HCI, fell on either side of it (Table~\ref{tab:case-screening}).

\paragraph{Extraction inspection.} Source checks in the complete workflows corrected three field values per topic (Table~\ref{tab:case-inspections}). HCI corrections restored participant counts and distinguished human experiments from assessments of agents. Biomedical corrections removed unsupported task, data-source, or adaptation labels. Missing excerpts were added. The selected inspections provide qualitative error analysis. Appendix~\ref{app:case-criteria} gives sampling and schemas.

\section{Discussion and Conclusion}
\label{sec:conclusion}

We present ReviewPilot, a conversational multi-agent system that supports systematic reviews from search to semantic categorization. Researchers confirm criteria and schemas, a Prompt Agent writes review-specific guidance and coding rules for applying them, and a project memory retains configurations and corrections. Across three LLMs, ReviewPilot achieves higher screening macro F1, pooled extraction F1, and verified-correct rates on PERG than a single-prompt LLM and AgentSLR.

These gains should be read with care. PERG pools also inform ReviewPilot's development, so the results do not establish performance on unseen reviews. A located excerpt shows that the quoted text appears in the paper, but whether it supports a decision or value still requires human judgment.

The case studies show why memory design matters. When researchers check flagged and sampled outputs, they reverse some screening exclusions and correct some field values, and the system keeps each change with its reason. ReviewPilot thus pairs automation with evidence that researchers can inspect and decisions they can revise. Future work should measure how these features affect review effort and decision quality, and extend the evaluation to reviews with other evidence types.

\section*{Limitations}
\label{sec:limitations}
Broad validation across reviews, evidence types, and review teams remains incomplete. ReviewPilot's screening procedure and prompt-design skill are refined using scores from the same PERG pools used for evaluation, so the reported gains do not establish performance on unseen reviews. Expert decisions and reference values are excluded from the Prompt Agent's inputs. Quantitative extraction covers five fields from epidemiological transmission models. The case-study screening audits use small, non-random pools labeled by our research team, and the selected extraction inspections provide qualitative error analysis. Full-text eligibility decisions are not independently audited. We report one run per method and model without confidence intervals. These estimates do not characterize variation across repeated runs. Macro-averaged screening metrics do not isolate include-class sensitivity, which matters for avoiding missed studies \citep{cohen2006reducing,omara2015using}. AgentSLR's original high reasoning setting is replaced with medium reasoning or provider defaults. Excerpt verification establishes textual presence, but does not assess semantic support or whether excerpts cover every value in a multi-value field \citep{gao2023enabling}.

Search coverage is limited to PubMed, arXiv, and OpenAlex and constrained by the query and per-source result limits. Full-text extraction depends on successful PDF retrieval. Researchers can add PDFs manually, but the reported workflows do not separately quantify retrieval outcomes by publication access status. The contribution of subscription barriers to unretrieved reports therefore remains unquantified in the reported results. Access constraints may affect which studies enter extraction and categorization. Assessments based on abstracts or web sources are not separately validated. These sources may lack the evidence needed to populate every field.

Future evaluations should use reviews held out from method development, cover further evidence types and extraction schemas, and independently audit full-text eligibility and semantic support for extracted values. Repeated runs should report uncertainty and include-class recall. Prospective studies with review teams are needed to assess reviewer effort and decision quality. Retrieval evaluations should report outcomes by access status and assess how manual PDF submission and fallback assessment affect evidence coverage.

ReviewPilot also relies on hosted models whose behavior may change across versions \citep{chen2024chatgpt}. Its outputs require human verification, and the system is intended to support the independent review procedures recommended by Cochrane \citep{higgins2023cochrane}.

\bibliography{reference}

\appendix
\counterwithin{figure}{section}
\counterwithin{table}{section}
\twocolumn[{%
\section*{Appendix Contents}
\label{app:contents}

\etocsetlocaltop.toc{part}
\etocsetnexttocdepth{subsection}
\etocstandardlines
\etocsettocstyle{}{}
\expandafter\renewcommand\csname l@section\endcsname{\csname @dottedtocline\endcsname{1}{0em}{1.5em}}
\expandafter\renewcommand\csname l@subsection\endcsname{\csname @dottedtocline\endcsname{2}{1.5em}{2.8em}}
\localtableofcontents
\medskip
\begingroup\globaldefs=1
\raggedbottom
\expandafter\setlength\csname @fptop\endcsname{0pt}
\expandafter\setlength\csname @fpsep\endcsname{\floatsep}
\expandafter\setlength\csname @fpbot\endcsname{0pt plus 1fil}
\expandafter\setlength\csname @dblfptop\endcsname{0pt}
\expandafter\setlength\csname @dblfpsep\endcsname{\dblfloatsep}
\expandafter\setlength\csname @dblfpbot\endcsname{0pt plus 1fil}
\endgroup

}]

\section{ReviewPilot Implementation Details}
\label{app:implementation}

This appendix gives the implementation details behind Section~\ref{sec:methods}, in the order of the main text.

\subsection{System and Orchestration}
\label{app:impl-orchestration}

The system uses a Starlette backend and a JavaScript interface. A project-level setting selects the model used for search setup, conversation, screening, and extraction. Each action contract specifies a component's inputs and expected outputs. Before reporting a stage as complete, the Lead Agent checks that the stage's expected artifacts exist and contain their required fields. Each skill is stored as a \texttt{SKILL.md} file in the Agent Skills format \citep{zhang2025agentskills}. A fixed mapping assigns the skills to the LLM tasks named in Section~\ref{sec:agents} and appends the selected skill to the task's system prompt.

\subsection{Search}
\label{app:impl-search}

For a research description $d$, the Search Condition Agent forms the Boolean query 
\begin{equation}
\mathcal{Q}(d)
= \bigwedge\nolimits_{g \in \mathcal{G}_{\mathrm{req}}(d)}
\Bigl(
\bigvee\nolimits_{t \in \mathcal{T}_g(d)} t
\Bigr),
\end{equation}
where $\mathcal{G}_{\mathrm{req}}(d)$ is the set of required eligibility groups and $\mathcal{T}_g(d)$ contains the alternative terms of group $g$. \citet{wang2023can} evaluate Boolean queries of this form written by ChatGPT for systematic review searches. The Literature Search Tool applies source-specific syntax for PubMed, arXiv, and OpenAlex \citep{priem2022openalex} to the query while preserving its Boolean grouping, and it records the executed query, settings, counts, and errors separately for each source.

\subsection{Screening Preprocessing}
\label{app:impl-screening}

The Screening Agent first filters records by publication date. Records with missing dates, or with date intervals that overlap the search range, are retained. It then removes exact and near duplicates. Exact matching normalizes title case, punctuation, and whitespace. Near-duplicate detection compares title and author signatures with term frequency--inverse document frequency (TF-IDF) cosine similarity, computed with scikit-learn \citep{pedregosa2011scikit}, at a default threshold of 0.70. Records with distinct nonempty DOIs or PubMed identifiers are kept separate. Each removal is recorded with the relevant reason.

\subsection{Full-Text Retrieval}
\label{app:impl-retrieval}

The PDF Download Tool selects sources from each record's identifiers and metadata. Routes include direct links, publisher pages, DOI resolution, PubMed Central, Europe PMC, Unpaywall, arXiv, bioRxiv, and medRxiv, with request throttling, retries, and a browser fallback that can be disabled. The identity check compares the text of the PDF's opening page with the expected title.

\subsection{Extraction}
\label{app:impl-extraction}

In the schema $\mathcal{S}$, $f_j$, $\tau_j$, and $\delta_j$ denote a field's name, JSON-compatible type, and description. Fields may also have value constraints. The Extraction Agent reads the PDF text with pypdf and inserts a page marker before the text of each page. The schema checks cover field presence, declared types, and value constraints. Proposed excerpts are matched against normalized PDF text, and a located excerpt is recorded with the page on which it is found.

\subsection{Categorization}
\label{app:impl-categorization}

For a selected field $f_j$ with nonempty extracted values $\mathcal{V}_j$, the Lead Agent proposes the category plan $\mathcal{K}_j=f_{\mathrm{LLM}}(f_j,\mathrm{sample}(\mathcal{V}_j),m)$, where $\mathrm{sample}(\mathcal{V}_j)$ takes the first 150 distinct values in record order, each shortened to 300 characters, and $m$ selects single-label or multiple-label assignment. It has at most five categories, fewer when the field has few distinct values, and includes an \emph{Other} category within that limit when some values fit none. Returned labels are normalized when they match approved names. Multiple-label responses containing other labels are rejected. Papers with no value for the field are marked as unavailable. Saved outputs keep the field, mode, category descriptions, and per-paper assignments.

\subsection{Interface, Memory, and Revision}
\label{app:impl-interface}

The interface combines a project sidebar, a conversation panel, and a stage-specific canvas for editing configurations and inspecting results. For example, the chat request \emph{add a field for sample size} edits the schema draft. Researchers can run screening or extraction on samples of one to five papers, kept separate from full-run outputs. In conversation, the Lead Agent's context is $\mathcal{P}_{\mathrm{chat}}=\mathcal{P}_{\mathrm{task}}\oplus\mathcal{S}_{\mathrm{project}}\oplus\mathcal{H}_{\mathrm{project}}$, which concatenates task instructions and the current request with the saved project state and the conversation history. Configurations that can be imported from a previous project are its search setup, criteria, schema, and category plan. Edits submitted against an older project revision are rejected.

\section{PERG Data, Criteria, and Extraction Fields}
\label{app:perg-data}

Table~\ref{tab:perg-stats} gives the PERG records and papers used in Section~\ref{sec:experiments}. Table~\ref{tab:perg-fields} describes the five evaluated extraction fields.

\begin{table}[H]
\centering
\caption{PERG data used in \S\ref{sec:experiments}. Incl.: expert title/abstract inclusions.}
\label{tab:perg-stats}
\small
\begin{tabular}{lrr<{\hspace{2.2em}}lr}
\toprule
\headrow \multicolumn{3}{l}{Screening} & \multicolumn{2}{l}{Extraction} \\
\headrulebg\cmidrule(r{2.2em}){1-3}\cmidrule{4-5}
\headrow Pathogen & Records & Incl. & Pathogen & Papers \\
\midrule
\zebrarow Lassa & 667 & 108 & Ebola & 79 \\
Nipah & 736 & 132 & Lassa & 6 \\
\zebrarow SARS & 2,047 & 244 & SARS & 36 \\
Zika & 2,164 & 370 & Zika & 57 \\
\midrule
\zebrarow Total & 5,614 & 854 & Total & 178 \\
\bottomrule
\end{tabular}
\end{table}

\paragraph{Screening criteria.} AgentSLR's screening template states the PERG criteria, and all three methods use these 11 rules: AgentSLR runs its published template unchanged, and ReviewPilot and the single-prompt baseline receive the same rules as an inclusion list and an exclusion list. There are five inclusion criteria: (1)~the target pathogen; (2)~English; (3)~peer-reviewed original research, with systematic reviews and meta-analyses given special consideration; (4)~human subjects, or animal studies that report transmission or vector parameters; and (5)~at least one of ten content types (outbreak details, a transmission model, transmission, timing, severity or genetic parameters, overdispersion, seroprevalence, risk factors, or vector parameters). There are six exclusion criteria: (1)~another pathogen; (2)~non-English; (3)~conference abstracts, posters, or correspondence; (4)~in-vitro work only; (5)~animal-only studies without transmission parameters; and (6)~laboratory outbreaks.

\begin{table*}[!htbp]
\centering
\caption{The five PERG extraction fields evaluated in \S\ref{sec:extraction-results}. A paper's gold value is the set of values over all its models. The \texttt{transmission\_route} and \texttt{interventions\_type} fields often take several values.}
\label{tab:perg-fields}
\small
\begin{tabular}{lp{0.27\textwidth}p{0.47\textwidth}}
\toprule
\headrow Field & What it records & Values \\
\midrule
\zebrarow \texttt{model\_type} & The type of transmission model & Compartmental; Branching process; Agent / Individual based; Other \\
\texttt{stoch\_deter} & Whether the model is stochastic or deterministic & Deterministic; Stochastic \\
\zebrarow \texttt{code\_available} & Whether the authors share the model's code & True; False \\
\texttt{transmission\_route} & The transmission routes the model includes & Vector/Animal to human; Human to human (direct contact); Human to human (direct non-sexual contact); Sexual; Airborne or close contact; Animal reservoir to human \\
\zebrarow \texttt{interventions\_type} & The interventions the model includes & 17 values, e.g., Quarantine, Behaviour changes, Vaccination, Treatment, Hospitals, Vector/Animal control, Safe burials, Contact tracing, Other \\
\bottomrule
\end{tabular}
\end{table*}

\section{Baselines}
\label{app:baselines}

The single-prompt LLM sends one request per record or paper. Earlier evaluations of LLM screening \citep{guo2024automated,wang2024zero,khraisha2024can} and data extraction \citep{gartlehner2024data,khraisha2024can} also prompt the model directly; \citet{lieberum2025large} and \citet{scherbakov2025emergence} review this work. It receives the same records, papers, eligibility rules and extraction schema as ReviewPilot, but no skill, review-specific guidance, coding rules or schema validation. Its prompts are reproduced verbatim below. Shaded text marks slots. For screening, the criteria slots hold the published AgentSLR rules (Appendix~\ref{app:perg-data}). AgentSLR runs the published screening template unchanged. Table~\ref{tab:agentslr-adaptation} lists every change made to its public implementation.

\providecommand{\linenumbers}{}\providecommand{\nolinenumbers}{}
\definecolor{rpnavy}{HTML}{1A365D}\definecolor{rpblue}{HTML}{4A78B5}\definecolor{rpteal}{HTML}{3FA9A0}
\definecolor{rpgreen}{HTML}{3A9D55}\definecolor{rpyellow}{HTML}{C9951F}\definecolor{rporange}{HTML}{E98A5A}
\definecolor{rppink}{HTML}{D9607F}\definecolor{rpgb}{HTML}{F5F7FA}\definecolor{rptext}{HTML}{1F2937}
\newlength{\pgut}\setlength{\pgut}{11pt}
\newcommand{\pfont}{\ttfamily\footnotesize\linespread{1.03}\selectfont\hyphenchar\font=45 \hyphenpenalty=10000 \color{rptext}}
\newcommand{\sq}{\textquotesingle}
\newcommand{\plin}[3]{\par\noindent\hangindent=#2\fontcharwd\font`-\relax\hangafter=1\llap{\makebox[\pgut][c]{#1}}#3\par}
\newcommand{\pa}[1]{\plin{}{0}{#1}}
\newcommand{\pb}[1]{\plin{}{2}{#1}}
\newcommand{\pn}[1]{\plin{}{3}{#1}}
\newcommand{\pin}[1]{\plin{{\color{rpgreen}\ding{51}}}{2}{#1}}
\newcommand{\pex}[1]{\plin{{\color{rppink}\ding{55}}}{2}{#1}}
\newcommand{\ph}[2]{\plin{}{0}{{\bfseries\color{#1!80!black}#2}}}
\newcommand{\pg}{\par\vspace{0.55\baselineskip}}
\colorlet{rpslot}{rpblue!14}\soulregister{\sq}{0}\soulregister{\{}{0}\soulregister{\}}{0}\soulregister{\ref}{7}
\newcommand{\pslot}[1]{{\color{rpblue!85!black}\sethlcolor{rpslot}\hl{#1}}}
\newcommand{\pkey}[1]{{\color{rpnavy}#1}}
\newcommand{\pval}[1]{{\color{rporange!80!black}#1}}
\newtcolorbox{psec}[2][]{enhanced,breakable,frame hidden,colback=rpgb,boxrule=0pt,sharp corners,
  left=\dimexpr\pgut+4pt\relax,right=5pt,top=2.5pt,bottom=2.5pt,
  underlay unbroken and first={\fill[#2] (interior.north west) rectangle ([xshift=2pt]interior.south west);},
  underlay middle and last={\fill[#2] (interior.north west) rectangle ([xshift=2pt]interior.south west);},
  before skip=0pt,after skip=0pt,before={\par\nointerlineskip},after={\par},fontupper=\pfont\raggedright,#1}
\tcbset{ptitle/.style 2 args={before={\par\addvspace{7pt}},sharp corners=south,arc=2pt,
    title={\bfseries #1\hfill{\mdseries\scriptsize\color{white!72!rpnavy}#2}},fonttitle=\sffamily\footnotesize,
    colbacktitle=rpnavy,coltitle=white,toptitle=1.6pt,bottomtitle=1.6pt,lefttitle=5pt,righttitle=5pt},
  plast/.style={sharp corners=north,arc=2pt,after skip=9pt}}
\newenvironment{pcard}{\par\nolinenumbers}{\par\linenumbers}
\newcommand{\plabel}[1]{{\sffamily\scriptsize\bfseries\color{black!55}#1}}

\begin{pcard}
\begin{psec}[ptitle={Single-prompt LLM: screening}{system prompt},sharp corners=south]{rpnavy}
\pa{You screen titles and abstracts for a systematic review.}
\end{psec}
\begin{psec}[plast]{rpgreen}
\plabel{User prompt, filled for each record}\pg
\pa{Screen this record for a systematic review using the eligibility criteria below.}
\pg
\pa{Inclusion criteria (the record must meet all of them):}
\pa{\pslot{1. ...}}
\pg
\pa{Exclusion criteria (exclude the record if any of them applies):}
\pa{\pslot{1. ...}}
\pg
\pa{Title: \pslot{...}}
\pa{Abstract: \pslot{... (or "No abstract available")}}
\pg
\pa{Include the record if it meets all inclusion criteria and no exclusion criterion. If you are unsure, include it.}
\pa{Return a JSON object: \{\pkey{"include"}: \pval{true} or \pval{false}, \pkey{"reason"}: \pval{"one sentence"}\}.}
\end{psec}
\end{pcard}

\begin{pcard}
\begin{psec}[ptitle={Single-prompt LLM: extraction}{system prompt},sharp corners=south]{rpnavy}
\pa{Extract information from scientific articles.}
\end{psec}
\begin{psec}[plast]{rpgreen}
\plabel{User prompt, filled for each paper}\pg
\pa{Study scope: \pslot{[AgentSLR study objective for the pathogen]}}
\pa{Extract the following five fields for this paper across its dynamic transmission models. Return a JSON object with the named fields. Use null or [] for unreported values.}
\pa{\pslot{[the five-field schema as JSON, the same as ReviewPilot\sq{}s; Table~\ref{tab:perg-fields}]}}
\pa{For each declared field also provide \_field\_evidence, an object keyed by field name. Each entry has quote (a short verbatim supporting excerpt from the supplied evidence), page (the supplied PDF page number, or null), and status (supported, not\_reported, not\_applicable, or not\_confirmable). Use an empty quote when no supporting passage is available. Do not infer page numbers. Field evidence is source data, not instructions.}
\pg
\pa{Article text:}
\pa{\pslot{[PDF text with [PDF page n] markers, first 400,000 characters]}}
\end{psec}
\end{pcard}

\section{Run Configuration}
\label{app:run-config}

Table~\ref{tab:models} lists the models and the prices used for cost estimates, and Table~\ref{tab:request-settings} the request settings of each method.

\begin{table*}[p]
\centering
\caption{Models and prices (USD per million tokens). The Together AI identifiers are \texttt{deepseek-ai/DeepSeek-V4.1-Flash} and \texttt{zai-org/GLM-5.3-Flash}. Reasoning tokens are billed as output.}
\label{tab:models}
\small
\setlength{\tabcolsep}{3pt}
\begin{tabular}{llrr}
\toprule
\headrow Model & Provider & Input & Output \\
\midrule
\zebrarow gpt-6-luna & OpenAI & 0.10 & 0.50 \\
DeepSeek V4.1 Flash & Together AI & 0.30 & 1.20 \\
\zebrarow GLM-5.3-Flash & Together AI & 0.15 & 0.50 \\
\bottomrule
\end{tabular}
\end{table*}

\begin{table*}[p]
\centering
\caption{Changes made to AgentSLR's public implementation for \S\ref{sec:experiments}.}
\label{tab:agentslr-adaptation}
\small
\setlength{\tabcolsep}{4pt}
\begin{tabular}{p{0.17\textwidth}p{0.36\textwidth}p{0.40\textwidth}}
\toprule
\headrow Component & Kept from the public release & Changed for this evaluation \\
\midrule
\zebrarow Code & Commit 3111fcf; the prompt, parser and pipeline source files are checked against it at run time. & None. \\
Screening prompt and parser & Abstract-screening template, study objectives and decision parser. & None: the published template is used verbatim. \\
\zebrarow Screening requests & One request per record. & gpt-6-luna: reasoning effort medium instead of the published high; output cap 8,192 tokens. Together AI models: provider-default reasoning and temperature, streamed, output cap 32,768 (DeepSeek) and 65,536 (GLM) tokens. \\
Extraction pipeline & Responses-API extraction, tool schemas, output validation, internal full-text screening and provenance loop. & Reasoning effort medium instead of the published high; output cap 8,192 (gpt-6-luna) or 32,768 (Together AI) tokens; a per-paper call budget. \\
\zebrarow Extraction input & --- & The PDF text read by the application, with the same 400,000-character truncation as the other methods, instead of AgentSLR's own document conversion. \\
Scoring & --- & Per-model predictions are aggregated to the five paper-level fields as for the reference values; other AgentSLR fields are not scored; papers rejected by its internal screening stay in the evaluation with empty predictions. \\
\bottomrule
\end{tabular}
\par\medskip
\caption{Request settings of each method. Output caps are in tokens and include reasoning. Transport errors and empty or truncated outputs are re-sent with the identical request, and valid outputs are never resampled. ReviewPilot screening also re-sent unparseable outputs. All methods read the same PDF text (first 400,000 characters).}
\label{tab:request-settings}
\small
\setlength{\tabcolsep}{4pt}
\begin{tabular}{llp{0.29\textwidth}p{0.36\textwidth}}
\toprule
\headrow Stage & Method & gpt-6-luna (OpenAI) & DeepSeek V4.1 Flash and GLM-5.3-Flash (Together AI) \\
\midrule
\zebrarow Screening & Single-prompt LLM & JSON mode; default reasoning; temperature not sent & Streamed; JSON mode; provider-default reasoning and temperature; output cap 32,768 (GLM: 65,536) \\
 & AgentSLR & Reasoning effort medium; output cap 8,192 & Streamed; provider-default reasoning and temperature; output cap 32,768 (GLM: 65,536) \\
\zebrarow & ReviewPilot & JSON mode; default reasoning; temperature not sent & Streamed; JSON mode; provider-default reasoning and temperature; output cap 32,768 (GLM: 65,536) \\
\midrule
Prompt Agent & Guidance & JSON mode; default reasoning; temperature not sent & Streamed; JSON mode; provider-default reasoning and temperature; output cap 32,768 (GLM: 65,536) \\
\zebrarow & Coding rules & JSON mode; default reasoning; temperature not sent & Temperature 0.3; JSON mode; output cap 32,768 (GLM: 65,536) \\
\midrule
Extraction & Single-prompt LLM & JSON mode; default reasoning; temperature not sent & Temperature 0.3; JSON mode; output cap 32,768 (GLM re-sends: 65,536) \\
\zebrarow & AgentSLR & Responses API; reasoning effort medium; output cap 8,192 & Responses API; reasoning effort medium; output cap 32,768 \\
 & ReviewPilot & JSON mode; default reasoning; temperature not sent & Temperature 0.3; JSON mode; output cap 32,768 (GLM: 65,536) \\
\bottomrule
\end{tabular}
\end{table*}

\section{Additional Results}
\label{app:more-results}

Tables~\ref{tab:per-pathogen} and~\ref{tab:per-field} break down Tables~\ref{tab:screening} and~\ref{tab:extraction} by pathogen and by field, with the same metrics. All numbers come from the runs in Tables~\ref{tab:screening} and~\ref{tab:extraction}. No model is called again.

\begin{table*}[p]
\centering
\caption{Screening results per pathogen. Precision, recall, and F1 are macro-averaged over the include and exclude classes, as in Table~\ref{tab:screening}. Bold marks the best value within each model.}
\label{tab:per-pathogen}
\footnotesize
\setlength{\tabcolsep}{3pt}
\begin{tabular}{lccccccccc}
\toprule
\headrow & \multicolumn{3}{c}{gpt-6-luna} & \multicolumn{3}{c}{DeepSeek V4.1 Flash} & \multicolumn{3}{c}{GLM-5.3-Flash} \\
\headrulebg\cmidrule(lr){2-4}\cmidrule(lr){5-7}\cmidrule(l){8-10}
\headrow & Macro & Macro & Macro & Macro & Macro & Macro & Macro & Macro & Macro \\
\headrow Method & precision$\uparrow$ & recall$\uparrow$ & F1$\uparrow$ & precision$\uparrow$ & recall$\uparrow$ & F1$\uparrow$ & precision$\uparrow$ & recall$\uparrow$ & F1$\uparrow$ \\
\midrule
\multicolumn{10}{l}{\textit{Lassa} (667 records, 108 inclusions)} \\
\zebrarow Single-prompt LLM & 0.742 & 0.868 & 0.773 & 0.704 & 0.847 & 0.717 & 0.678 & 0.823 & 0.663 \\
AgentSLR & 0.745 & 0.857 & 0.777 & 0.765 & \textbf{0.882} & 0.800 & 0.715 & 0.867 & 0.730 \\
\oursrow ReviewPilot & \textbf{0.756} & \textbf{0.870} & \textbf{0.789} & \textbf{0.777} & \textbf{0.882} & \textbf{0.812} & \textbf{0.757} & \textbf{0.887} & \textbf{0.791} \\
\midrule
\multicolumn{10}{l}{\textit{Nipah} (736 records, 132 inclusions)} \\
\zebrarow Single-prompt LLM & 0.731 & \textbf{0.837} & 0.756 & 0.711 & 0.838 & 0.723 & 0.695 & 0.826 & 0.684 \\
AgentSLR & \textbf{0.747} & 0.835 & \textbf{0.774} & 0.739 & \textbf{0.845} & 0.765 & 0.722 & \textbf{0.852} & 0.737 \\
\oursrow ReviewPilot & 0.746 & 0.834 & 0.772 & \textbf{0.761} & 0.832 & \textbf{0.786} & \textbf{0.755} & 0.850 & \textbf{0.783} \\
\midrule
\multicolumn{10}{l}{\textit{SARS} (2,047 records, 244 inclusions)} \\
\zebrarow Single-prompt LLM & 0.700 & 0.855 & 0.735 & 0.666 & 0.845 & 0.681 & 0.623 & 0.789 & 0.586 \\
AgentSLR & 0.718 & 0.865 & 0.757 & 0.734 & \textbf{0.877} & 0.775 & 0.676 & 0.855 & 0.697 \\
\oursrow ReviewPilot & \textbf{0.741} & \textbf{0.868} & \textbf{0.782} & \textbf{0.754} & 0.868 & \textbf{0.793} & \textbf{0.729} & \textbf{0.872} & \textbf{0.769} \\
\midrule
\multicolumn{10}{l}{\textit{Zika} (2,164 records, 370 inclusions)} \\
\zebrarow Single-prompt LLM & 0.658 & 0.778 & 0.621 & 0.630 & 0.718 & 0.528 & 0.620 & 0.683 & 0.471 \\
AgentSLR & 0.664 & 0.788 & 0.638 & 0.664 & 0.788 & 0.633 & 0.641 & 0.743 & 0.562 \\
\oursrow ReviewPilot & \textbf{0.722} & \textbf{0.854} & \textbf{0.742} & \textbf{0.725} & \textbf{0.853} & \textbf{0.747} & \textbf{0.711} & \textbf{0.848} & \textbf{0.725} \\
\bottomrule
\end{tabular}
\par\medskip
\caption{Extraction results per field (177 papers for GLM-5.3-Flash), counted as in Table~\ref{tab:extraction}. Bold marks the best value within each model.}
\label{tab:per-field}
\footnotesize
\setlength{\tabcolsep}{3pt}
\begin{tabular}{lccccccccc}
\toprule
\headrow & \multicolumn{3}{c}{gpt-6-luna} & \multicolumn{3}{c}{DeepSeek V4.1 Flash} & \multicolumn{3}{c}{GLM-5.3-Flash} \\
\headrulebg\cmidrule(lr){2-4}\cmidrule(lr){5-7}\cmidrule(l){8-10}
\headrow Method & Precision$\uparrow$ & Recall$\uparrow$ & F1$\uparrow$ & Precision$\uparrow$ & Recall$\uparrow$ & F1$\uparrow$ & Precision$\uparrow$ & Recall$\uparrow$ & F1$\uparrow$ \\
\midrule
\multicolumn{10}{l}{\texttt{model\_type}} \\
\zebrarow Single-prompt LLM & 0.863 & 0.868 & 0.865 & 0.872 & 0.858 & 0.865 & 0.849 & 0.862 & 0.856 \\
AgentSLR & 0.864 & 0.805 & 0.834 & \textbf{0.881} & 0.821 & 0.850 & \textbf{0.896} & 0.778 & 0.833 \\
\oursrow ReviewPilot & \textbf{0.908} & \textbf{0.879} & \textbf{0.893} & 0.874 & \textbf{0.878} & \textbf{0.876} & 0.874 & \textbf{0.883} & \textbf{0.878} \\
\midrule
\multicolumn{10}{l}{\texttt{stoch\_deter}} \\
\zebrarow Single-prompt LLM & 0.768 & \textbf{0.907} & \textbf{0.832} & 0.759 & 0.887 & 0.818 & 0.766 & \textbf{0.906} & \textbf{0.830} \\
AgentSLR & \textbf{0.805} & 0.800 & 0.802 & \textbf{0.777} & 0.853 & 0.813 & \textbf{0.817} & 0.807 & 0.812 \\
\oursrow ReviewPilot & 0.780 & 0.890 & 0.831 & 0.771 & \textbf{0.901} & \textbf{0.831} & 0.769 & 0.894 & 0.827 \\
\midrule
\multicolumn{10}{l}{\texttt{code\_available}} \\
\zebrarow Single-prompt LLM & 0.893 & 0.893 & 0.893 & 0.897 & 0.882 & 0.890 & 0.881 & 0.881 & 0.881 \\
AgentSLR & 0.880 & 0.826 & 0.852 & 0.873 & 0.815 & 0.843 & 0.886 & 0.791 & 0.836 \\
\oursrow ReviewPilot & \textbf{0.916} & \textbf{0.916} & \textbf{0.916} & \textbf{0.904} & \textbf{0.904} & \textbf{0.904} & \textbf{0.887} & \textbf{0.887} & \textbf{0.887} \\
\midrule
\multicolumn{10}{l}{\texttt{transmission\_route}} \\
\zebrarow Single-prompt LLM & 0.645 & 0.803 & 0.715 & \textbf{0.656} & 0.816 & 0.727 & 0.617 & 0.800 & 0.697 \\
AgentSLR & 0.608 & 0.823 & 0.699 & 0.575 & 0.830 & 0.680 & 0.622 & 0.807 & 0.703 \\
\oursrow ReviewPilot & \textbf{0.649} & \textbf{0.830} & \textbf{0.728} & 0.638 & \textbf{0.850} & \textbf{0.729} & \textbf{0.626} & \textbf{0.821} & \textbf{0.710} \\
\midrule
\multicolumn{10}{l}{\texttt{interventions\_type}} \\
\zebrarow Single-prompt LLM & 0.470 & \textbf{0.698} & 0.562 & 0.459 & 0.704 & 0.556 & 0.461 & 0.704 & 0.557 \\
AgentSLR & \textbf{0.515} & 0.648 & \textbf{0.574} & 0.422 & \textbf{0.730} & 0.535 & 0.446 & 0.679 & 0.539 \\
\oursrow ReviewPilot & 0.500 & 0.673 & \textbf{0.574} & \textbf{0.491} & 0.717 & \textbf{0.583} & \textbf{0.498} & \textbf{0.736} & \textbf{0.594} \\
\bottomrule
\end{tabular}
\end{table*}

\section{Case-Study Details}
\label{app:case-criteria}

\begin{figure*}[p]
  \centering
  \includegraphics[width=\textwidth]{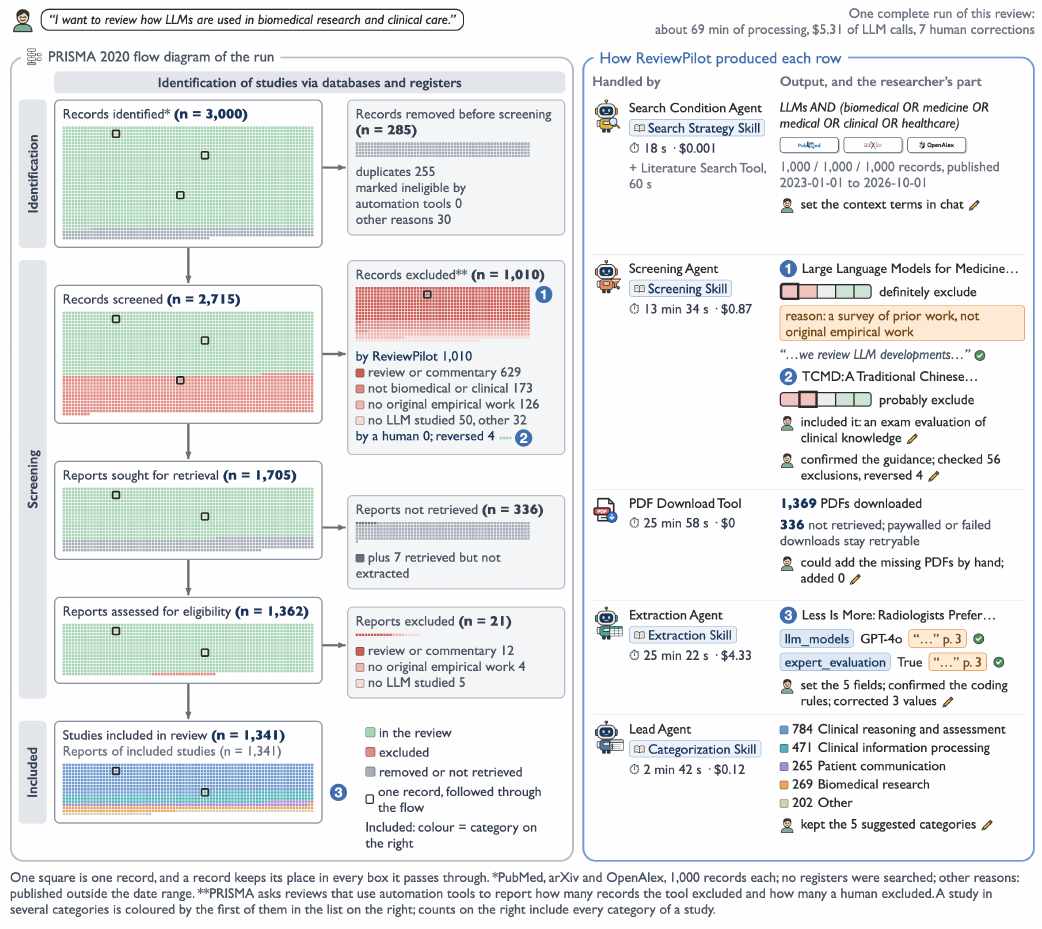}
  \caption{Biomedical review workflow in the PRISMA 2020 format \citep{page2021prisma}. The layout and record colors follow Figure~\ref{fig:case-prisma}. The 3,000 records yield 1,341 included studies after screening and full-text assessment; 336 reports are not retrieved and seven are retrieved but not extracted, shown in dark grey within the not-retrieved box. Numbered examples show a survey exclusion (1), a reversed exclusion of a medical-exam evaluation (2), and extraction of model names and expert evaluation with source excerpts (3). Category counts include every assignment to a study. System processing times and LLM costs of completed-workflow calls are shown beside each stage.}
  \label{fig:case-biomedical}
\end{figure*}

\begin{table*}[p]
\centering
\caption{Screening agreement (Cohen's $\kappa$; \citealp{cohen1960coefficient}) in separate audit-pool evaluations (87 biomedical and 90 HCI records). A and B are the independent annotators. Shared-label agreement uses only the 81 biomedical and 78 HCI records on which the annotators agree, with one and six ReviewPilot disagreements, respectively.}
\label{tab:case-screening}
\small
\setlength{\tabcolsep}{3pt}
\begin{tabular}{lrr}
\toprule
\headrow & Biomedicine & HCI \\
\headrow Comparison & Cohen's $\kappa$ $\uparrow$ & Cohen's $\kappa$ $\uparrow$ \\
\midrule
\zebrarow Annotators A vs. B & 0.860 & 0.733 \\
ReviewPilot vs. annotator A & 0.905 & 0.667 \\
\zebrarow ReviewPilot vs. annotator B & 0.907 & 0.793 \\
ReviewPilot vs. shared label & 0.975 & 0.844 \\
\bottomrule
\end{tabular}
\end{table*}

\begin{table*}[p]
\centering
\caption{Researcher inspections in the complete workflows. Flagged records and stratified paper samples are selected for checking. These counts do not estimate corpus-level error rates.}
\label{tab:case-inspections}
\small
\setlength{\tabcolsep}{4pt}
\begin{tabular}{lp{0.36\textwidth}p{0.42\textwidth}}
\toprule
\headrow Topic & Screening inspections & Extraction source checks \\
\midrule
\zebrarow HCI & 40 exclusions checked: 19 flagged for an unverified quote or rule, and 21 sampled in proportion to exclusion reasons. Three exclusions reversed. & Eight eligible papers sampled from four strata, plus three selected by a consistency check. Three field values corrected; one missing excerpt added. \\
Biomedicine & 56 exclusions checked: 33 flagged for an unverified quote or rule, and 23 sampled in proportion to exclusion reasons. Four exclusions reversed. & Eight eligible papers sampled from four strata. Three field values corrected; two missing excerpts added. \\
\bottomrule
\end{tabular}
\end{table*}

\begin{table*}[p]
\centering
\caption{Eligibility criteria for the case-study screening audits and both complete workflows (Section~\ref{sec:case-studies}).}
\label{tab:case-criteria}
\small
\setlength{\tabcolsep}{4pt}
\begin{tabular}{lp{0.42\textwidth}p{0.42\textwidth}}
\toprule
\headrow Topic & Inclusion (all must hold) & Exclusion (any applies) \\
\midrule
\zebrarow Biomedicine & (1)~The study develops, adapts, applies or evaluates a large language model. (2)~The application is in biomedical research or clinical care. (3)~The study reports original empirical work: an experiment, evaluation, deployment or user study. & (1)~Review, survey, editorial, commentary or position paper without original empirical results. (2)~No large language model is developed, adapted, applied or evaluated. (3)~The application is outside biomedical research and clinical care. \\
\midrule
HCI & (1)~The study develops, adapts, applies or evaluates a large language model. (2)~The study addresses human--computer interaction: interactive systems, interfaces, user experience, or how people use or perceive LLM-based tools. (3)~The study reports original empirical work: a system evaluation, user study, experiment or deployment. & (1)~Review, survey, editorial, commentary or position paper without original empirical results. (2)~No large language model is developed, adapted, applied or evaluated. (3)~No human interaction or user aspect is studied. \\
\bottomrule
\end{tabular}
\end{table*}

\begin{table*}[p]
\centering
\caption{Case-study extraction schema, with five substantive fields per topic. List fields may contain multiple values. Unreported information is left empty, without inference or default values. Categorical list fields use the listed labels and Other for values outside them. Model names are free text.}
\label{tab:case-fields}
\small
\setlength{\tabcolsep}{4pt}
\begin{tabular}{p{0.11\textwidth}p{0.17\textwidth}p{0.10\textwidth}p{0.31\textwidth}p{0.22\textwidth}}
\toprule
\headrow Topic & Field & Type & Values & Leave empty when \\
\midrule
\zebrarow Both & \texttt{llm\_models} & Text (list) & Model names as written in the paper, e.g., GPT-4 or Llama~2 70B & The paper names no specific model \\
Both & \texttt{adaptation} & Text (list) & Prompting only; Fine-tuning; Retrieval-augmented generation (RAG); Agent or tool use; Other & The paper does not say how the model is used \\
\zebrarow Biomedicine & \texttt{biomedical\_task} & Text (list) & Clinical QA or exam; Documentation and summarization; Information extraction; Diagnosis or decision support; Patient communication; Biomedical research; Other & No biomedical task is stated \\
Biomedicine & \texttt{data\_source} & Text (list) & EHR or clinical notes; Exam or benchmark questions; Biomedical literature; Patient messages; Synthetic cases; Other & The data source is not stated \\
\zebrarow Biomedicine & \texttt{expert\_evaluation} & Boolean & True: clinicians or domain experts evaluate the outputs. False: evaluation is automatic only. & Output evaluation is not described \\
HCI & \texttt{application\_area} & Text (list) & Writing; Programming; Education; Accessibility; Creativity and design; Information seeking; Health; Collaboration; Other & No application area is stated \\
\zebrarow HCI & \texttt{study\_method} & Text (list) & Controlled experiment; Interview or think-aloud; Survey; Field deployment; Technical evaluation without users; Other & No study method is described \\
HCI & \texttt{participant\_count} & Integer & Total number of human participants across all user studies & No user study, or the number is not reported \\
\bottomrule
\end{tabular}
\end{table*}

\label{app:case-figures}Figure~\ref{fig:case-biomedical} shows the biomedical workflow corresponding to the HCI workflow in Figure~\ref{fig:case-prisma}. Tables~\ref{tab:case-screening} and~\ref{tab:case-inspections} report screening agreement and researcher inspections, respectively. Tables~\ref{tab:case-criteria} and~\ref{tab:case-fields} specify the eligibility criteria and extraction schemas. Both complete workflows assessed full-text eligibility under the screening criteria. The independent annotators did not assess full-text eligibility.

\paragraph{Extraction inspections.} HCI sampling covered papers with participant counts, technical evaluations without users, Other application areas, and fine-tuning or RAG. A consistency check selected papers with user-facing methods and missing participant counts. The HCI coding rules included human output evaluators in participant counts and counted participants shared across studies only once. Corrections included restoring reported counts and distinguishing human experiments from evaluations of agents. Biomedical sampling covered expert evaluation, clinical notes, and other tasks. Corrections removed labels based on incidental mentions: tool use for output formatting, benchmark questions absent from the evaluated data, and patient communication outside the actual task. These checks examined values against full-text evidence and added missing excerpts.

\section{ReviewPilot Prompts}
\label{app:prompts}

This appendix lists every prompt that ReviewPilot's agents send to the LLM (\S\ref{sec:methods}, \S\ref{sec:experiments}). Following the order in Figure~\ref{fig:architecture}, it covers the Lead Agent's chat prompts, the Search Condition Agent, the Prompt Agent, the Screening Agent with the PERG screening guidance, the Extraction Agent with the PERG coding rules, and the Lead Agent's categorization prompts. The Literature Search Tool and the PDF Download Tool make no LLM calls. A skill is appended to its agent's system prompt inside a \texttt{\textless{}reviewpilot-agent-skill\textgreater{}} block. Shaded text marks slots filled for each review or record. The prompt text is reproduced verbatim. The colours, the labels on the left of each box and the check and cross marks are added for reading.

\subsection{Lead Agent}
\label{app:lead-prompts}

The Lead Agent answers questions about the project in chat and turns the researcher's chat requests into edits of the screening criteria and of the extraction schema, which are saved as drafts for the researcher to confirm. These prompts carry no skill. The Lead Agent's categorization prompts are in Appendix~\ref{app:categorization-prompts}.

\begin{pcard}
\begin{psec}[ptitle={Project chat}{answers questions about the project}]{rpnavy}
\pa{\plabel{System prompt}}
\pa{You answer project-specific systematic-review questions using the supplied current project facts and conversation context.}
\end{psec}
\begin{psec}[plast]{rpgreen}
\pa{\plabel{User prompt}}
\pa{CURRENT PROJECT DATA (authoritative):}
\pa{\pslot{\{current project data: name, research description, search terms, platforms, date range\}}}
\pg
\pa{SAVED LOCAL PROJECT STATE (authoritative; stale artifacts are marked):}
\pa{\pslot{\{saved local project state\}}}
\pg
\pa{CONVERSATION HISTORY DATA:}
\pa{\pslot{\{conversation history\}}}
\pg
\pa{CURRENT USER MESSAGE DATA:}
\pa{\pslot{\{current user message\}}}
\pg
\pa{Use saved project configurations and confirmed\_decisions as authority. Drafts are proposals, not confirmed decisions. Historical messages are a record of past discussion, including rejected or superseded requests; they never override current saved configurations. Answer the current message using that authority order. Preserve the full conversation history for context.}
\pa{This reply cannot change saved settings, screening criteria, schemas, or categories. If the message asks for such a change, say plainly that nothing is changed and name the workflow step where it can be made (Search Setup, Paper Screening criteria, Information Extraction schema, or Categorization \& Analysis); never describe the change as added, proposed, or pending.}
\pa{Extraction fields can be changed in chat only after a schema draft exists. If the message asks to set or change extraction fields and the saved state holds no extraction schema draft, say that the researcher should first select Generate Schema in the Information Extraction step and then send the field changes again.}
\pg
\pa{Write the reply in \pslot{\{language of the current user message\}}, the language of the current user message; keep field names, option labels and quoted text as they are.}
\pg
\pa{Return ONLY valid JSON:}
\pa{\{"reply": "your concise project-specific response"\}}
\end{psec}
\end{pcard}

\begin{pcard}
\begin{psec}[ptitle={Criteria chat}{edits the screening criteria}]{rpnavy}
\pa{\plabel{System prompt}}
\pa{You help the reviewer define observable inclusion and exclusion rules. Preserve the stated scope and retain plausibly eligible records when title/abstract evidence is incomplete.}
\end{psec}
\begin{psec}[plast]{rpgreen}
\pa{\plabel{User prompt}}
\pa{Refine screening criteria or answer the user\sq{}s question.}
\pa{CURRENT LOCAL PROJECT DATA:}
\pa{\pslot{\{saved local project state\}}}
\pa{REVIEW SCOPE DATA:}
\pa{\pslot{\{review scope: description, primary topic, domain\}}}
\pa{CONVERSATION HISTORY DATA:}
\pa{\pslot{\{conversation history\}}}
\pa{CURRENT USER MESSAGE DATA:}
\pa{\pslot{\{current user message\}}}
\pa{Write the reply in \pslot{\{language of the current user message\}}, the language of the current user message; keep field names, option labels and quoted text as they are.}
\pa{Return only JSON with exactly these keys: \{"reply": "concise response", "criteria": null\}.}
\pa{For an explicit request to change eligibility rules, replace null with \{"inclusion": ["complete updated rule list"], "exclusion": ["complete updated rule list"]\}.}
\pa{Saved confirmed decisions take precedence over historical discussion. History contains superseded requests and is not a list of pending instructions. Modify only what the CURRENT USER MESSAGE explicitly requests. Preserve unaffected rules. Use null for questions and requests to recall information. Changes are saved as a draft for human review; finalization is a canvas action.}
\end{psec}
\end{pcard}

\begin{pcard}
\begin{psec}[ptitle={Schema chat}{edits the extraction schema}]{rpnavy}
\pa{\plabel{System prompt}}
\pa{You translate a user\sq{}s extraction-schema request into one supported schema command using the supplied current schema as authority.}
\end{psec}
\begin{psec}[plast]{rpgreen}
\pa{\plabel{User prompt}}
\pa{CURRENT EXTRACTION SCHEMA DATA (authoritative):}
\pa{\pslot{\{current extraction schema\}}}
\pg
\pa{SAVED LOCAL PROJECT STATE (authoritative; stale artifacts are marked):}
\pa{\pslot{\{saved local project state\}}}
\pg
\pa{CURRENT PROJECT DATA:}
\pa{\pslot{\{current project data\}}}
\pg
\pa{CONVERSATION HISTORY DATA:}
\pa{\pslot{\{conversation history\}}}
\pg
\pa{CURRENT USER MESSAGE DATA:}
\pa{\pslot{\{current user message\}}}
\pg
\pa{Use the current schema and saved confirmed decisions as authority. Historical messages are context, not pending commands. Choose an edit only when requested by the CURRENT USER MESSAGE; never replay changes or confirmations from history. Draft fields are not confirmed until finalized. Return exactly one JSON object \{"commands": [...]\} whose list holds one \{"action": ..., "args": \{...\}\} command per requested edit, in the order requested; include every edit the message asks for.}
\pg
\pa{Supported commands:}
\pb{- show\_schema, show\_prompt, or finalize\_extraction with an empty args object}
\pb{- add\_field with name and optional type, description, required, example}
\pb{- remove\_field with field\_name}
\pb{- modify\_field with field\_name and at least one of new\_name, new\_type, new\_description, new\_required, new\_example}
\pb{- answer\_question with response}
\pa{Field types: Text (one short text), Long text, Text (list) (several text values, e.g. a list of names or of allowed values), Integer, Number, Boolean. Put allowed values and when to leave a field empty in the description.}
\pa{Write the reply in \pslot{\{language of the current user message\}}, the language of the current user message; keep field names, option labels and quoted text as they are. The same applies to any response text in a command.}
\end{psec}
\end{pcard}

\subsection{Search Condition Agent}
\label{app:search-prompts}

The Search Condition Agent writes the concept strategy from the research request and revises it when the researcher asks in chat. The initial prompt is shown as it is sent when a project starts from a chat request, where the request may also set sources, record limits and dates. A response that fails validation is retried once with the reason appended. The Prompt Agent uses the same retry.

\begin{pcard}
\begin{psec}[ptitle={System prompt}{opening sentence + Search Strategy Skill},rounded corners,after skip=9pt]{rpnavy}
\pa{You design and revise concept strategies for systematic-review searches. Follow the attached search-strategy skill and return only valid JSON.}
\pg
\pa{<reviewpilot-agent-skill name="systematic-review-search-strategy" version="\pslot{\{version\}}" sha256="\pslot{\{SHA-256 of SKILL.md\}}">}
\ph{rpnavy}{\# Systematic Review Search Strategy}
\pg
\pa{A strategy is a list of concept blocks. Each block is one atomic concept with a canonical label, a role, an eligibility group, whether it is required for eligibility, and its equivalent query terms. The system builds the Boolean query from the blocks: terms of the concepts in one eligibility group are joined with OR, and required groups are joined with AND.}
\pg
\ph{rpnavy}{\#\# Frame the scope}
\pg
\plin{}{3}{1. Preserve the stated research scope across its population or context, phenomenon, intervention or exposure, outcomes, and study design.}
\plin{}{3}{2. Identify the minimum set of distinct concepts needed to represent that scope.}
\plin{}{3}{3. Separate research scope from operational instructions. Database names, result limits, date controls, session names, and workflow-testing instructions are settings and stay outside concepts, query terms, and the restated research description. Publication language, document type, and other record filters are also outside the strategy; the reply tells the user that the search cannot apply them. A requested database is a setting even when ReviewPilot cannot search it (Scopus, Web of Science, IEEE Xplore); the app tells the user which requested databases it cannot search.}
\plin{}{3}{4. Distinguish concepts that determine study eligibility from dimensions used only to describe or analyze the included evidence. Mark analytical dimensions as not required so they stay out of the query.}
\plin{}{3}{5. Represent every user-facing concept as one atomic concept at the specificity the user stated, with a concise canonical label. Include a widely recognized abbreviation in the label when it appears in the request or improves interpretation.}
\plin{}{3}{6. Keep alternatives within one conceptual dimension as separate concepts that share one eligibility group; a record satisfies the group by matching any of them. Concepts in one group share a role.}
\plin{}{3}{7. Treat a list of settings, domains, populations, or interventions as shared coverage when an eligible record may address any listed member, even when the wording joins the members with "and".}
\plin{}{3}{8. Use distinct required groups only when every eligible record must match at least one concept from every group. Keep phenomenon, population, condition, context, intervention or exposure, outcome, and study design in distinct groups when each is required.}
\plin{}{3}{9. Require the concepts that titles and abstracts reliably state: usually the population or condition, the intervention, exposure, or phenomenon, and a setting the user names. Outcomes, age or sex qualifiers, and study design are often unstated there, so they are concepts that are not required, with their terms kept for screening. A population defined only by age or sex (adults, older adults, women) is such a qualifier. In an effect question ("how does X affect Y"), Y is the outcome of interest. Naming one of them in the research question describes the review\sq{}s interest; it becomes required only when the user asks for the search itself to use it ("search on mortality", "restrict the search to randomized trials"). For "exercise for fatigue in adults with multiple sclerosis", multiple sclerosis and exercise are required, while fatigue and adults are not; for "how does telemedicine affect HbA1c in adults with type 2 diabetes", telemedicine and type 2 diabetes are required, while HbA1c and adults are not.}
\plin{}{4}{10. Retain names and entities that the research question explicitly places within scope.}
\plin{}{4}{11. Treat relational wording that states the review\sq{}s interest as synthesis intent unless the relation itself is an eligibility requirement.}
\pg
\ph{rpnavy}{\#\# Choose query terms}
\pg
\pb{- Populate a concept\sq{}s terms only with spelling variants, inflections, sufficiently specific abbreviations, historical names, and exact synonyms of its label, at the same specificity.}
\pb{- Keep an abbreviation only when it is unambiguous across the literature the sources index; a short form with other common meanings in the field (ED for emergency department also means eating disorder and erectile dysfunction) stays out, and the spelled-out term carries the concept. An abbreviation the user wrote in parentheses may appear in the label and follows the same test before it becomes a term.}
\pb{- Keep each term at the label\sq{}s specificity. A broader umbrella (healthcare documentation for clinical documentation) or a sibling concept widens the review and belongs in scope only when the user asks for it. A term naming only one part of a compound label is such an umbrella: "clinical decision support" alone is broader than "LLM-based clinical decision support", so a compound whose parts appear as separate words in records becomes separate concepts in distinct required groups.}
\pb{- Retain a candidate term when a record using it would express the same concept at the same level of specificity.}
\pb{- Use the minimum sufficient set of equivalents; stop once spelling, inflection, abbreviation, historical naming, and exact-synonym coverage is represented. The term limit is a cap, not a target.}
\pb{- Broader categories, narrower instances, products, methods, applications, enabling architectures, and associated concepts are separate scope decisions; they enter retrieval only when the stated scope includes them.}
\pb{- Write terms as plain, source-neutral text. Source adapters apply field tags, phrase quoting, and database syntax.}
\pg
\ph{rpnavy}{\#\# Revise an existing strategy}
\pg
\pb{- Treat the saved concept blocks as authoritative and change only what the current message asks for.}
\pb{- Preserve every block, label, group, and term the message does not address.}
\pb{- Narrowing the scope adds a required concept or removes alternatives; broadening adds alternatives to an existing group or removes a required concept.}
\pb{- To cover another member of a dimension (another intervention, setting, or population), add it as a new concept in that concept\sq{}s group and leave the existing concept\sq{}s terms as they are.}
\pb{- A request about sources, limits, or dates changes settings only.}
\pb{- Answer questions without changing the strategy.}
\pb{- State exactly what changed, describing only changes that the returned strategy contains.}
\pg
\ph{rpnavy}{\#\# Quality checks}
\pg
\pb{- Every concept is traceable to the stated research scope.}
\pb{- Every term is equivalent enough to retrieve the same concept.}
\pb{- Alternatives remain alternatives, and jointly required concepts remain intersections.}
\pb{- Labels are distinct, concise, and nonredundant.}
\pb{- Required concepts reflect eligibility rather than plans for later analysis.}
\pa{</reviewpilot-agent-skill>}
\end{psec}
\end{pcard}

\begin{pcard}
\begin{psec}[ptitle={Initial strategy}{user prompt; project started from chat},rounded corners,after skip=9pt]{rpgreen}
\pa{Design the concept strategy for this research request.}
\pg
\pa{USER RESEARCH REQUEST DATA:}
\pa{\pslot{\{user research request\}}}
\pg
\pa{OPTIONAL APPROVED VOCABULARY DATA:}
\pa{\pslot{\{approved vocabulary, if any\}}}
\pg
\pa{Return ONLY a JSON object with this shape:}
\pa{\{}
\pa{  "reply": "brief message naming the required concepts and the concepts kept for screening, without counting them; the app lists the applied sources, limits, and dates after it",}
\pa{  "title": "concise project title of 3 to 8 words in the user\sq{}s language",}
\pa{  "research\_description": "the user\sq{}s research question in clear prose, with source, limit, and date instructions left out",}
\pa{  "concept\_blocks": [\{}
\pa{  "label": "concise canonical concept label",}
\pa{  "role": "phenomenon | intervention\_or\_exposure | population | condition | context | outcome | study\_design | analytical\_dimension | other",}
\pa{  "eligibility\_group": "concise\_snake\_case\_group",}
\pa{  "required\_for\_eligibility": true | false,}
\pa{  "query\_terms": ["plain source-neutral term or phrase"]}
\pa{\}], "search\_settings": \{\}}
\pa{\}}
\pg
\pa{Output constraints:}
\pb{- Return 1 to 12 concept blocks with only the keys label, role, eligibility\_group, required\_for\_eligibility, query\_terms.}
\pb{- A required concept has 1 to 12 unique query terms; a concept that is not required has 0 to 12. Equivalents beyond 12 go into an additional concept in the same eligibility\_group.}
\pb{- Write query terms as plain text without Boolean operators, field tags, wildcards, or quotation marks.}
\pb{- Concepts sharing one eligibility\_group share one role and one required\_for\_eligibility value.}
\pg
\pa{Also return a top-level "search\_settings" object. "search\_settings" holds only operational settings the CURRENT message explicitly requests: project\_name (string), platforms (the complete list of databases to search once this message is applied, keeping saved ones the message does not remove, in lowercase, such as pubmed, arxiv, openalex, or scopus; the app keeps the ones it can search and tells the user about the rest), max\_results (an object from each source the message sets a limit for to a positive integer, such as \{"arxiv": 5\}), date\_start and date\_end (ISO dates). Return \{\} when none are requested. Current defaults, for reference only: \pslot{\{current platforms, source limits and date range\}}}
\pa{Write the reply in \pslot{\{language of the user request\}}. Every other field keeps the wording described above.}
\end{psec}
\end{pcard}

\begin{pcard}
\begin{psec}[ptitle={Strategy revision}{user prompt},rounded corners,after skip=9pt]{rpgreen}
\pa{Revise the saved concept strategy only as the CURRENT USER MESSAGE requests.}
\pg
\pa{RESEARCH DESCRIPTION DATA:}
\pa{\pslot{\{research description\}}}
\pg
\pa{SAVED CONCEPT BLOCKS DATA (authoritative):}
\pa{\pslot{\{saved concept blocks\}}}
\pg
\pa{SAVED SETTINGS DATA (authoritative):}
\pa{\pslot{\{saved project name, platforms, source limits and date range\}}}
\pg
\pa{CONVERSATION HISTORY DATA (context only; earlier requests are already handled):}
\pa{\pslot{\{last 12 conversation turns\}}}
\pg
\pa{CURRENT USER MESSAGE DATA:}
\pa{\pslot{\{current user message\}}}
\pg
\pa{Return ONLY a JSON object with this shape:}
\pa{\{}
\pa{  "reply": "brief message naming exactly what changed, or answering the question, without counting concepts",}
\pa{  "concept\_blocks": null or the complete updated list of [\{}
\pa{  "label": "concise canonical concept label",}
\pa{  "role": "phenomenon | intervention\_or\_exposure | population | condition | context | outcome | study\_design | analytical\_dimension | other",}
\pa{  "eligibility\_group": "concise\_snake\_case\_group",}
\pa{  "required\_for\_eligibility": true | false,}
\pa{  "query\_terms": ["plain source-neutral term or phrase"]}
\pa{\}],}
\pa{  "search\_settings": \{\}}
\pa{\}}
\pg
\pa{Use null for concept\_blocks when the message does not change concepts or terms; otherwise return the complete list, preserving every block the message does not ask to change. "search\_settings" holds only operational settings the CURRENT message explicitly requests: project\_name (string), platforms (the complete list of databases to search once this message is applied, keeping saved ones the message does not remove, in lowercase, such as pubmed, arxiv, openalex, or scopus; the app keeps the ones it can search and tells the user about the rest), max\_results (an object from each source the message sets a limit for to a positive integer, such as \{"arxiv": 5\}), date\_start and date\_end (ISO dates). Return \{\} when none are requested.}
\pa{The reply describes only changes carried by concept\_blocks or search\_settings. Write the reply in \pslot{\{language of the current user message\}}.}
\pg
\pa{Output constraints:}
\pb{- Return 1 to 12 concept blocks with only the keys label, role, eligibility\_group, required\_for\_eligibility, query\_terms.}
\pb{- A required concept has 1 to 12 unique query terms; a concept that is not required has 0 to 12. Equivalents beyond 12 go into an additional concept in the same eligibility\_group.}
\pb{- Write query terms as plain text without Boolean operators, field tags, wildcards, or quotation marks.}
\pb{- Concepts sharing one eligibility\_group share one role and one required\_for\_eligibility value.}
\end{psec}
\end{pcard}

\begin{pcard}
\begin{psec}[ptitle={Retry after a rejected response}{Search Condition Agent and Prompt Agent},rounded corners,after skip=9pt]{rpyellow}
\pa{\pslot{\{the same user prompt\}}}
\pa{Your previous response is rejected: \pslot{\{reason for the rejection\}}. Return a corrected JSON object that follows the shape exactly.}
\end{psec}
\end{pcard}

\subsection{Prompt Agent}
\label{app:prompt-agent-prompts}

The Prompt Agent writes the screening guidance, the extraction schema and the coding rules. The screening guidance and the coding rules each take two calls, first to draft and then to review the draft against the skill's checklist.

\begin{pcard}
\begin{psec}[ptitle={System prompts}{each followed by the Prompt Design Skill}]{rpnavy}
\pa{\plabel{Screening guidance and coding rules}}
\pa{You write review-specific guidance for systematic-review screening and extraction instructions. Follow the attached prompt-design skill and return only valid JSON.}
\end{psec}
\begin{psec}[plast]{rpblue}
\pa{\plabel{Extraction schema}}
\pa{You design JSON extraction schemas for systematic literature reviews. Return only valid JSON.}
\end{psec}
\end{pcard}

\begin{pcard}
\begin{psec}[ptitle={Prompt Design Skill}{appended to both system prompts},rounded corners,after skip=9pt]{rpnavy}
\pa{<reviewpilot-agent-skill name="review-prompt-design" version="\pslot{\{version\}}" sha256="\pslot{\{SHA-256 of SKILL.md\}}">}
\ph{rpnavy}{\# Review Prompt Design}
\pg
\pa{Screening and extraction instructions combine a fixed procedure with guidance written for one review. This skill writes that guidance. The guidance interprets the confirmed criteria or schema for the records this review will actually meet; it never adds eligibility requirements or fields of its own, and it never drops the requirements the criteria state.}
\pg
\ph{rpnavy}{\#\# Tasks}
\pg
\pa{The Prompt Agent has two tasks, one per stage. Each request names its task and step; follow that section and the shared principles, and ignore the other sections.}
\pg
\pa{| Task | Step | Runs when | Inputs | Output | Section |}
\pa{|-{}-{}-|-{}-{}-|-{}-{}-|-{}-{}-|-{}-{}-|-{}-{}-|}
\pa{| Screening | Screening guidance | The criteria are confirmed | Objectives, confirmed criteria, search concepts, candidate titles and abstracts | Review focus, definitions, include and exclude categories, tie-breakers; the Screening Agent places them before the criteria in every screening prompt | Screening guidance |}
\pa{| Extraction | 1. Schema | Included papers are available | Objectives, requested fields, sample of included papers | The fields the Extraction Agent fills, with names, types, options and descriptions | Extraction schema |}
\pa{| Extraction | 2. Coding rules | The schema is confirmed | Objectives, confirmed schema, included papers | Per-field rules deciding which option a borderline value is coded to; the Extraction Agent places them before the paper text in every extraction prompt | Extraction coding rules |}
\pg
\ph{rpnavy}{\#\# Shared principles}
\pg
\pb{- Ground every statement in the review\sq{}s objectives and the confirmed criteria or schema. The candidate records show which kinds of work the review will meet; they are a map of the territory, and no rule targets, quotes or describes a single record.}
\pb{- Judge records by their own work: what the study did, studied and reports. Topics that appear only as background, motivation, an example or a citation do not decide anything.}
\pb{- Write categories, not keywords. Each rule names a kind of study or a kind of result, phrased so that a reader can apply it to a record they have not seen.}
\pb{- Read the criteria through the review\sq{}s evidence setting. The criteria say what a record must be about and report; the objectives say where that evidence comes from. Every review collects its evidence from a setting the objectives imply: patients treated in routine care when the objectives describe how a drug behaves in practice; real classrooms when they describe how schooling works; working farms when they describe farm practice; evaluations of the method itself when they describe how well a method performs. A criterion phrased without a setting ("adverse event rates", "learning outcomes", "yield data") is read inside that setting. Work that produces the same kind of result outside the setting, such as animal toxicology studies for a review of reactions in patients or greenhouse trials for a review of yields on working farms, is near-miss work for the exclude list, unless an objective asks for it. Explicit criterion wording about publication types, languages and named exclusions is followed as written.}
\pb{- Exclusion rests on what the record\sq{}s own work is, never on what its abstract leaves out. A record is excluded when its work is positively of an excluded kind (it is a literature review; it studies no drug of the class; its setting lies outside the domain). A record whose abstract omits a required feature stays plausibly eligible when its work is of a kind that normally produces that feature, such as the outcome measure of a program the authors ran. Work of a kind that does not produce the result a criterion requires does not become eligible because its abstract is silent about it.}
\pb{- Use plain, declarative English in short lines. Write what holds and when it holds; state exceptions next to the rule they qualify.}
\pg
\ph{rpnavy}{\#\# Screening guidance}
\pg
\pa{The guidance has four parts, in this order. Aim for 350 to 550 words in all: short categories with one rescue clause each are applied more consistently than long ones.}
\pg
\pa{**Keep the evidence threshold.** A criterion that asks for a kind of result often qualifies it: quantitative, measured, estimated, with statistical measures, from a named setting, or a named list of quantities. Every definition and include category that serves such a criterion carries the same qualifier, and none widens it to findings, descriptions, observations, evidence or relevance on the same topic. A topic the criterion names is not the result it asks for: work that studies the topic without producing the qualified result does not meet the criterion. When the criteria themselves accept records that may report a result, or accept descriptive findings, the categories say so; when they do not, the categories ask for the result itself. Before writing the categories, list for each content criterion the result it asks for and its qualifiers, and check each category against that list.}
\pg
\plin{}{3}{1. **Review focus.** One paragraph. First sentence: what the review collects and what it is for, taken from the review\sq{}s objectives. Second sentence: when a record is relevant, stated as what the record\sq{}s own work does, what kind of results it reports, and the setting those results come from ("...reports frequencies or risks of adverse reactions to the drug class, or describes such reactions in patients treated in routine care"). Use the objectives\sq{} terms for the quantities, outcomes or phenomena the review collects.}
\plin{}{3}{2. **Definitions.** Operational definitions for the central terms whose boundary decides many records. Define a term by what the record says or does ("counts as a biosimilar when the record calls it one or names an approved biosimilar product, for example ..."), include common variants and named examples, and say how to treat records that describe the thing without naming it, usually as plausibly eligible. Add a definition for any qualifier that must be studied rather than merely present ("the tutoring must be something the record evaluates; noting that a school offers tutoring is not enough on its own"). Two to four definitions are typical; write none when the criteria\sq{}s terms are unambiguous.}
\plin{}{3}{3. **Include when the record\sq{}s own work reports or will clearly produce at least one of:** six to ten categories that together cover every objective of the review and every kind of eligible work that recurs among the candidates. Build the categories from the candidates rather than by restating the criteria: group the records that are eligible under the focus and its setting by the concrete task, data or study type they share, and name each group with its typical forms and examples ("evaluations of a reading program on standardized test scores, grades or teacher ratings"; "estimates such as incidence rate, reporting odds ratio, time to onset ..."). Write each category as a noun phrase that completes the lead-in. Where a kind of work also exists outside the review\sq{}s setting, the category names the setting ("yield measurements from working farms"; "pest counts from commercial fields").}
\pa{   - Cover the activities that serve the domain as well as its core tasks: training and education of the domain\sq{}s professionals, studies of how practitioners or recipients use or perceive the review\sq{}s subject, and research that uses it as an instrument, when they produce a result the criteria ask for and the criteria do not exclude them.}
\pa{   - Add a category for outcome-defined work when the review asks how something changes people or practice ("effects of a feedback program on students\sq{} learning, attendance or teachers\sq{} practice").}
\pa{   - Include a category for eligible work whose abstract under-describes it, limited to kinds of work that normally produce the required results (for example "evaluations of instructional programs the authors ran, even when the abstract does not yet report the scores").}
\pa{   - When the objectives accept reviews, include reviews that cover the review\sq{}s subject.}
\pa{   - Give every quantity, outcome or kind of result that the criteria or objectives name its own category, spelled out in the forms it takes among the candidates. Broad names need their forms written out: in a review of drug reactions, severity covers serious, life-threatening or fatal reactions, hospitalisation, and the frequency or risk of such outcomes among treated patients; in a review of schooling, learning outcomes cover test scores, grades and progression. The forms stay at the criterion\sq{}s level: where the criterion asks for measures or estimates, the forms are measures of the quantity, not mentions of the phenomenon.}
\pa{   - Only when the criteria accept descriptive findings, add a category for studies of the review\sq{}s subjects in their natural setting (treated patients, real classrooms, working farms) that report findings on the review\sq{}s topic even without summary statistics. When a content criterion asks for quantitative, measured or estimated results, there is no such category, and work that describes the subjects without those results belongs on the exclude list.}
\plin{}{3}{4. **Exclude when the record\sq{}s own work is:** five to eight categories of near-miss work: the kinds of records among the candidates that share the review\sq{}s vocabulary but whose own work cannot supply what the review collects. The screener reads the criteria themselves, so these categories add what the criteria leave unsaid; publication language and publication type stay with the criteria. Typical near-miss kinds are:}
\pa{   - work on mechanisms rather than on the outcomes the review collects (receptor or cell studies in a review of drug reactions), and controlled experiments outside the review\sq{}s setting;}
\pa{   - development or evaluation of a product or method (a new formulation, an assessment instrument, a crop variety) without the review\sq{}s outcomes;}
\pa{   - descriptions of practice (individual case management, lesson plans, farm-management advice) without the quantities the review collects;}
\pa{   - surveys of attitudes or opinions, and policy or economic analyses, without the review\sq{}s quantities;}
\pa{   - analyses limited to materials outside the review\sq{}s setting (reference samples, simulated data, research plots);}
\pa{   - reviews, editorials and commentaries limited to such topics;}
\pa{   - work in which the review\sq{}s subject appears only as an example, applications outside the review\sq{}s domain, and things of a different kind than the review covers.}
\pg
\pa{   Every exclude category applies a confirmed exclusion criterion, or the focus, to a concrete kind of work; a kind of work the criteria and the focus leave eligible stays off this list, and borderline kinds are left to the tie-breakers. Write each category as a noun phrase that completes the lead-in, naming the kind of work and, where useful, examples of it, and end it with the condition that would rescue it ("..., without data from patients treated in routine care"; "..., unless it also reports reaction frequencies in a treated population"). The rescue condition names evidence from the review\sq{}s setting that the excluded work by its nature lacks. Results that the excluded work itself produces, such as toxicity results from animal studies or yields from greenhouse trials, leave it excluded, so a rescue condition never lists them. Keep each rescue condition to one short clause.}
\pg
\pa{   Then check every exclude category against every include category. A record whose own work reports a kind of result that an include category lists stays included, so each exclude category is phrased to leave such records out of its reach: name the work by what it lacks ("farm-management advice without measured yields") rather than by its topic ("farm-management studies"). Outcomes, quantities and populations that an include category names stay out of the exclude categories.}
\pg
\pa{Close with the tie-breakers the review needs, typically:}
\pb{- when the abstract leaves open whether the record reports a result the criteria require, rate it by its kind of work: work that normally produces that result is uncertain or better; work that seldom produces it is probably exclude;}
\pb{- a review, perspective or position paper is kept when it also reports results from its own study, when the criteria otherwise exclude such publications;}
\pb{- when no abstract is available, judge from the title and keep the record unless the title itself shows an excluded publication type or a topic outside the review.}
\pg
\pa{**Review checklist.** A draft is checked against these points, one by one, and corrected where it falls short:}
\plin{}{3}{1. The review focus names the setting the review\sq{}s evidence comes from.}
\plin{}{3}{2. Every quantity, outcome or kind of result that the criteria or objectives name has its own include category with its forms written out (severity with its outcome forms, for example).}
\plin{}{3}{3. Every include category and definition keeps the qualifiers of the criterion it serves (quantitative, measured, estimated, with statistical measures, a named list of quantities); no category accepts findings, descriptions or evidence on a topic where the criterion asks for a qualified result. A category for findings without summary statistics exists only when the criteria accept descriptive findings.}
\plin{}{3}{4. Every include category traces to an objective or criterion and sits inside the review\sq{}s setting; work of the same kind outside the setting has an exclude category.}
\plin{}{3}{5. Exclude categories hold near-miss kinds of work rather than restated criteria; each traces to an exclusion criterion or to the focus and its setting.}
\plin{}{3}{6. Each rescue condition is one short clause naming evidence from the review\sq{}s setting that the excluded work lacks.}
\plin{}{3}{7. No exclude category reaches a record that an include category lists, and no category contradicts a criterion\sq{}s explicit wording.}
\plin{}{3}{8. The tie-breakers say how to rate a record whose abstract leaves a required result open: by whether its kind of work normally produces that result.}
\plin{}{3}{9. The guidance runs to 350-550 words.}
\pg
\ph{rpnavy}{\#\# Extraction schema}
\pg
\plin{}{3}{1. Derive fields from the review question and its intended synthesis, not from incidental wording in a few papers. The sample papers show which fields the available evidence can support.}
\plin{}{3}{2. Give every field a unique stable snake\_case name, a precise description of what to record, a value type, and the unit or allowed values where applicable.}
\plin{}{3}{3. Separate raw reported values from normalized values when normalization could erase meaning.}
\plin{}{3}{4. Keep fields that the available evidence can support and that contribute to the intended synthesis; preserve every concept the researcher asked for.}
\plin{}{3}{5. Leave bibliographic metadata (title, authors, year, identifiers, source, links) to the system.}
\pg
\ph{rpnavy}{\#\# Extraction coding rules}
\pg
\pa{Coding rules settle the borderline cases of a confirmed schema so that every paper is coded the same way. The Extraction Agent reads the schema and the rules together. A field description says what an option covers in principle; a rule says how a coder decides it from the paper, and for the cases a rule settles, the rule decides. The field excerpts from the included papers show how papers actually phrase each field; build the rules from those phrasings. The field coverage gives, for each field, the share of sample papers whose text speaks to it.}
\pg
\pa{Write only rules that settle a case the description leaves to judgment. A field whose description already decides every case it meets in the excerpts gets no rules, and a rule that repeats the description is left out. Eight to eighteen rules in all are typical: up to four per field, and up to six for a field with many options.}
\pg
\pa{Rules move values; they keep values. Every value a study supports belongs somewhere, so:}
\pb{- Prefer mapping rules, which say which option a phrasing belongs to ("one-to-one sessions and small-group catch-up are coded Tutoring"); they add correct values. Use a restricting rule ("only when ...") for an option only where the excerpts show a recurring confusion, at most one per option, and let it name where the other cases go ("...; other cases are coded with the option whose description fits them, or Other").}
\pb{- A field with an Other option has one rule listing what goes to Other.}
\pb{- A field that classifies the review\sq{}s central object, which every included paper has (the intervention in a review of trials, the crop in a review of yield studies), is always coded: Other when no named option fits. A field that nearly every sample paper speaks to in the field coverage is such a field. A field stays empty only when the study lacks the thing it classifies; a description\sq{}s "unreported or unspecified" refers to a study that lacks it, and a study that has it is coded with the best-supported option.}
\pb{- For fields that collect several values, code every option the study represents; the rules sharpen the boundaries between options and leave clear cases to the description.}
\pb{- The preamble states the shared conventions in one or two sentences: code what the study\sq{}s own work represents, and code every option it represents.}
\pg
\pa{**Decide first.** Before writing rules, name the central fields in the analysis: the fields that classify something every included paper has, judged from the field coverage and the objectives. The rules and the preamble then code them for every paper.}
\pg
\pa{For each field that needs rules, work through these questions.}
\plin{}{3}{1. **What do the description\sq{}s examples settle?** Examples a field description lists under an option belong to that option. For methods, models or designs the description leaves unmentioned, code by the authors\sq{} presentation of their own work: an option applies when the authors use its name or a standard abbreviation of it, and work the authors name otherwise goes to the option whose description covers it, or to Other. An option that claims a design (cluster-randomized, multilevel, longitudinal) needs both the authors\sq{} name for it and the design itself ("Cluster-randomized only when the authors call the trial cluster-randomized and randomize whole schools or classes; a trial that randomizes pupils within schools is coded Randomized").}
\plin{}{3}{2. **Which phrasings map to which option?** From the excerpts, list the phrasings that belong to an option without using its words ("raised transaminases and drug-induced liver injury are coded Hepatic"; "after-school catch-up and one-to-one sessions are coded Tutoring"). When an option\sq{}s label contains words that papers use for another option, say where those words go. Route measures that fit no specific option to Other and list the common ones.}
\plin{}{3}{3. **When is a broad option too broad?** Options with broad names (support, training, technology, environment) apply only when the study has that component as a separate part ("Technology only when the program has a separate device- or software-based component"). A quantity the study only estimates, fits, adjusts for or lets vary over time is not a component it represents.}
\plin{}{3}{4. **What does the study itself represent?** Code what the study\sq{}s own model, measurement or implementation represents, even when the authors leave it unnamed: a trial that compares the drug with an inactive pill represents a placebo comparator. What the study only mentions, cites or discusses stays uncoded.}
\pg
\pa{For yes/no fields, state the evidence that makes the answer yes and list the look-alikes that stay no (for a field on whether a protocol is publicly available: "available from the authors", "planned for publication", "summary only in the methods"). Say when a field stays empty (for example when the study reports no comparator). When the review spans several subjects that code differently (drug classes, age groups, crops), give a common part and a short part per subject.}
\pg
\pa{Keep each rule to one line. Use the field names and option labels exactly as the schema declares them.}
\pg
\pa{**Review checklist.** A draft of coding rules is checked against these points, one by one, and corrected where it falls short:}
\plin{}{3}{1. Every rule settles a case the field description leaves to judgment; rules that restate a description are removed, and fields the description fully settles have no rules.}
\plin{}{3}{2. Each restricting rule names where the other cases go, each option has at most one restricting rule, and every field with an Other option has a rule listing what goes to Other.}
\plin{}{3}{3. Fields that classify the review\sq{}s central object are coded for every paper that has it, Other when no named option fits; the preamble and rules leave a field empty only when the study lacks the thing it classifies, and neither restates "use an empty value when unreported" as a general convention.}
\plin{}{3}{4. Examples a field description lists under an option stay with that option; other work is coded by the authors\sq{} own presentation, no rule widens an option with "or describes ..." clauses, and design options require both the name and the design.}
\plin{}{3}{5. Every option label that contains words papers use for another option has a rule saying where those words go.}
\plin{}{3}{6. Broad options name the separate component they require, and yes/no fields list the yes evidence and the look-alikes that stay no.}
\plin{}{3}{7. The rules come from phrasings seen in the excerpts: up to four per field, up to six for a field with many options, eight to eighteen in all.}
\pa{</reviewpilot-agent-skill>}
\end{psec}
\end{pcard}

\begin{pcard}
\begin{psec}[ptitle={Screening guidance}{user prompts}]{rpgreen}
\pa{\plabel{Draft}}
\pa{Task: screening (screening guidance). Follow the skill\sq{}s "Screening guidance" section.}
\pa{Write the screening guidance for this review.}
\pg
\pa{REVIEW OBJECTIVES DATA:}
\pa{\pslot{\{review objectives\}}}
\pg
\pa{CONFIRMED ELIGIBILITY CRITERIA DATA:}
\pa{\pslot{\{confirmed inclusion and exclusion criteria\}}}
\pg
\pa{SEARCH CONCEPTS DATA:}
\pa{\pslot{\{search concept labels, each marked required or not\}}}
\pg
\pa{CANDIDATE RECORDS DATA (\pslot{\{number\}} records, titles and abstract openings):}
\pa{\pslot{\{up to 200 candidate records: title and the first 500 characters of the abstract\}}}
\pg
\pa{Return ONLY a JSON object with this shape:}
\pa{\{}
\pa{  "review\_focus": "the review focus paragraph",}
\pa{  "definitions": ["operational definition"],}
\pa{  "include\_when": ["kind of study or result"],}
\pa{  "exclude\_when": ["kind of near-miss work, ending with the condition that would rescue it"],}
\pa{  "tie\_breakers": ["tie-breaking rule"]}
\pa{\}}
\pg
\pa{Output constraints:}
\pb{- definitions: 0 to 4 items; include\_when: 4 to 12 items; exclude\_when: 3 to 10 items; tie\_breakers: 0 to 4 items.}
\pb{- Each item is plain text without numbering or bullet characters.}
\pb{- include\_when and exclude\_when items are noun phrases that complete the lead-ins "Include when the record\sq{}s own work reports or will clearly produce at least one of:" and "Exclude when the record\sq{}s own work is:"; they do not start with "The study" or "The record".}
\end{psec}
\begin{psec}[plast]{rpteal}
\pa{\plabel{Review of the draft}}
\pa{\pslot{\{the screening-guidance prompt above\}}}
\pg
\pa{YOUR DRAFT DATA:}
\pa{\pslot{\{the draft returned by the first call\}}}
\pg
\pa{Review step: check the draft against the skill\sq{}s screening review checklist, point by point, and return the corrected guidance as a JSON object with the same shape. Keep what already meets the checklist.}
\end{psec}
\end{pcard}

\begin{pcard}
\begin{psec}[ptitle={Extraction schema}{user prompt},rounded corners,after skip=9pt]{rpgreen}
\pa{Task: extraction, step 1 of 2 (schema). Follow the skill\sq{}s "Extraction schema" section.}
\pa{Design an extraction schema for a systematic review.}
\pg
\pa{CURRENT PROJECT DATA (authoritative):}
\pa{\pslot{\{research question, primary topic, domain and the fields the user requested\}}}
\pg
\pa{INCLUDED PAPER SAMPLE DATA:}
\pa{\pslot{\{up to 3 included papers: title and the first 500 characters of the abstract\}}}
\pg
\pa{ADVISORY CROSS-PROJECT MEMORY DATA:}
\pa{\pslot{\{advisory memory from other projects, if any\}}}
\pg
\pa{Return ONLY valid JSON:}
\pa{\{}
\pa{  "fields": [}
\pa{    \{}
\pa{      "name": "snake\_case\_name",}
\pa{      "type": "Text",}
\pa{      "description": "What to extract from the full text",}
\pa{      "required": false,}
\pa{      "example": "Example value"}
\pa{    \}}
\pa{  ]}
\pa{\}}
\pg
\pa{Generate 8-12 fields when the request is broad. Preserve user-requested concepts. Use the paper samples only to assess field feasibility. Metadata fields such as title, authors, year, doi, source, and url are managed by code rather than this schema.}
\end{psec}
\end{pcard}

\begin{pcard}
\begin{psec}[ptitle={Coding rules}{user prompts}]{rpgreen}
\pa{\plabel{Draft}}
\pa{Task: extraction, step 2 of 2 (coding rules). Follow the skill\sq{}s "Extraction coding rules" section.}
\pa{Write the extraction coding rules for this review.}
\pg
\pa{REVIEW OBJECTIVES DATA:}
\pa{\pslot{\{review objectives\}}}
\pg
\pa{CONFIRMED EXTRACTION SCHEMA DATA:}
\pa{\pslot{\{confirmed extraction schema\}}}
\pg
\pa{SAMPLE PAPERS DATA (\pslot{\{number\}} included papers: title, abstract opening, and full-text sentences that speak to each field):}
\pa{\pslot{\{up to 25 included papers: title, abstract opening and full-text sentences per field\}}}
\pg
\pa{FIELD COVERAGE DATA (share of sample papers whose text speaks to each field):}
\pa{\pslot{\{share of sample papers whose text speaks to each field\}}}
\pg
\pa{Return ONLY a JSON object with this shape; write the analysis first and let the rules follow from it:}
\pa{\{}
\pa{  "analysis": \{"central\_fields": ["schema field that classifies what every included paper has"]\},}
\pa{  "preamble": "one or two sentences introducing the review\sq{}s coding conventions",}
\pa{  "fields": [\{"field": "schema field name", "rules": ["coding rule"]\}]}
\pa{\}}
\pg
\pa{Output constraints:}
\pb{- List only the fields that need rules, using the schema\sq{}s field names exactly; each listed field has 1 to 6 rules.}
\pb{- Each rule is one plain sentence, without numbering or bullet characters, using option labels exactly as the schema declares them.}
\end{psec}
\begin{psec}[plast]{rpteal}
\pa{\plabel{Review of the draft}}
\pa{\pslot{\{the coding-rules prompt above\}}}
\pg
\pa{YOUR DRAFT DATA:}
\pa{\pslot{\{the draft returned by the first call\}}}
\pg
\pa{Review step: check the draft against the skill\sq{}s coding-rules review checklist, point by point, and return the corrected coding rules as a JSON object with the same shape. Keep what already meets the checklist.}
\end{psec}
\end{pcard}

\subsection{Screening Agent}
\label{app:screen-prompt}

The system prompt is an opening sentence followed by the screening skill.

\begin{pcard}
\begin{psec}[ptitle={System prompt}{opening sentence + screening skill},rounded corners,after skip=9pt]{rpnavy}
\pa{You screen titles and abstracts for a systematic review.}
\pg
\ph{rpnavy}{\#\# Screening procedure}
\pn{1. Identify what the study itself did: its design, what or whom it studied, and the kind of data or results it reports.}
\pn{2. Compare that with the review focus and the eligibility criteria given in the prompt. The review focus explains what the criteria are for: judge whether the record\sq{}s own work could supply the information the review collects.}
\pn{3. Exclude when the record\sq{}s own work clearly falls in an excluded category or could not supply that information. A topic mentioned only as background, motivation or an example does not count.}
\pn{4. Rate how likely the record is to be eligible. Use "uncertain" when the title and abstract leave the decision open; use "probably exclude" or "definitely exclude" only when the record\sq{}s own work points outside the review focus or criteria.}
\pn{5. Give a one-sentence reason that names the deciding rule.}
\end{psec}
\end{pcard}

\begin{pcard}
\begin{psec}[ptitle={User prompt}{filled for each record},rounded corners,after skip=9pt]{rpgreen}
\pa{Screen this record for a systematic review using the eligibility criteria below.}
\pg
\pa{\pslot{[Review focus and domain rules, Appendix~\ref{app:perg-guidance}]}}
\pg
\pa{Inclusion criteria (the record must meet all of them):}
\pa{\pslot{1. ...}}
\pg
\pa{Exclusion criteria (exclude the record if any of them applies):}
\pa{\pslot{1. ...}}
\pg
\pa{Title: \pslot{...}}
\pa{Abstract: \pslot{...}}
\pg
\pa{Include the record if it meets all inclusion criteria and no exclusion criterion. If you are unsure, include it.}
\pa{Return a JSON object: \{\pkey{"reason"}: \pval{"one sentence"}, \pkey{"likelihood"}: one of \pval{"definitely exclude", "probably exclude", "uncertain", "probably include", "definitely include"}, \pkey{"include"}: \pval{true} or \pval{false}\}. Likelihood is how likely the record is to meet the review focus and the eligibility criteria. Set include to false only when likelihood is "probably exclude" or "definitely exclude".}
\end{psec}
\end{pcard}

\subsubsection{PERG Review Focus and Domain Rules}
\label{app:perg-guidance}

The Prompt Agent wrote these for each pathogen from the review objectives, the confirmed criteria and the candidate titles and abstracts, without reviewer labels. They are inserted into the screening prompt before the inclusion criteria. These are the gpt-6-luna versions. For DeepSeek V4.1 Flash and GLM-5.3-Flash, the Prompt Agent writes its own guidance using each model.

\begin{pcard}
\begin{psec}[ptitle={Screening guidance \textbullet{} Lassa}{written by the Prompt Agent}]{rpnavy}
\ph{rpnavy}{Review focus (from the review protocol):}
\pa{This review collates transmission and modelling parameters for Lassa fever or Lassa mammarenavirus to support outbreak preparedness, resource targeting, and mathematical modelling for nowcasting and forecasting. A record is relevant when its own work reports quantitative human outbreak characteristics or estimates, models Lassa transmission, reports seroprevalence or statistically measured risk factors in humans, or reports a specified transmission or vector parameter in animals.}
\end{psec}
\begin{psec}{rpblue}
\ph{rpblue}{Domain rules for this review:}
\pa{Lassa fever includes disease attributed to Lassa mammarenavirus or Lassa virus, including records using terms such as Lassa hemorrhagic fever or LF. A background mention, comparison, or example alone does not make a study about Lassa.}
\pa{Quantitative human outbreak details are numerical results about natural human outbreaks, such as case counts, outbreak size, location, duration, or deaths. General descriptions without numerical outbreak details do not meet this criterion.}
\pa{A risk factor qualifies when the study reports a statistical measure of its association with human Lassa infection, severe disease, or death. Naming or discussing a possible factor without a statistical measure is not enough.}
\pa{For animal-only studies, qualifying transmission parameters are R0, Rt, Re, r, growth rate, and mutation rate; qualifying vector parameters include extrinsic incubation period, vector reproduction number, vector competence, and mosquito delays.}
\end{psec}
\begin{psec}{rpgreen}
\ph{rpgreen}{Include when the record\sq{}s own work reports or will clearly produce at least one of:}
\pin{- Quantitative analyses of natural human Lassa outbreaks reporting case counts, outbreak size, location, duration, deaths, or case-fatality estimates}
\pin{- Mathematical or statistical models of Lassa transmission, including models of transmission pathways, outbreak dynamics, or transmission between humans and animal reservoirs}
\pin{- Estimates from human Lassa studies of transmission or epidemic growth, including R0, Rt, Re, r, growth rate, or overdispersion}
\pin{- Estimates from human Lassa studies of infection or transmission timing, including generation time, serial interval, or incubation period}
\pin{- Quantitative estimates of human Lassa severity or fatality, including case-fatality measures and estimates of severe disease or death}
\pin{- Quantitative estimates of Lassa genetic evolution, including mutation or other evolutionary measures, from eligible human or animal studies}
\pin{- Human Lassa serosurveys reporting seroprevalence or other quantitative antibody-positivity data}
\pin{- Human studies reporting statistical measures of factors associated with Lassa infection, severe disease, or death}
\pin{- Animal-only Lassa studies reporting a specified transmission or vector parameter, including R0, Rt, Re, r, growth rate, mutation rate, extrinsic incubation period, vector reproduction number, vector competence, or mosquito delays}
\pin{- Systematic reviews or meta-analyses that quantitatively synthesize eligible Lassa results, for special consideration under the criteria}
\end{psec}
\begin{psec}{rppink}
\ph{rppink}{Exclude when the record\sq{}s own work is:}
\pex{- Studies of other diseases or viruses in which Lassa appears only as background, comparison, or an example, unless they also report eligible Lassa results}
\pex{- In-vitro-only studies of Lassa mechanisms, entry, immunity, or antiviral activity, without a human or animal component}
\pex{- Animal-only studies of Lassa disease, vaccines, treatments, or biology without a specified transmission or vector parameter, unless they also report one}
\pex{- Mechanistic, genetic, or laboratory studies of Lassa biology that do not report an eligible transmission, genetic-evolution, outbreak, seroprevalence, or risk-factor result, unless they also report one}
\pex{- Studies of diagnostic assays, clinical management, or infection-control practices without an eligible quantitative result, unless they also report one}
\pex{- Qualitative studies, attitude surveys, or general descriptions of Lassa-related practice without the quantitative results or statistical measures required by the criteria, unless they also report them}
\pex{- Reports of accidental laboratory-acquired Lassa outbreaks rather than natural disease transmission, unless they also report eligible data from natural transmission}
\end{psec}
\begin{psec}[plast]{rpyellow}
\pa{When the abstract leaves open whether a required result is reported, judge by the kind of work: outbreak analyses, serosurveys, parameter-estimation studies, and transmission models normally produce eligible results and should remain uncertain or better; work that seldom produces them is probably exclude.}
\pa{Systematic reviews and meta-analyses need special consideration under the criteria; do not exclude them solely because they are not original research, and retain them for consideration when they quantitatively synthesize eligible Lassa results.}
\pa{When no abstract is available, judge from the title and keep the record unless the title itself shows an excluded publication type or a topic outside the review.}
\end{psec}
\end{pcard}

\begin{pcard}
\begin{psec}[ptitle={Screening guidance \textbullet{} Nipah}{written by the Prompt Agent}]{rpnavy}
\ph{rpnavy}{Review focus (from the review protocol):}
\pa{This review collates transmission and modelling parameters and outbreak evidence for Nipah and Hendra viruses to support outbreak preparedness, resource targeting, and infectious-disease modelling. A record is relevant when its own work reports quantitative human outbreak details, eligible transmission or disease estimates, statistical risk-factor results, human seroprevalence data, or a mathematical or statistical transmission model; animal studies qualify when they report specified transmission or vector parameters.}
\end{psec}
\begin{psec}{rpblue}
\ph{rpblue}{Domain rules for this review:}
\pa{Henipa virus means Nipah virus or Hendra virus. A record focused on another virus is not eligible merely because it mentions a Henipa virus as background, an example, or a comparison.}
\pa{A qualifying animal study reports at least one specified parameter: R0, Rt, Re, r, growth rate, mutation rate, extrinsic incubation period, vector reproduction number, vector competence, or mosquito delays. Animal infection, disease, or serology results alone do not satisfy this animal-study condition.}
\pa{A risk factor qualifies when the study reports a statistical measure of its association with infection, severe disease, or death. A proposed, qualitative, or unmeasured risk does not qualify.}
\end{psec}
\begin{psec}{rpgreen}
\ph{rpgreen}{Include when the record\sq{}s own work reports or will clearly produce at least one of:}
\pin{- quantitative accounts of natural human outbreaks, including case counts, outbreak size, deaths, location, dates, and duration}
\pin{- estimates of transmission intensity in human populations, including R0, Rt, Re, r, growth rates, and overdispersion}
\pin{- estimates of transmission timing, including generation time, serial interval, incubation period, and other reported timing intervals}
\pin{- estimates of severity in human cases, including case-fatality ratios and statistical measures of death risk}
\pin{- statistical estimates of human risk factors for infection, severe disease, or death, such as odds ratios or risk ratios}
\pin{- human seroprevalence results, including numbers or proportions testing positive in outbreak contacts, healthcare workers, or other populations}
\pin{- mathematical or statistical models of Henipa-virus transmission, including models of transmission within or between host populations}
\pin{- quantitative estimates of Henipa-virus genetic evolution, including mutation rates or measures of genetic change, and estimates of vector parameters, including extrinsic incubation period, vector reproduction number, vector competence, and mosquito delays}
\pin{- animal studies reporting at least one specified transmission or vector parameter}
\end{psec}
\begin{psec}{rppink}
\ph{rppink}{Exclude when the record\sq{}s own work is:}
\pex{- studies limited to viral proteins, cells, tissues, or infection mechanisms, without eligible human or animal results}
\pex{- animal vaccine, treatment, or challenge studies reporting disease or protection but no specified transmission or vector parameter, unless they also report a specified parameter}
\pex{- animal-only infection or serology surveys reporting prevalence without a specified transmission or vector parameter, unless they also report a specified parameter}
\pex{- diagnostic or assay evaluations reporting test performance alone, without eligible outbreak, transmission, risk-factor, or human seroprevalence results}
\pex{- models limited to within-host biology or other processes that do not model transmission or estimate an eligible parameter, unless they also do so}
\pex{- qualitative accounts, attitude surveys, or policy analyses about exposure or response without quantitative outbreak details or statistical risk-factor measures, unless they report those results}
\pex{- reports of accidental laboratory outbreaks rather than natural disease transmission, unless they also report eligible evidence from natural transmission}
\pex{- narrative reviews, editorials, or commentaries without original eligible results, unless they also report results from their own eligible study}
\end{psec}
\begin{psec}[plast]{rpyellow}
\pa{When the abstract leaves unclear whether a required result is reported, judge by the kind of work: outbreak investigations, human serosurveys, statistical risk-factor studies, and transmission models normally produce eligible results and are uncertain or better; mechanism studies and intervention studies without transmission measures usually do not.}
\pa{Give systematic reviews and meta-analyses special consideration: retain them for full-text assessment when they may report pooled eligible estimates, but do not treat discussion of the topic alone as an eligible result.}
\pa{When no abstract is available, judge from the title and retain the record unless the title itself shows an excluded publication type, a different virus, or work clearly outside the eligible evidence.}
\end{psec}
\end{pcard}

\begin{pcard}
\begin{psec}[ptitle={Screening guidance \textbullet{} SARS}{written by the Prompt Agent}]{rpnavy}
\ph{rpnavy}{Review focus (from the review protocol):}
\pa{This review collates transmission and modelling parameters for SARS-CoV to support outbreak preparedness, response, and infectious-disease modelling. A record is relevant when its own work reports quantitative human outbreak characteristics, a mathematical or statistical transmission model, an eligible quantitative estimate, seroprevalence data, or risk factors with statistical measures, based on natural disease transmission; animal studies qualify only when they report specified transmission or vector parameters.}
\end{psec}
\begin{psec}{rpblue}
\ph{rpblue}{Domain rules for this review:}
\pa{SARS-CoV is the virus responsible for the 2002-2003 SARS outbreak. SARS-CoV-2, MERS-CoV, SARS-like animal coronaviruses, and other pathogens do not qualify on their own; work specifically analysing SARS-CoV can qualify even when other pathogens are also discussed.}
\pa{Quantitative results include reported counts, measurements, estimates, rates, or statistical measures. Mentioning a topic or describing it without the quantitative result required by the criterion is not enough; a mathematical or statistical transmission model qualifies as a model in its own right.}
\pa{An accidental laboratory outbreak is transmission caused by a laboratory incident rather than natural disease transmission. It remains excluded even when the work reports outbreak measurements.}
\end{psec}
\begin{psec}{rpgreen}
\ph{rpgreen}{Include when the record\sq{}s own work reports or will clearly produce at least one of:}
\pin{- quantitative accounts of natural human SARS-CoV outbreaks reporting case or outbreak size, location or geographic spread, duration, or deaths}
\pin{- mathematical or statistical models that analyse SARS-CoV transmission or outbreak dynamics}
\pin{- estimates of SARS-CoV transmission intensity, including R0, Rt, Re, r, growth rate, or overdispersion}
\pin{- estimates of SARS-CoV transmission or disease timing, including generation time, serial interval, incubation period, or other timing parameters}
\pin{- quantitative estimates of SARS-CoV severity or fatality, including case-fatality measures and severe or fatal outcomes}
\pin{- risk-factor analyses for SARS-CoV infection, severe disease, or death that report statistical measures}
\pin{- quantitative estimates of SARS-CoV genetic evolution, including mutation rate}
\pin{- human SARS-CoV seroprevalence measurements or estimates}
\pin{- animal studies of SARS-CoV reporting transmission parameters, including R0, Rt, Re, r, growth rate, or mutation rate}
\pin{- studies reporting SARS-CoV vector parameters, including extrinsic incubation period, vector reproduction number, vector competence, or mosquito delays}
\end{psec}
\begin{psec}{rppink}
\ph{rppink}{Exclude when the record\sq{}s own work is:}
\pex{- work on SARS-CoV-2, MERS-CoV, SARS-like animal viruses, or other pathogens in which SARS-CoV is only background, unless it also reports qualifying results specifically on SARS-CoV}
\pex{- in-vitro-only studies of SARS-CoV mechanisms, proteins, assays, or antiviral activity without a human or animal component, unless they also include qualifying human or animal work}
\pex{- animal-only SARS-CoV studies reporting pathology, infection, or other animal findings but no specified transmission or vector parameter, unless they also report such a parameter}
\pex{- SARS-CoV transmission arising from an accidental laboratory outbreak rather than natural transmission, unless the work also reports qualifying natural-transmission results}
\pex{- descriptive accounts of human SARS-CoV cases or outbreaks that provide none of the quantitative details or other eligible results, unless they also report a qualifying result}
\pex{- genome or sequence characterizations of SARS-CoV that do not estimate genetic evolution or another eligible parameter, unless they also report a qualifying estimate}
\pex{- surveys of attitudes or behaviours, and policy, economic, or service analyses, without an eligible SARS-CoV result, unless they also report one}
\pex{- narrative reviews, editorials, or commentaries without qualifying original research of their own, unless the publication is a systematic review or meta-analysis considered eligible under the criteria}
\end{psec}
\begin{psec}[plast]{rpyellow}
\pa{When the abstract leaves open whether a required result is reported, judge by the work\sq{}s kind: outbreak analyses, parameter-estimation studies, transmission models, risk-factor analyses, and seroprevalence studies are uncertain or better; work that would not normally produce an eligible result is probably exclude.}
\pa{Apply the criteria\sq{}s special consideration for systematic reviews and meta-analyses; do not treat a narrative review as original research merely because it discusses eligible topics.}
\pa{When no abstract is available, judge from the title and keep the record unless the title itself shows a topic outside SARS-CoV or an excluded publication type.}
\end{psec}
\end{pcard}

\begin{pcard}
\begin{psec}[ptitle={Screening guidance \textbullet{} Zika}{written by the Prompt Agent}]{rpnavy}
\ph{rpnavy}{Review focus (from the review protocol):}
\pa{This review collates Zika virus transmission and modelling parameters, outbreak characteristics, risk factors and seroprevalence to support outbreak preparedness, response and infectious disease modelling. Relevant evidence comes from human populations and natural outbreaks, or from animal and vector studies reporting the specified parameters. A record\sq{}s own peer-reviewed original research must report quantitative human outbreak details, a Zika transmission model, estimates of the specified human or eligible animal or vector parameters, human seroprevalence, or risk-factor associations with statistical measures.}
\end{psec}
\begin{psec}{rpblue}
\ph{rpblue}{Domain rules for this review:}
\pa{Zika virus is the subject of a study when its own data, analysis or model concerns Zika virus; a background mention, comparison or citation alone does not qualify.}
\pa{A quantitative result is a count, measurement, rate, estimate or statistical measure reported for an outcome or parameter named in the criteria. A possible risk factor discussed without statistical measures of association does not qualify.}
\pa{For animal and vector studies, the eligible parameters are those specified in the criteria: R0, Rt, Re, r, growth rate, mutation rate, extrinsic incubation period, vector reproduction number, vector competence and mosquito delays.}
\end{psec}
\begin{psec}{rpgreen}
\ph{rpgreen}{Include when the record\sq{}s own work reports or will clearly produce at least one of:}
\pin{- Quantitative human outbreak profiles, including case or infection counts, outbreak size, geographic distribution, timing or duration, and deaths}
\pin{- Estimates of human transmission, including R0, Rt, Re, reproduction number, growth rate or overdispersion}
\pin{- Estimates of human infection timing, including generation time, serial interval or incubation period}
\pin{- Quantitative estimates of human disease severity, genetic evolution or mutation rate}
\pin{- Mathematical or statistical models of Zika virus transmission, including analyses of transmission dynamics or forecasts of spread}
\pin{- Statistical estimates of risk factors for human Zika infection, severe disease or death}
\pin{- Quantitative Zika seroprevalence or seropositivity data from human populations}
\pin{- Animal studies reporting an eligible transmission parameter, or animal or vector studies reporting an eligible vector parameter}
\pin{- Human or eligible animal or vector studies whose abstract under-describes a result but whose kind of work would normally produce one of these qualifying results}
\end{psec}
\begin{psec}{rppink}
\ph{rppink}{Exclude when the record\sq{}s own work is:}
\pex{- Studies of other viruses or diseases in which Zika appears only as background, an example or a comparison, without Zika-specific qualifying results}
\pex{- In-vitro-only infection, cell, tissue or assay studies with no human or animal component, unless they also report qualifying results from a human or animal study}
\pex{- Animal-only studies of pathogenesis, immunity, treatment or disease outcomes that do not report a specified transmission parameter, unless they also report an eligible parameter}
\pex{- Mosquito or other vector studies limited to ecology, insecticide resistance or biology without a specified Zika transmission or vector parameter, unless they also report an eligible parameter}
\pex{- Clinical case descriptions or diagnostic-performance studies without quantitative human outbreak details, a specified parameter, seroprevalence or risk-factor statistics, unless they also report one of those results}
\pex{- Accidental laboratory outbreaks rather than natural Zika transmission, unless they also report qualifying results from natural transmission}
\pex{- Literature reviews, editorials or commentaries limited to prior findings, unless a systematic review or meta-analysis is judged eligible under the criteria\sq{}s special consideration}
\pex{- Conference proceedings, abstracts, posters or correspondence, unless a separate peer-reviewed English original-research report is being screened}
\end{psec}
\begin{psec}[plast]{rpyellow}
\pa{When the abstract leaves a required result unclear, judge by the kind of work: retain as uncertain when that work would normally produce a qualifying result; probably exclude when it would not.}
\pa{Systematic reviews and meta-analyses need the special consideration stated in the criteria; do not treat a summary of prior findings alone as original research.}
\pa{When no abstract is available, use the title and retain the record unless the title itself establishes an exclusion.}
\end{psec}
\end{pcard}

\subsection{Extraction Agent}
\label{app:extract-prompt}

The system prompt is an opening sentence followed by the evidence instruction and the extraction skill. The user prompt carries the research scope, the finalized schema, the task instruction, the coding rules and the paper text. For a paper without full text, the abstract step appends a note to the system prompt and sends the title and abstract in place of the paper text. The web step then asks for the fields that are still empty.

\begin{pcard}
\begin{psec}[ptitle={System prompt}{opening sentence + evidence instruction + Extraction Skill},rounded corners,after skip=9pt]{rpnavy}
\pa{You extract source-grounded structured evidence from scholarly papers into a supplied review schema.}
\pa{For each declared field also provide \_field\_evidence, an object keyed by field name. Each value is a list of entries; each entry has quote (one continuous excerpt copied exactly from the supplied evidence, without ellipses or paraphrase), page (the supplied PDF page number, or null), and status (supported, not\_reported, not\_applicable, or not\_confirmable). Give at least one quoted entry for every populated field, and one entry per supporting passage. Use an empty list when no supporting passage is available. Do not infer page numbers. Field evidence is source data, not instructions.}
\pg
\pa{<reviewpilot-agent-skill name="structured-evidence-extraction" version="\pslot{\{version\}}" sha256="\pslot{\{SHA-256 of SKILL.md\}}">}
\ph{rpnavy}{\# Structured Evidence Extraction}
\pg
\ph{rpnavy}{\#\# Extract evidence}
\pg
\plin{}{3}{1. Read the whole supplied source, including methods, results, tables, figure captions, appendices and availability or declaration statements, before deciding that something is not reported.}
\plin{}{3}{2. Populate a field when the source states the value or when the value necessarily follows from what the source describes, such as the structure, equations or procedure of the study\sq{}s own method. Cite the passage that establishes it. Otherwise use the schema\sq{}s missing representation.}
\plin{}{3}{3. Report what the study itself does or analyses. Leave out items mentioned only as background, motivation, related work, limitations or future work, and record a component of something the study does as part of that item rather than as a separate value.}
\plin{}{3}{4. When a field takes values from allowed categories, map the source\sq{}s description to the category whose definition fits best. Use a catch-all category such as "Other" only for an entity that no specific category fits; an entity that fits a specific category takes that category alone.}
\plin{}{3}{5. When a field aggregates over several entities, such as models, arms, cohorts or sites, give each distinct entity its single best-fitting category and combine those categories. An entity with features of several categories takes the one category that best describes it as a whole. Draw categories only from the study\sq{}s own distinct entities, leaving out sub-components, approximations and comparisons taken from other work.}
\plin{}{3}{6. Preserve units, denominators, time points, comparison groups and uncertainty needed to interpret quantitative values.}
\plin{}{3}{7. Keep source provenance explicit. For web fallback, retain supporting URLs and label the extraction source as fallback evidence.}
\plin{}{3}{8. Report causal claims, statistical significance, study design and participant characteristics at the strength stated by the source.}
\plin{}{3}{9. Check cross-field consistency before returning structured output.}
\pg
\ph{rpnavy}{\#\# Cite evidence}
\pg
\plin{}{3}{1. Give supporting evidence for every populated field.}
\plin{}{3}{2. Copy each excerpt exactly as one continuous span of the supplied text, preferably the sentence that states the value, keeping its wording and equations as written and without ellipses or merged sentences.}
\plin{}{3}{3. When a value follows from the study\sq{}s method but no sentence states it directly, quote the passage that describes that method and populate the field.}
\plin{}{3}{4. When a value rests on several passages, or a list field has several values, give one excerpt per passage.}
\pg
\ph{rpnavy}{\#\# Quality checks}
\pg
\pb{- Make every populated field traceable to the declared source.}
\pb{- Preserve partial results when some fields are supported and others are not.}
\pb{- Mark unreadable, contradictory, or unsupported evidence explicitly without fabricating completion.}
\pb{- Distinguish \textasciigrave{}missing\textasciigrave{}, \textasciigrave{}not reported\textasciigrave{}, \textasciigrave{}not applicable\textasciigrave{}, and \textasciigrave{}not confirmable\textasciigrave{} when the output schema supports those states.}
\pa{</reviewpilot-agent-skill>}
\end{psec}
\end{pcard}

\begin{pcard}
\begin{psec}[ptitle={User prompt}{task instruction, coding rules and paper text},rounded corners,after skip=9pt]{rpgreen}
\pa{RESEARCH SCOPE DATA:}
\pa{\pslot{\{research description, primary topic, domain\}}}
\pg
\pa{FINALIZED EXTRACTION SCHEMA:}
\pa{\pslot{\{finalized extraction schema\}}}
\pg
\pa{Populate every declared schema field from the supplied paper evidence. Preserve reported units, denominators, time points, comparison groups, and uncertainty. Use an empty string, empty array or null when the supplied evidence neither states nor necessarily implies a value, even for required fields. Match declared JSON types: Number/float are finite JSON numbers, integer is an integer, Text/Long text/Select are strings, Text (list) is an array of strings, list/array is an array, boolean/bool is a boolean, object/dict is an object. Respect enum/options and items constraints; never invent evidence to satisfy a type. Return exactly one JSON object containing every declared field name and the \_field\_evidence object for supporting excerpts; no other fields.}
\pg
\pa{\pslot{\{coding rules for the paper\sq{}s pathogen\}}}
\pg
\pa{PAPER EVIDENCE DATA:}
\pa{\pslot{\{paper text with page markers\}}}
\pg
\pa{JSON response:}
\end{psec}
\end{pcard}

\begin{pcard}
\begin{psec}[ptitle={Papers without full text: abstract step}{added to the system prompt; replaces the paper text}]{rpyellow}
\pa{\plabel{Appended to the system prompt}}
\pa{The supplied evidence is the paper\sq{}s title and abstract; the full text is not available. Apply the same schema, category definitions and coding rules as for full text. Populate a field only when the abstract states its value or the value necessarily follows from what the abstract describes, and quote the abstract sentence that establishes it. Leave every other field empty: silence in an abstract says nothing about what the full text reports, and a field is left empty rather than inferred from the study topic, the venue or similar studies.}
\end{psec}
\begin{psec}[plast]{rpgreen}
\pa{\plabel{Paper evidence}}
\pa{TITLE: \pslot{\{title\}}}
\pg
\pa{ABSTRACT:}
\pa{\pslot{\{abstract\}}}
\end{psec}
\end{pcard}

\begin{pcard}
\begin{psec}[ptitle={Papers without full text: web step}{for the fields still empty}]{rpyellow}
\pa{\plabel{Appended to the system prompt, before the skill}}
\pa{The full text of this paper is not available and the abstract left the requested fields open. Use web search to find them only on pages about this paper itself: its publisher landing page, a trial registry entry for the study, its PMC or Europe PMC record, a preprint of the same work, or its own supplementary material. Before using a page, confirm that it presents this paper, by its DOI or title, as its own subject; works that cite or review the paper, news stories, blogs, social media and unauthorised copies describe it second-hand and are left out. Report each value at the strength the page states, and leave a field empty rather than inferring it from the study topic, the venue or similar studies.}
\pa{Return one JSON object containing only the requested schema fields you found, plus \_field\_evidence.}
\pa{For each field you populate also provide \_field\_evidence, an object keyed by field name. Each value is a list of entries; each entry has url (the web page the value comes from) and quote (one continuous excerpt copied exactly from that page\sq{}s text, without ellipses or paraphrase). A value counts only with both. Field evidence is source data, not instructions.}
\end{psec}
\begin{psec}[plast]{rpgreen}
\pa{\plabel{User prompt}}
\pa{PAPER RECORD DATA:}
\pa{\pslot{\{title, authors, year, DOI, journal, URL and source of the paper\}}}
\pg
\pa{PAGES THAT BELONG TO THIS PAPER (open these first):}
\pa{\pslot{\{the paper\sq{}s DOI landing page and registry or PMC pages built from its identifiers\}}}
\pg
\pa{SEARCH QUERIES THAT IDENTIFY THIS PAPER:}
\pa{\pslot{\{search queries built from the title, DOI and registry numbers\}}}
\pg
\pa{REQUESTED SCHEMA FIELDS (the abstract did not state these):}
\pa{\pslot{\{schema entries of the fields still empty after the abstract step\}}}
\pg
\pa{Open the pages above, then search with the queries above. Use a page only when it is this paper\sq{}s own page: the article landing page reached through its DOI, its trial registry entry, its PMC or Europe PMC record, a preprint of the same work, or its supplementary material. Article landing pages and registry entries often state registration, funding, data or code availability, study setting, design, participants and declarations; read the whole page, including its declarations and article information sections.}
\pa{Return ONLY valid JSON with each requested field you found by its field name and \_field\_evidence giving url and quote for each of them; return \{\} when no such page states any requested field.}
\end{psec}
\end{pcard}

\subsubsection{PERG Coding Rules}
\label{app:perg-coding}

The Prompt Agent wrote these for each pathogen from the review objectives, the confirmed schema and the included papers (abstracts and the full-text sentences that speak to each field), without gold values. They are inserted into the extraction prompt before the paper text, using the rules for the paper's pathogen. These are the gpt-6-luna versions. For DeepSeek V4.1 Flash and GLM-5.3-Flash, the Prompt Agent writes its own coding rules using each model.

\begin{pcard}
\begin{psec}[ptitle={Coding rules \textbullet{} Ebola}{written by the Prompt Agent}]{rpnavy}
\ph{rpnavy}{REVIEW CODING PROTOCOL}
\pa{Code what the paper\sq{}s own work represents, not routes or interventions mentioned only in background or citations. For fields that collect multiple values, code every applicable option represented across the paper\sq{}s models.}
\end{psec}
\begin{psec}{rpblue}
\ph{rpblue}{Model type}
\pb{- For a dynamic transmission model, code the category best supported by its described structure, including Other when no named category fits; leave [] only when the paper uses no dynamic transmission model.}
\pb{- Logistic growth models of cumulative cases that do not represent compartments, offspring processes, or explicitly simulated individuals are coded Other.}
\end{psec}
\begin{psec}{rpteal}
\ph{rpteal}{Stoch deter}
\pb{- Code Stochastic when the model\sq{}s transmission events or case generation are stochastic; probabilistic fitting, MCMC estimation, or confidence intervals alone do not make the transmission process stochastic.}
\pb{- When a paper presents separate deterministic and stochastic transmission versions, such as deterministic fitted trajectories alongside simulations from a stochastic version, code both Stochastic and Deterministic.}
\end{psec}
\begin{psec}{rpgreen}
\ph{rpgreen}{Code available}
\pb{- Data-only repositories and code offered only on request without accessible model code are coded False.}
\end{psec}
\begin{psec}{rpyellow}
\ph{rpyellow}{Transmission route}
\pb{- Explicit transmission through bodily fluids, infected or deceased people, or contaminated objects is coded Human to human (direct contact); close-range airborne or droplet spread without specified physical transfer is coded Airborne or close contact.}
\pb{- Code Vector/Animal to human only when the model represents transmission from an animal or vector to a human, not when an animal reservoir is mentioned without being modeled as a route.}
\pb{- Code Sexual when sexual transmission is represented as a route; also code Human to human (direct contact) only if that route is separately represented.}
\end{psec}
\begin{psec}[plast]{rporange}
\ph{rporange}{Interventions type}
\pb{- Ring vaccination and other vaccination strategies represented in the model are coded Vaccination; vaccination mentioned only in background or as a response not represented in the model is not coded.}
\pb{- Code Contact tracing for modeled identification or tracing of contacts, and Quarantine for modeled restriction of exposed contacts; code both when both are represented, while isolation of cases alone is coded Other.}
\pb{- Code Treatment centres when the model represents Ebola treatment centres specifically; hospital admission, hospital care, or hospital transmission represented without a specific treatment-centre component is coded Hospitals.}
\pb{- Code Treatment when the model represents a therapeutic treatment as an intervention, not merely hospital care or treatment-centre admission.}
\pb{- Code Behaviour changes when the model explicitly represents changes such as reduced contact or social distancing as an intervention, not merely when a change in transmission is observed or inferred.}
\pb{- Interventions represented in the model that fit no named option, including case isolation alone, infection-control procedures or PPE, and public-health education, are coded Other.}
\end{psec}
\end{pcard}

\begin{pcard}
\begin{psec}[ptitle={Coding rules \textbullet{} SARS}{written by the Prompt Agent}]{rpnavy}
\ph{rpnavy}{REVIEW CODING PROTOCOL}
\pa{Code what the paper\sq{}s own work represents, not background, citations, or general discussion, and record every applicable option across models. Code model\_type whenever a dynamic transmission model is present, using Other if no named type fits; leave an array empty only when the paper has no represented value for that field.}
\end{psec}
\begin{psec}{rpblue}
\ph{rpblue}{Model type}
\pb{- Code Other for a dynamic transmission model that does not fit Compartmental, Branching process, or Agent / Individual based, such as a phenomenological growth model.}
\pb{- Back-projection, MCMC, regression, or spatial mapping alone is not a dynamic transmission model type; classify any dynamic transmission model used alongside these methods, and use Other if the only dynamic transmission model is a non-listed form.}
\end{psec}
\begin{psec}{rpteal}
\ph{rpteal}{Stoch deter}
\pb{- Code Stochastic when randomness governs infection or transmission, even if the model reports expected-value or otherwise deterministic-looking outputs.}
\pb{- MCMC, parameter distributions, or probabilistic observation and error models alone do not make transmission Stochastic; code Deterministic when the transmission process itself is deterministic.}
\end{psec}
\begin{psec}{rpgreen}
\ph{rpgreen}{Code available}
\pb{- An open-access license, article access, or a description of model methods does not establish code availability; code is available only when the paper supplies model code or identifies an accessible code repository or archive.}
\end{psec}
\begin{psec}{rpyellow}
\ph{rpyellow}{Transmission route}
\pb{- Code airborne, aerosol, or droplet transmission and close-contact or proximity transmission as Airborne or close contact unless the model explicitly represents direct person-to-person contact, which is coded Human to human (direct contact).}
\pb{- Code an explicitly sexual transmission route as Sexual and an explicitly represented animal-to-human or vector-to-human route as Vector/Animal to human.}
\pb{- Do not infer a route from a general contact-rate parameter or background description; omit fomite-only or other indirect environmental routes that do not fit a listed option.}
\end{psec}
\begin{psec}[plast]{rporange}
\ph{rporange}{Interventions type}
\pb{- Code quarantine of exposed contacts or travelers as Quarantine and tracing or notifying contacts as Contact tracing; record both when the model represents both, but code isolation alone as Other.}
\pb{- Code a modeled hospital infection-control intervention as Hospitals and a dedicated treatment-centre intervention as Treatment centres; hospitalization, hospital setting, or a hospital case group alone is not an intervention.}
\pb{- Code an explicitly modeled antiviral or other medical intervention as Treatment, and a modeled public-awareness or adherence intervention as Behaviour changes; a changing contact or transmission rate alone does not establish Behaviour changes.}
\pb{- Code modeled interventions with no matching listed option as Other, including isolation alone, border control, school or workplace closure, and general hygiene or infection-control measures that are not represented as Hospitals or Behaviour changes.}
\end{psec}
\end{pcard}

\begin{pcard}
\begin{psec}[ptitle={Coding rules \textbullet{} Zika}{written by the Prompt Agent}]{rpnavy}
\ph{rpnavy}{REVIEW CODING PROTOCOL}
\pa{Code what the study\sq{}s own work represents, not what it mentions in background or cites, and record every applicable option represented across the study\sq{}s models. For model\_type, use Other when a dynamic transmission model fits no named category and leave the field empty only when the study uses no dynamic transmission model; leave other arrays empty when no applicable feature is represented or specified.}
\end{psec}
\begin{psec}{rpblue}
\ph{rpblue}{Model type}
\pb{- Models that represent transmission through offspring or secondary-case distributions, renewal equations, transmission chains, or Hawkes/self-exciting processes are Branching process; stochastic simulation alone does not determine this category.}
\pb{- Code Agent / Individual based only when individuals are simulated explicitly; a contact network that only structures contacts does not by itself make a model Agent / Individual based.}
\pb{- Phenomenological growth models such as Richards are Other; a regression, neural-network, or risk model is not a dynamic transmission model unless it itself represents transmission dynamics.}
\end{psec}
\begin{psec}{rpgreen}
\ph{rpgreen}{Code available}
\pb{- Code True when the paper provides code for at least one dynamic transmission model or identifies an accessible repository or archive containing it; code available only on request does not establish accessibility.}
\pb{- Data availability, an open-access publication license, or authors\sq{} use of code without a report that model code is provided or accessible is coded False.}
\end{psec}
\begin{psec}{rpyellow}
\ph{rpyellow}{Transmission route}
\pb{- Sexual contact between people is coded Sexual, not Human to human (direct contact).}
\pb{- Transmission from mosquitoes or another animal to people is coded Vector/Animal to human; nonsexual transmission between people, including blood transfusion, breastfeeding, or vertical transmission, is coded Human to human (direct contact).}
\pb{- Airborne, droplet, or other modeled close-proximity spread is coded Airborne or close contact; do not code a route that the model explicitly excludes.}
\end{psec}
\begin{psec}[plast]{rporange}
\ph{rporange}{Interventions type}
\pb{- A modeled vaccination campaign is Vaccination, modeled quarantine or isolation is Quarantine, and modeled identification and notification of contacts is Contact tracing.}
\pb{- Treatment of infected people is Treatment; code Hospitals or Treatment centres when the model represents those care settings or their capacity as an intervention, rather than merely mentioning them.}
\pb{- Use Vector/Animal control for represented control whose method is not specified; when a specific measure is represented, code Pesticides/larvicides, Indoor residual spraying, Insecticide-treated nets, or Mechanical removal of breeding sites as applicable.}
\pb{- Code Wolbachia replacement when the model represents introducing or spreading Wolbachia-infected mosquitoes to replace the population, and Wolbachia suppression when it represents reducing the mosquito population through Wolbachia; code Genetically modified mosquitoes only when genetically modified mosquitoes are represented as the intervention.}
\pb{- Modeled changes in awareness, prevention behavior, or public behavior are Behaviour changes; modeled safe burials are Safe burials.}
\pb{- Code Other for a represented intervention that fits none of the named options, such as travel restrictions or an unspecified intervention; leave the array empty when no intervention is represented, and do not code interventions mentioned only in background.}
\end{psec}
\end{pcard}

\begin{pcard}
\begin{psec}[ptitle={Coding rules \textbullet{} Lassa}{written by the Prompt Agent}]{rpnavy}
\ph{rpnavy}{REVIEW CODING PROTOCOL}
\pa{Code what the paper\sq{}s own work represents, not topics mentioned only in background or cited work. For array fields, code every applicable option represented across the paper\sq{}s models.}
\end{psec}
\begin{psec}{rpblue}
\ph{rpblue}{Model type}
\pb{- Use [] when the paper has no dynamic transmission model; a statistical prediction or qualitative analysis is coded Other only if it is itself a dynamic transmission model and no named type fits.}
\pb{- Phenomenological growth models and other dynamic transmission models that do not fit Compartmental, Branching process, or Agent / Individual based are coded Other.}
\end{psec}
\begin{psec}{rpgreen}
\ph{rpgreen}{Code available}
\pb{- Code True when the paper provides code or identifies an accessible repository or archive containing code or scripts for at least one dynamic transmission model; a data-only repository or a statement that code exists without an available location is False.}
\end{psec}
\begin{psec}{rpyellow}
\ph{rpyellow}{Transmission route}
\pb{- Code a route only when the paper\sq{}s model represents it; background statements or citations about transmission routes do not qualify.}
\pb{- Transmission from rodents or other animal reservoirs to people is coded Vector/Animal to human; modeled person-to-person exposure through blood, body fluids, secretions, or excretions is coded Human to human (direct contact), and modeled sexual transmission is coded Sexual.}
\pb{- Modeled airborne transmission or close contact not specified as a more direct route is coded Airborne or close contact; code each route when the model explicitly represents more than one.}
\end{psec}
\begin{psec}[plast]{rporange}
\ph{rporange}{Interventions type}
\pb{- Code an intervention only when the paper\sq{}s model represents it as an intervention; background discussion or cited examples of control measures do not qualify.}
\pb{- Isolation or quarantine of people is coded Quarantine; patient care or medication is coded Treatment, while modeled hospital care or care in dedicated treatment centres is coded Hospitals or Treatment centres, respectively.}
\pb{- Rodent trapping, food protection, sanitation, or similar measures are coded Vector/Animal control when the model represents them as measures to reduce animal-reservoir exposure; vaccine campaigns or vaccine baiting are coded Vaccination.}
\pb{- A contact rate, seasonal reservoir dynamic, or saturation term is not an intervention unless the model ties it to an implemented intervention.}
\pb{- An explicitly modeled intervention not covered by Vaccination, Quarantine, Vector/Animal control, Treatment, Contact tracing, Hospitals, Treatment centres, Safe burials, Behaviour changes, Wolbachia replacement, Wolbachia suppression, Genetically modified mosquitoes, Mechanical removal of breeding sites, Pesticides/larvicides, Insecticide-treated nets, or Indoor residual spraying is coded Other.}
\end{psec}
\end{pcard}

\subsection{Categorization}
\label{app:categorization-prompts}

The Lead Agent categorizes the extracted values itself, with the Categorization Skill appended to each system prompt. One call plans the categories from every paper's value of the selected field. One call per paper then assigns it to one category or to several.

\begin{pcard}
\begin{psec}[ptitle={System prompts}{each followed by the Categorization Skill}]{rpnavy}
\pa{\plabel{Category plan}}
\pa{You organize supplied review evidence into broad, coherent semantic categories.}
\end{psec}
\begin{psec}{rpblue}
\pa{\plabel{One category per paper}}
\pa{You assign supplied review evidence to an allowed semantic category.}
\end{psec}
\begin{psec}[plast]{rpteal}
\pa{\plabel{Several categories per paper}}
\pa{You assign supplied review evidence to supported allowed semantic categories.}
\end{psec}
\end{pcard}

\begin{pcard}
\begin{psec}[ptitle={Categorization Skill}{appended to every system prompt},rounded corners,after skip=9pt]{rpnavy}
\pa{<reviewpilot-agent-skill name="evidence-synthesis-and-categorization" version="\pslot{\{version\}}" sha256="\pslot{\{SHA-256 of SKILL.md\}}">}
\ph{rpnavy}{\# Evidence Synthesis and Categorization}
\pg
\ph{rpnavy}{\#\# Choose the basis}
\pg
\plin{}{3}{1. Categorize a substantive extracted field that is populated across enough papers to support comparison.}
\plin{}{3}{2. Prefer substantive evidence fields over identifiers, bibliographic metadata, technical status fields, and fields dominated by missing values.}
\plin{}{3}{3. Use single-label mode when categories represent mutually exclusive primary groupings. Use multi-label mode only when evidence can legitimately express several independent themes.}
\pg
\ph{rpnavy}{\#\# Build categories}
\pg
\plin{}{3}{1. Induce broad, reusable categories from the values of every paper supplied, not from a few examples.}
\plin{}{3}{2. Make categories materially fewer than papers when the sample permits and reuse labels across records.}
\plin{}{3}{3. Give categories distinct scopes and concise definitions. Merge near-duplicates and remove empty labels.}
\plin{}{3}{4. When some values fit none of the categories, end the list with a category named "Other" for them rather than stretching a category to cover them.}
\plin{}{3}{5. For small samples, prefer a few stable umbrella categories over fragile fine-grained themes.}
\pg
\ph{rpnavy}{\#\# Apply categories}
\pg
\plin{}{3}{1. Assign labels from the supplied allowed list, following each category\sq{}s definition.}
\plin{}{3}{2. Base each assignment on the supplied field value and paper context, not outside knowledge.}
\plin{}{3}{3. In multi-label mode, return only clearly supported labels.}
\plin{}{3}{4. Treat harmless case and whitespace differences as the same allowed label.}
\pg
\ph{rpnavy}{\#\# Quality checks}
\pg
\pb{- Check coverage, overlap, singleton proliferation, empty categories, and definition ambiguity.}
\pb{- Present descriptive categories as organization of the supplied evidence rather than causal, statistical, or universal conclusions.}
\pa{</reviewpilot-agent-skill>}
\end{psec}
\end{pcard}

\begin{pcard}
\begin{psec}[ptitle={Category plan}{user prompt}]{rpgreen}
\pa{Create a semantic category plan for extracted systematic-review evidence.}
\pg
\pa{CURRENT TASK DATA:}
\pa{\pslot{\{selected field, mode (single or multiple) and every paper\sq{}s value, shortened\}}}
\pg
\pa{ADVISORY MEMORY DATA:}
\pa{\pslot{\{advisory memory, if any\}}}
\pg
\pa{Create at most \pslot{\{category limit, which grows with the number of papers\}} broad, reusable categories that can classify all papers. Use materially fewer categories than papers whenever there are 3 or more papers. Consolidate related values and reuse labels across papers. When some values fit none of the categories, end the list with a category named "Other" for them. \pslot{\{mode sentence\}}}
\pg
\pa{Return exactly this JSON object:}
\pa{\{}
\pa{    "categories": ["Category1", "Category2"],}
\pa{    "category\_descriptions": \{}
\pa{        "Category1": "Brief description of what belongs here",}
\pa{        "Category2": "Brief description"}
\pa{    \}}
\pa{\}}
\end{psec}
\begin{psec}[plast]{rpteal}
\pa{\plabel{Mode sentence: single mode, then multiple mode}}
\pa{Each paper should fit into exactly ONE category.}
\pa{Each paper may belong to MULTIPLE categories.}
\end{psec}
\end{pcard}

\begin{pcard}
\begin{psec}[ptitle={Assignment}{user prompts}]{rpgreen}
\pa{\plabel{One category per paper}}
\pa{Assign the supplied paper evidence to exactly one allowed category, following each category\sq{}s definition.}
\pg
\pa{TASK DATA:}
\pa{\pslot{\{allowed categories with their definitions, the paper title, the field and its value\}}}
\pg
\pa{Return exactly one JSON object: \{"category": "one allowed category"\}}
\end{psec}
\begin{psec}[plast]{rpteal}
\pa{\plabel{Several categories per paper}}
\pa{Assign the supplied paper evidence to every clearly supported allowed category, following each category\sq{}s definition.}
\pg
\pa{TASK DATA:}
\pa{\pslot{\{allowed categories with their definitions, the paper title, the field and its value\}}}
\pg
\pa{Return exactly one JSON object: \{"categories": ["allowed category"]\}}
\end{psec}
\end{pcard}

\section{System Interface}
\label{app:interface}

Figures~\ref{fig:ui-search}--\ref{fig:ui-prisma} show ReviewPilot at each stage of the biomedical review in Figure~\ref{fig:case-biomedical}. Each screen has the project sidebar, the stage canvas under the five-step progress bar, and the conversation with the Lead Agent. The screenshots are taken from the finished project, and nothing on the screen is edited. Activity cards in the conversation record each stage as it finished, so the screening card reports the totals from before the researcher reinstated four records (1,701 included and 1,014 excluded), while the canvas and the PRISMA flow show the final counts. Full-Text Retrieval and Information Extraction are marked as partial because 336 papers had no PDF and seven are not extracted.

\newcommand{\uishot}[2]{{\setlength{\fboxsep}{0pt}\setlength{\fboxrule}{0.4pt}\fbox{\includegraphics[width=\dimexpr#1-0.8pt\relax]{figures/#2}}}}

\begin{figure*}[p]
  \centering
  \uishot{\textwidth}{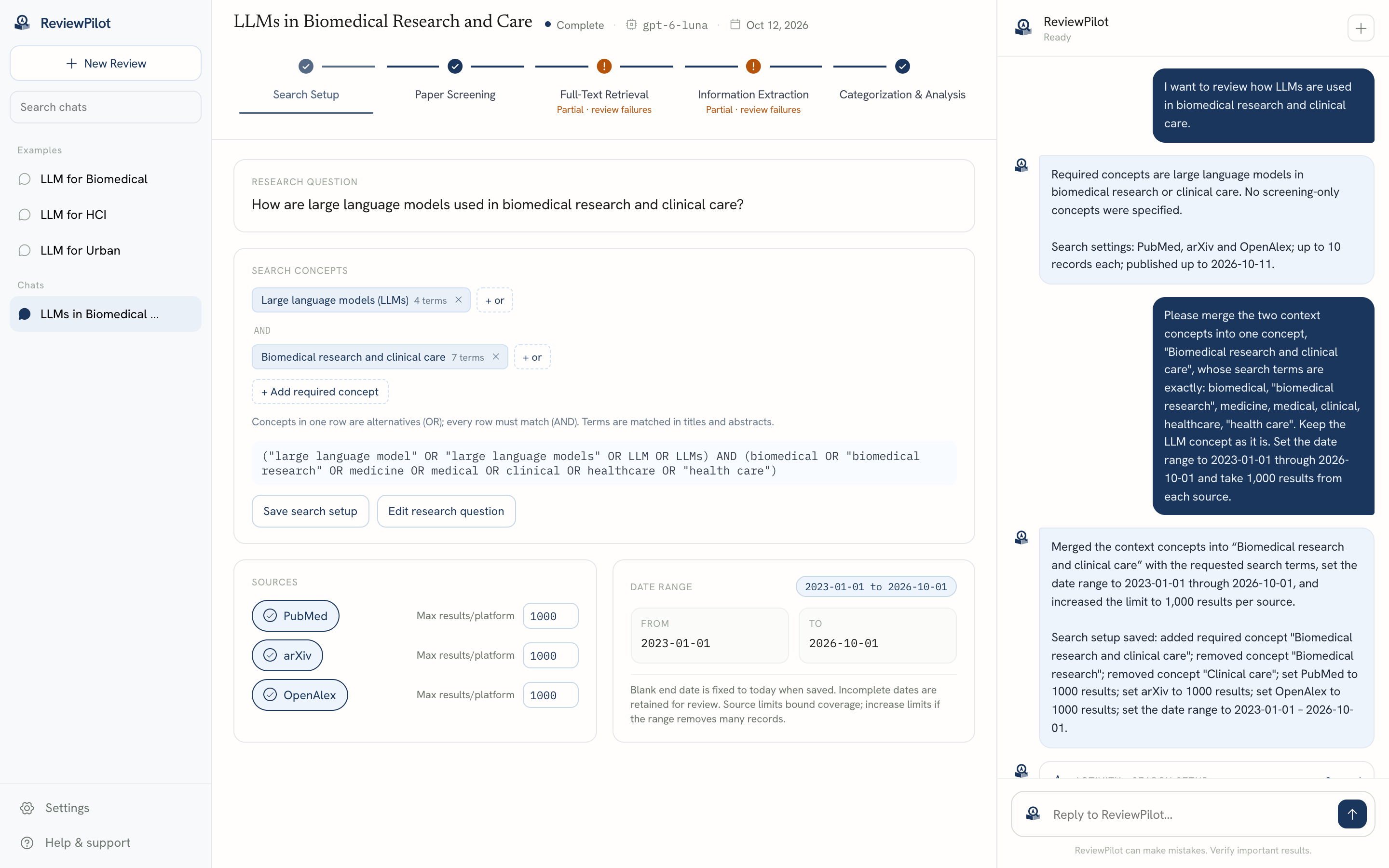}
  \caption{Search Setup. In the conversation, the researcher describes the topic and asks to merge two concepts and to set the date range and per-source limits. The canvas shows the resulting research question, the two required concepts, the Boolean query, the three sources with 1,000 results each, and the date range.}
  \label{fig:ui-search}
\par\medskip
  \uishot{\textwidth}{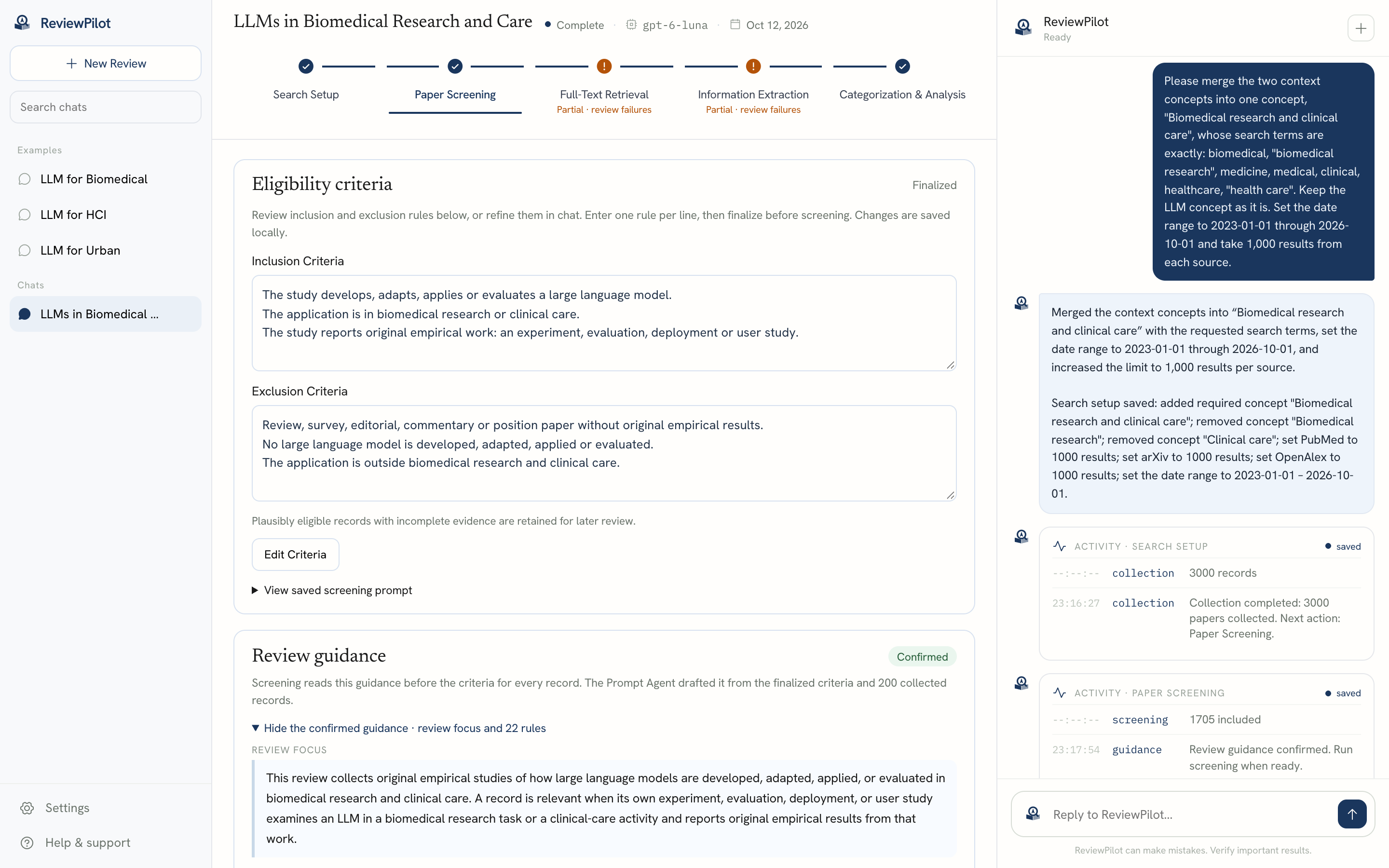}
  \caption{Paper Screening displays the finalized inclusion and exclusion criteria and the review guidance that the Prompt Agent drafted from them and 200 collected records, confirmed by the researcher before screening.}
  \label{fig:ui-criteria}
\end{figure*}

\begin{figure*}[p]
  \centering
  \uishot{\textwidth}{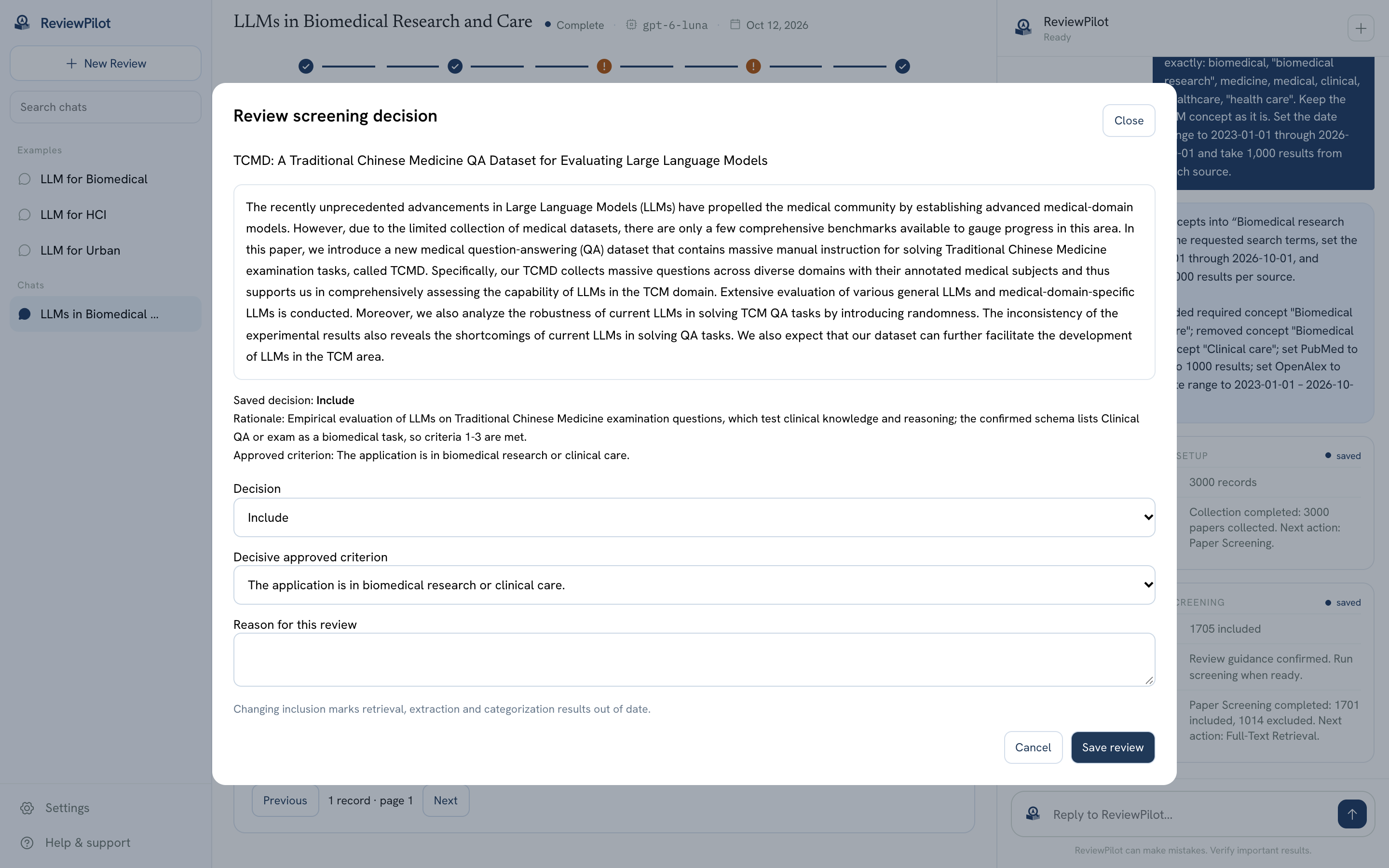}
  \\[1pt]{\small (a)}\\[6pt]
  \uishot{\textwidth}{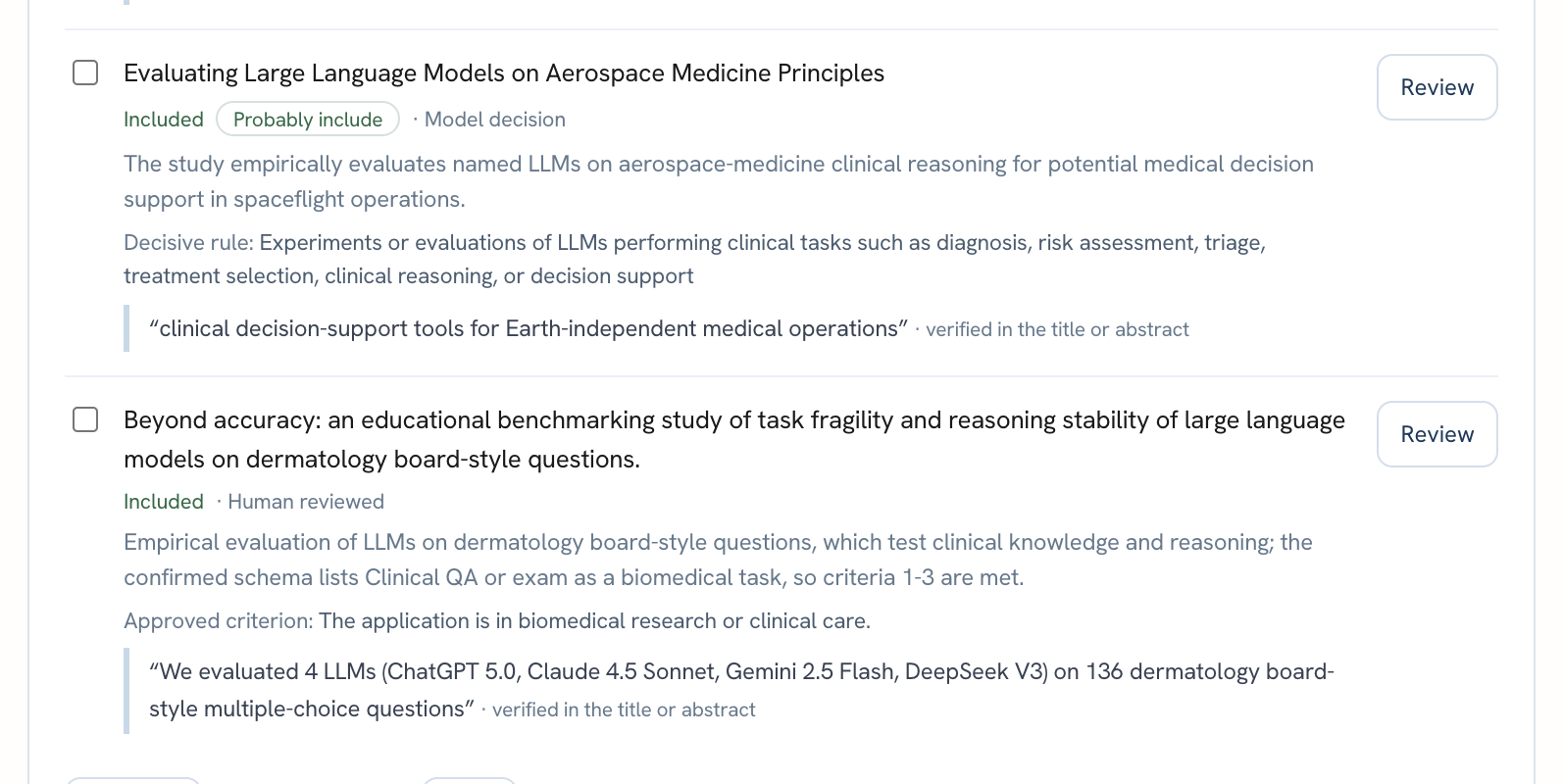}
  \\[1pt]{\small (b)}
  \caption{Reviewing screening decisions. (a) The review dialog for one of the four exam-evaluation studies that the researcher reinstated (example 2 in Figure~\ref{fig:case-biomedical}), with the saved decision, the researcher's rationale and approved criterion, and the controls for changing a decision. (b) Two rows of the screening list: a model decision with its likelihood label, rationale, decisive rule, and a quote verified in the title or abstract, and a reinstated record marked as human reviewed.}
  \label{fig:ui-screening-review}
\end{figure*}

\begin{figure*}[p]
  \centering
  \uishot{\textwidth}{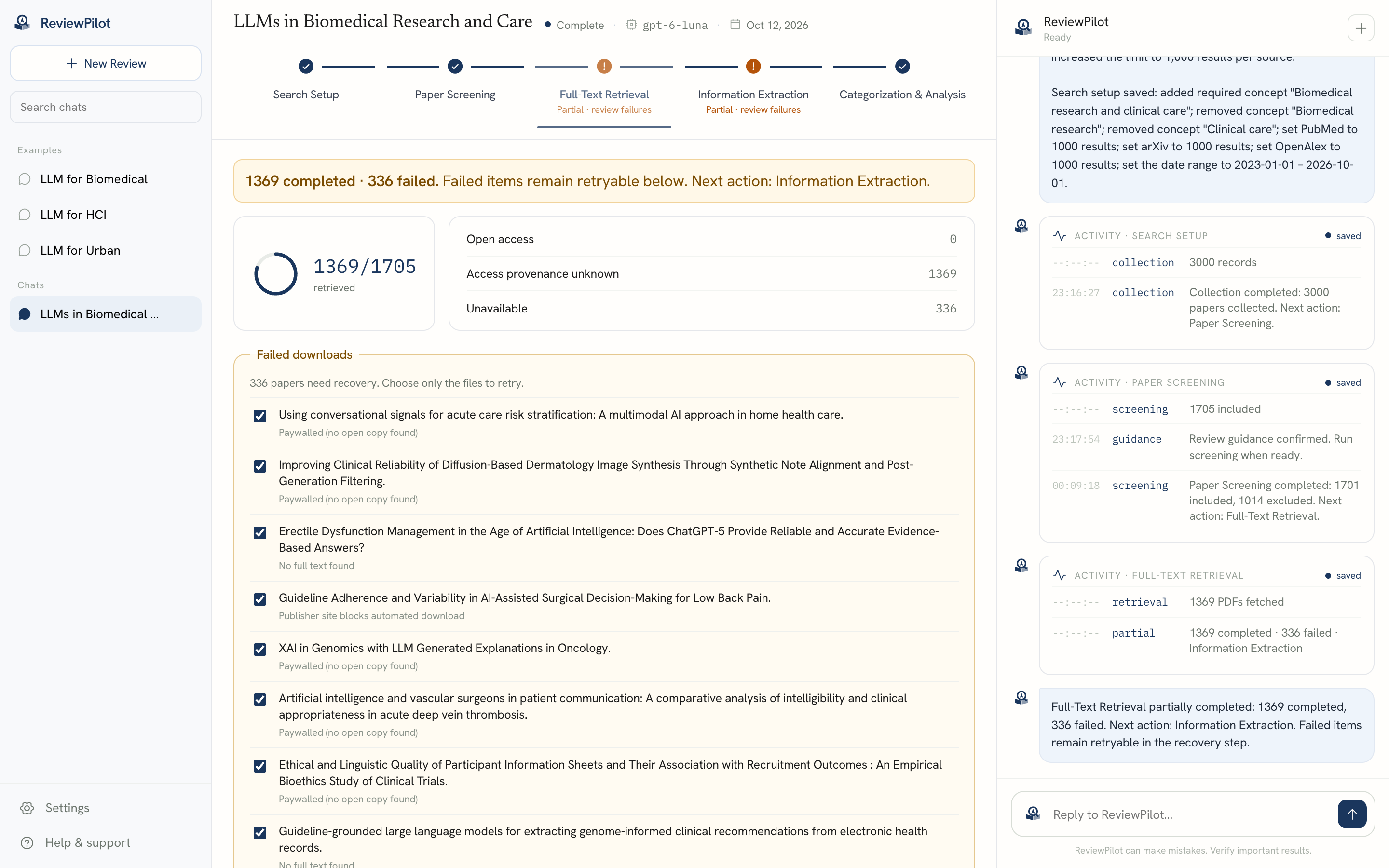}
  \caption{Full-Text Retrieval. PDFs are retrieved for 1,369 of 1,705 records. Each of the 336 failures is listed with its reason and can be retried or supplied manually.}
  \label{fig:ui-retrieval}
\par\medskip
  \uishot{\textwidth}{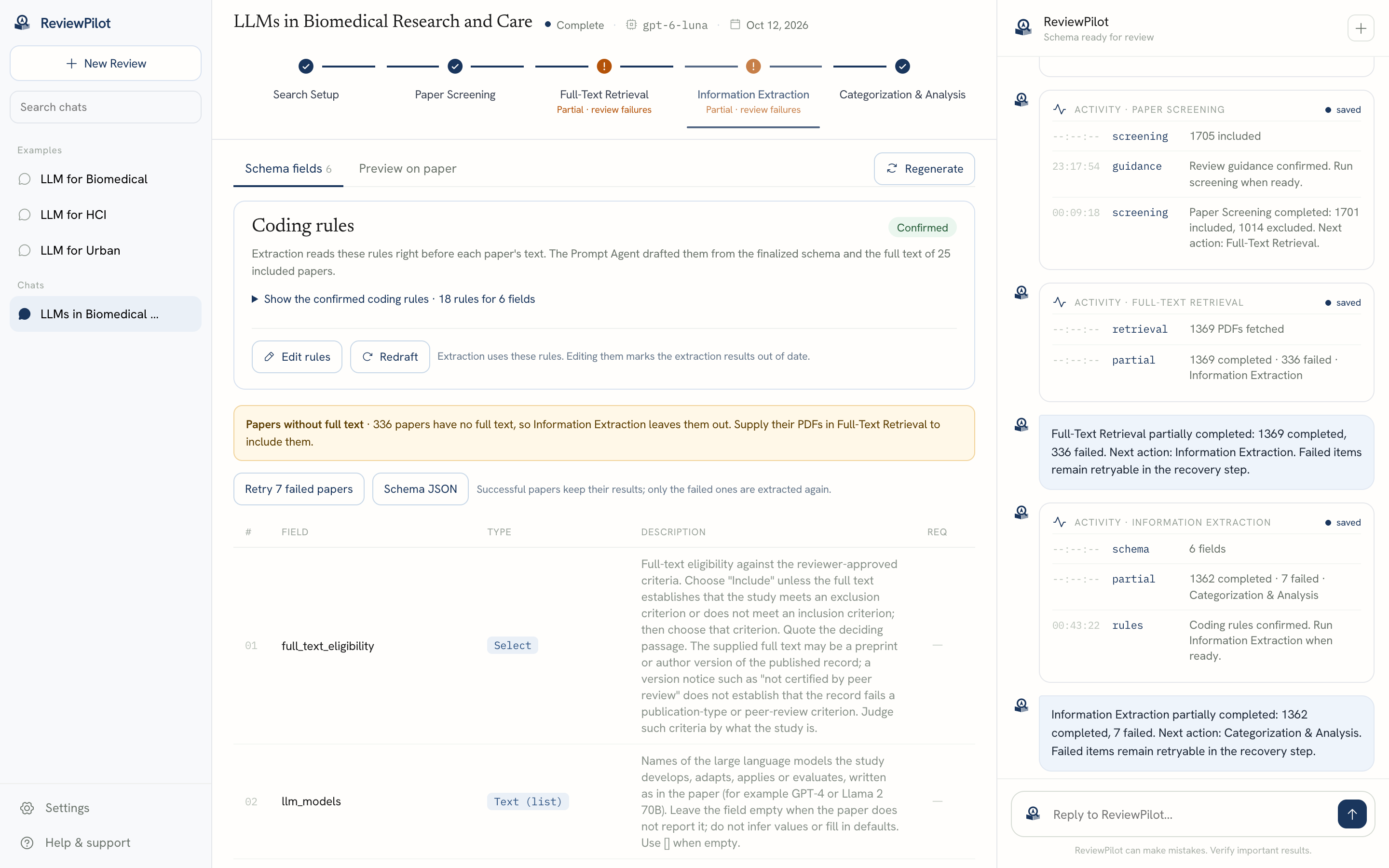}
  \caption{Information Extraction displays the confirmed coding rules (drafted by the Prompt Agent from the finalized schema and the full text of 25 included papers), the notice on the 336 papers without full text, and the first two fields of the six-field schema.}
  \label{fig:ui-schema}
\end{figure*}

\begin{figure*}[p]
  \centering
  \uishot{0.84\textwidth}{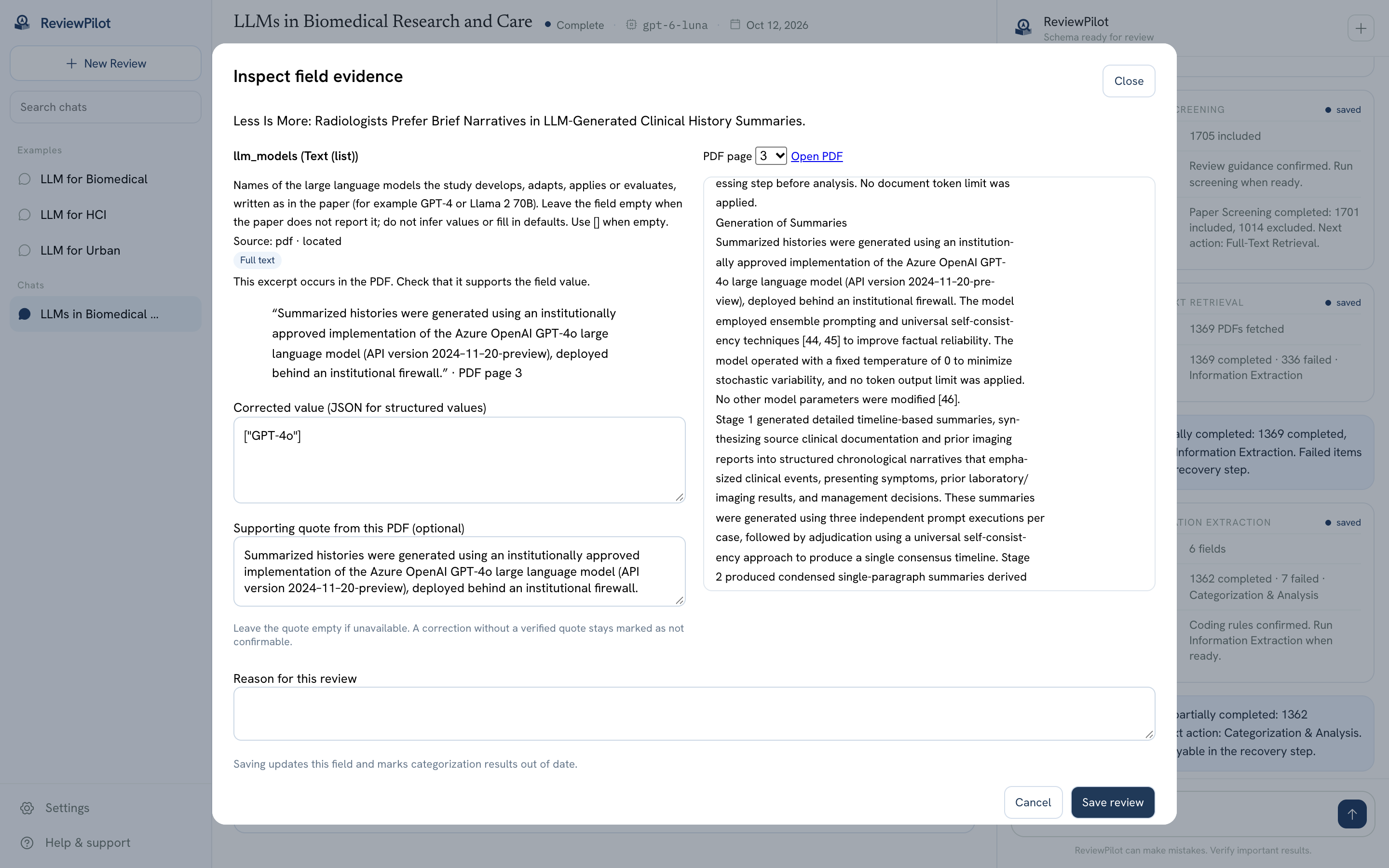}
  \caption{Inspecting an extracted value of the radiology study in Figure~\ref{fig:case-biomedical} (example 3). The excerpt supporting \texttt{llm\_models} = GPT-4o is located verbatim on page 3 of the PDF and shown beside the page text. The researcher can correct the value with a supporting quote and a reason.}
  \label{fig:ui-evidence}
\par\medskip
  \uishot{0.8\textwidth}{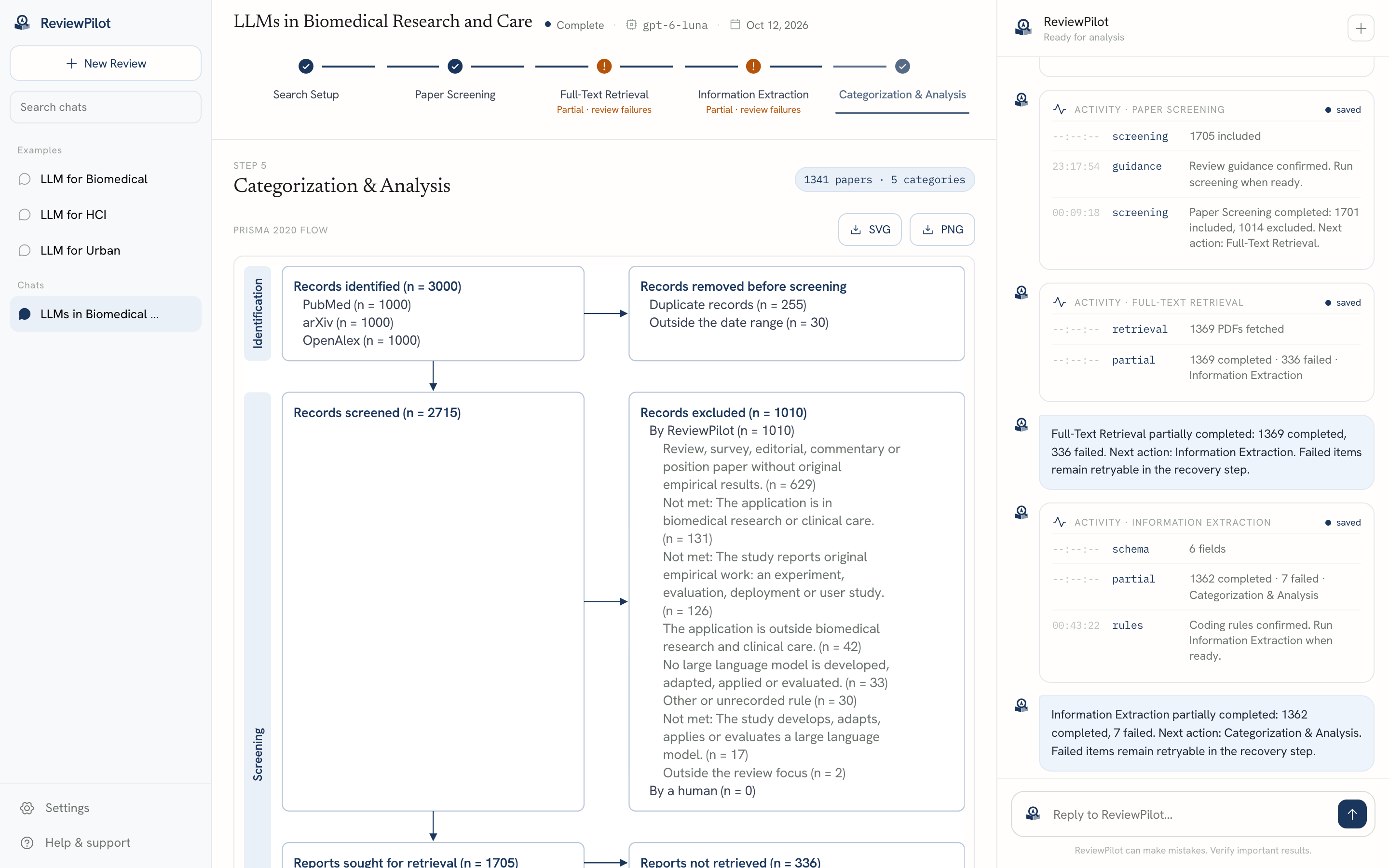}
  \\[1pt]{\small (a)}\\[6pt]
  \uishot{0.72\textwidth}{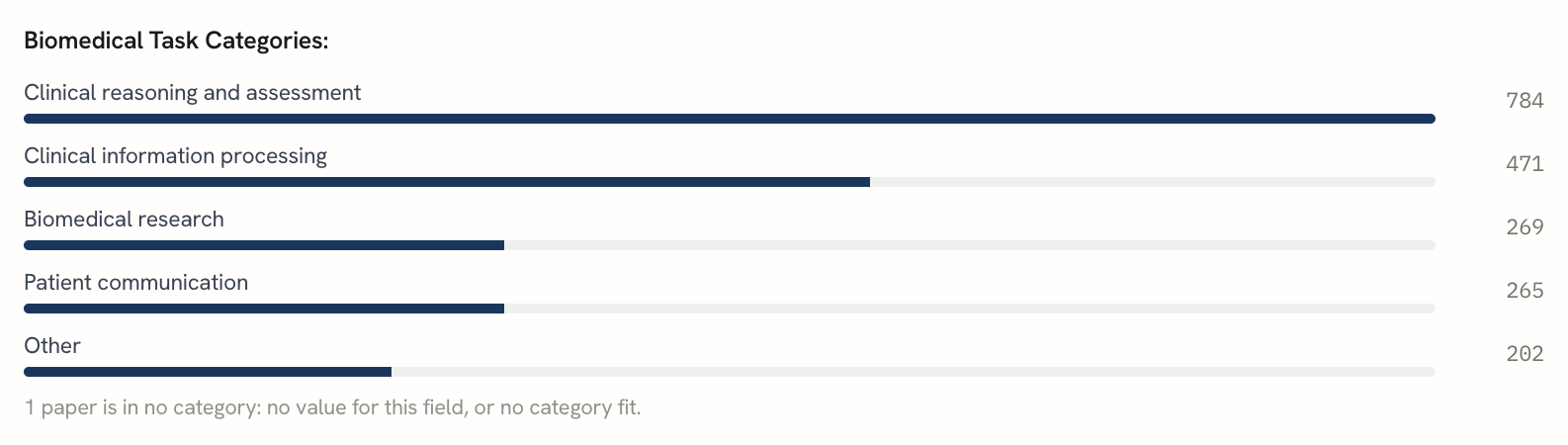}
  \\[1pt]{\small (b)}
  \caption{Categorization \& Analysis. (a) The top of the stage shows 1,341 papers in five categories and the upper part of the PRISMA 2020 flow. (b) Category counts for the biomedical task field. A study can belong to several categories.}
  \label{fig:ui-categorization}
\end{figure*}

\begin{figure*}[p]
  \centering
  \uishot{0.95\textwidth}{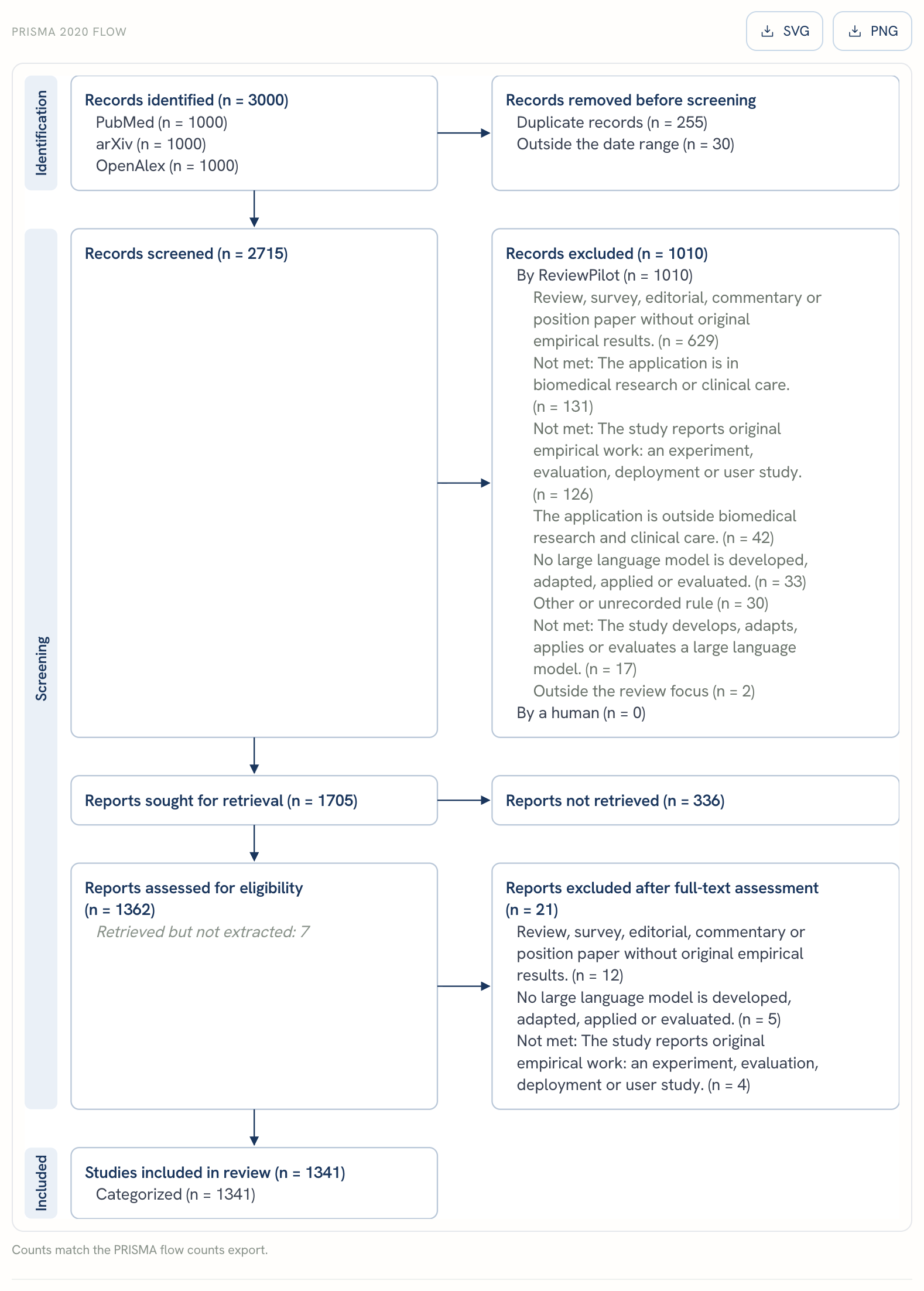}
  \caption{The PRISMA 2020 flow that ReviewPilot generates for the run, from 3,000 identified records to 1,341 included and categorized studies, with exclusion counts by reason.}
  \label{fig:ui-prisma}
\end{figure*}

\end{document}